# Explaining What Machines See: XAI Strategies in Deep Object Detection Models

FatemehSadat Seyedmomeni [0009-0005-9772-0593] and Mohammad Ali Keyvanrad [0000-0002-7654-1001]

[1] Faculty of Electrical & Computer Engineering Malek Ashtar University of Technology Tehran,
Tehran, Iran
[2] Faculty of Electrical & Computer Engineering Malek Ashtar University of Technology Tehran,
Tehran, Iran

f.seyedmomeni@mut.ac.ir ,keyvanrad@mut.ac.ir

**Abstract.** In recent years, deep learning has achieved unprecedented success in various computer vision tasks, particularly in object detection. However, the black-box nature and high complexity of deep neural networks pose significant challenges for interpretability, especially in critical domains such as autonomous driving, medical imaging, and security systems. Explainable Artificial Intelligence (XAI) aims to address this challenge by providing tools and methods to make model decisions more transparent, interpretable, and trustworthy for humans.

This review provides a comprehensive analysis of state-of-the-art explainability methods specifically applied to object detection models. The paper begins by categorizing existing XAI techniques based on their underlying mechanisms—perturbation-based, gradient-based, backpropagation-based, and graph-based methods. Notable methods such as D-RISE, BODEM, D-CLOSE, and FSOD are discussed in detail. Furthermore, the paper investigates their applicability to various object detection architectures, including YOLO, SSD, Faster R-CNN, and EfficientDet.

Statistical analysis of publication trends from 2022 to mid-2025 shows an accelerating interest in explainable object detection, indicating its increasing importance. The study also explores common datasets and evaluation metrics, and highlights the major challenges associated with model interpretability.

By providing a structured taxonomy and a critical assessment of existing methods, this review aims to guide researchers and practitioners in selecting suitable explainability techniques for object detection applications and to foster the development of more interpretable AI systems.

**Keywords:** Explainable Artificial Intelligence, Explainable AI, XAI, Object Detection, Deep Learning, Saliency Maps.

## 1 Introduction

In recent decades, machine learning has emerged as a pivotal branch of artificial intelligence, experiencing rapid development. Among its subfields, deep learning has distinguished itself as a significant domain, leveraging deep neural networks to deliver remarkable capabilities in extracting complex features from raw data. By employing multiple hidden layers, deep neural networks play a crucial role in identifying intricate patterns and uncovering latent structures within datasets [1].

Despite the remarkable achievements of deep learning across various domains, one of its fundamental challenges remains the explainability and high complexity of its models. Due to the multilayered structure and nonlinear behavior of these models, understanding their decision-making processes poses significant difficulties for humans. Furthermore, artificial intelligence models do not always learn effectively from data, as inappropriate selection of hyperparameters or model architecture can lead to suboptimal performance. This issue is particularly critical in sensitive applications, such as object detection, where precision is paramount [2].

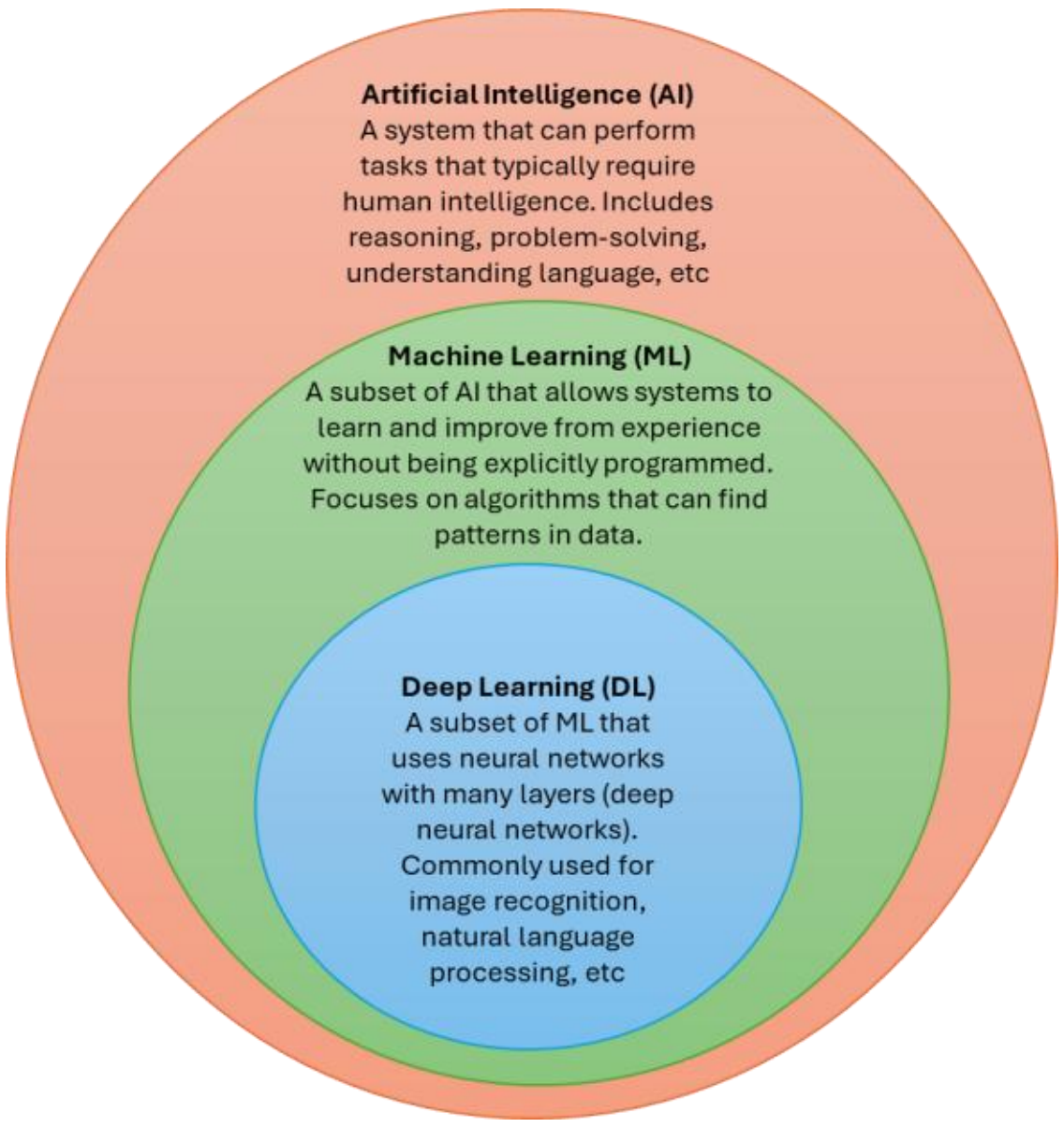

*Figure 1: Hierarchical Relationship Between Artificial Intelligence (AI), Machine Learning (ML), and Deep Learning (DL). Machine learning is a subset of artificial intelligence that focuses on the development of algorithms capable of learning from data. Deep learning, in turn, is a branch of machine learning that utilizes deep neural networks to perform complex tasks, such as natural language processing and image recognition.*

Explainability methods in machine learning can be broadly categorized into two types based on the model: white-box and black-box. White-box models, owing to their simple and transparent structure, are inherently interpretable and do not require complex techniques to understand their functionality. Examples of such models include short decision trees, simple linear models, and regressions with a limited number of features. In contrast, black-box models, such as deep neural networks, random forests, and support vector machines, possess complex and opaque structures, necessitating the development of specialized methods to interpret their decisions. These methods aim to enhance the transparency and reliability of deep learning models, particularly in sensitive applications such as object detection, and have garnered significant attention [3].

To investigate the trends in research conducted in the field of explainable artificial intelligence (AI) for object detection, searches were performed on the Google Scholar database using combined keywords, including "Explainable AI" and "Object Detection," as well as "Explainable Artificial Intelligence" and "Object Detection." The results of these searches indicate that scientific output in this field has exhibited a growing trend in recent years.

Based on the retrieved statistics, a total of approximately 9,860 articles were found using the combined keywords "Explainable AI" and "Object Detection." When categorized by year, 1,070 articles were published in 2022, with an average monthly publication rate of approximately 89.2 articles. This figure increased significantly in 2023 to 1,910 articles, reflecting an average monthly rate of about 159.2 articles. In 2024, the number of articles rose to 3,680, with an average monthly rate of approximately 306.7 articles. Finally, in the first five months of 2025, 1,880 articles were published, corresponding to an average monthly rate of about 376 articles. A comparison of these statistics with previous years reveals substantial growth; specifically, the average monthly publication rate in 2025 is approximately 4.2 times higher than in 2022, 2.4 times higher than in 2023, and 1.2 times higher than in 2024.

A similar trend is observed when using the more comprehensive keyword combination "Explainable Artificial Intelligence" and "Object Detection." This combination yielded a total of approximately 6,810 articles. In 2022, 864 articles were published (average monthly rate: 72 articles). This number increased to 1,520 articles in 2023 (average monthly rate: 126.7 articles) and further to 2,380 articles in 2024 (average monthly rate: 198.3 articles). In the first five months of 2025, 1,030 articles were published, corresponding to an average monthly rate of approximately 206 articles. Consequently, the average monthly publication rate in 2025 is approximately 2.9 times higher than in 2022, 1.6 times higher than in 2023, and slightly higher than in 2024.

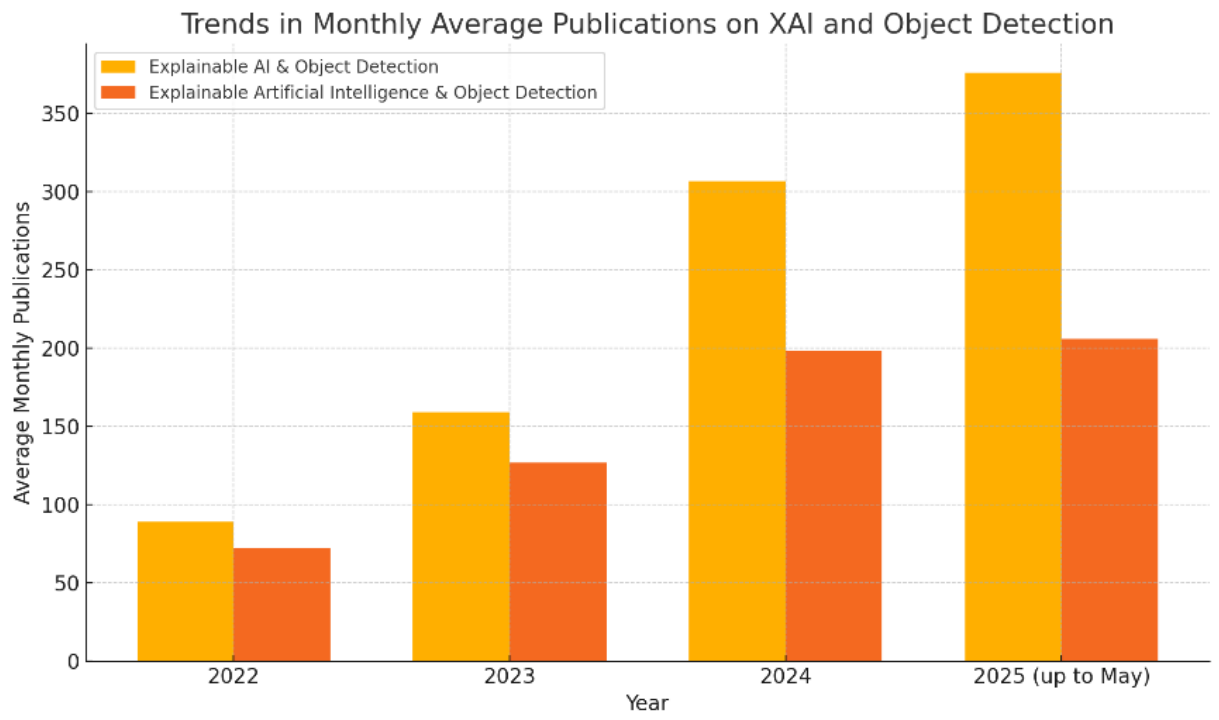

*Figure 2: The trend in the average monthly publication rates for articles using the combined keywords "Explainable AI" and "Object Detection," as well as "Explainable Artificial Intelligence" and "Object Detection," over the period from 2022 to May 2025 is presented. As observed, there has been a substantial increase in the number of articles in recent years, particularly from 2023 onward, indicating growing interest within the research community in the topic of explainable artificial intelligence in applications related to object detection.*

Overall, the increase in the average monthly publication rate for articles using both keyword combinations in 2025 (up to the end of May) compared to previous years reflects a burgeoning research interest in the topic of explainable artificial intelligence (AI) within the domain of object detection. This upward trend, particularly pronounced in 2024 and continuing into 2025, is likely attributable to the development and expanding applications of explainable AI technologies in critical systems, such as autonomous vehicles, surveillance systems, and security and medical applications. It is noteworthy that significant articles published prior to these dates were also reviewed. However, despite the high volume of articles, only a subset has been thoroughly examined. Additionally, many of these articles appeared in searches across different keyword combinations, indicating a close conceptual overlap and interconnection between the searched terms in the fields of explainable AI and object detection.

In this study, we first explore the fundamental concepts and categorize existing methods in the field of explainable artificial intelligence (XAI). We introduce several prominent methods related to classification. Subsequently, we analyze and review the models proposed for explainability in object detection. Following this, the datasets used in this domain and the existing evaluation methods for assessing the quality and degree of explainability of models are examined. The objective of this study is to provide a comprehensive perspective on recent advancements and to identify the challenges and opportunities in developing explainable models for object detection.

# 1 Fundamental Concepts

In this section, we first review the foundational concepts related to explainable artificial intelligence and object detection. Subsequently, we examine the primary mechanisms of the explainability methods employed in this study.

## 1.1 Explainable Artificial Intelligence

Explainable Artificial Intelligence (XAI) refers to a collection of methods and techniques developed to elucidate the decisions and predictions of complex machine learning models. Models such as deep neural networks, random forests, and support vector machines, despite their robust performance, are often regarded as "black-box" models due to their nonlinear and multilayered structures, which render their internal logic challenging for human comprehension [4].

XAI methods aim to enhance the transparency of these models by generating explanations that are understandable to humans. These explanations may take the form of saliency maps, simple surrogate models, feature attribution analyses, or textual descriptions. The objectives of these methods include fostering a better understanding of the decision-making processes of complex models, increasing user trust in the outputs of AI systems, and facilitating the safe and effective use of these systems in critical applications [5], [6].

Enhancing the Explainability of machine learning models offers numerous advantages, including aiding developers in debugging models, ensuring compliance with legal requirements in critical domains, and enabling the evaluation or contestation of automated decisions. Particularly in fields such as object detection, where model outputs can have significant practical implications, the need for reliable methods to explain model decisions becomes increasingly critical [7].

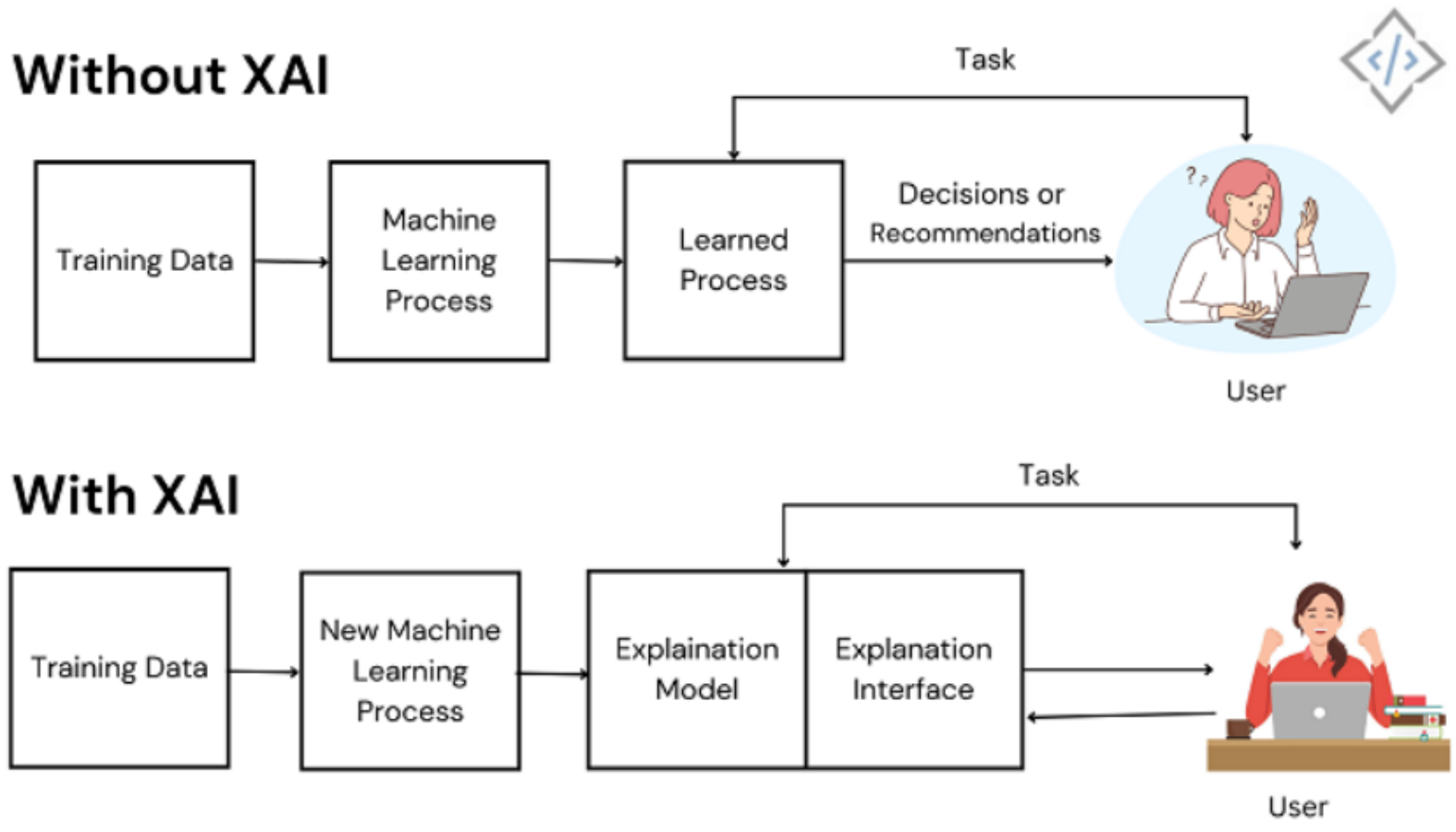

*Figure 3: A comparison of the decision-making process in machine learning systems without (top) and with (bottom) the use of Explainable Artificial Intelligence (XAI). In the absence of XAI, machine learning models provide only the final output, without offering any insight into the reasoning behind their decisions, thereby leaving users unable to comprehend the model's logic. In contrast, XAI introduces an explanatory model and an explanation interface that enable users to understand the rationale behind the model's behavior. This added transparency fosters greater trust, user acceptance, and accountability in intelligent system [8].*

Despite significant advancements in developing explainability methods for classification models, research on the explainability of object detection models remains limited and relatively scarce [9]. The present study addresses this gap by examining explainable methods and techniques in the domain of object detection.

## 1.2 Object Detection in Deep Learning

Object detection is a fundamental task in computer vision that aims to identify instances of visual objects from specific classes (such as humans, animals, vehicles, or buildings) within digital images or video frames. In this task, deep learning models are employed to recognize various objects and determine their precise locations. In other words, addressing the question of "what objects are where?" is central to this process [10].

In certain methods examined within the domain of explainable artificial intelligence (XAI), their applicability is limited to specific types of object detection model architectures. Specifically, these methods may be designed exclusively for single-stage or two-stage models. Accordingly, based on the approach to the identification and prediction process, object detection models can be broadly categorized into two main groups: two-stage models and single-stage models [10].

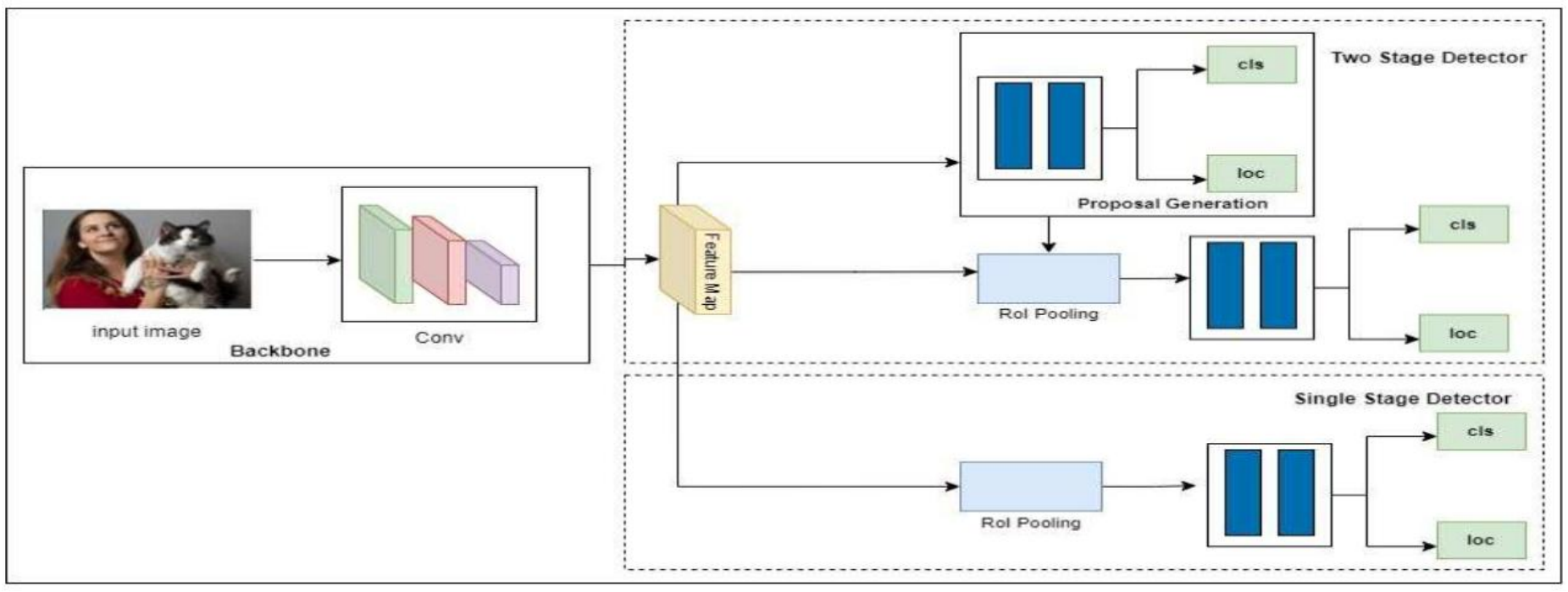

*Figure 4: Comparison of Two-Stage and Single-Stage Object Detection Architectures. In two-stage detectors (top), the process begins with the generation of region proposals, which are subsequently classified and refined into final detections. In contrast, single-stage detectors (bottom) perform object localization and classification in a unified step by directly predicting bounding boxes and associated class scores from the input image [11].*

### 1.1.1 Two-Stage Models

In these approaches, the detection process comprises two distinct steps: first, region proposals, which are areas likely to contain objects, are extracted; then, each region is individually analyzed to predict the object's class and location. This structure typically offers higher accuracy but incurs greater computational costs [10].

#### *1.1.1.1 R-CNN*

R-CNN (Regions with Convolutional Neural Networks) is a foundational approach in object detection that is built upon the idea of utilizing region proposals in conjunction with convolutional neural networks (CNNs). This algorithm begins by extracting approximately 2,000 class-independent region proposals from an input image using techniques such as Selective Search. Each proposed region is then resized (warped) to match the fixed input dimensions required by the neural network. A deep convolutional network is subsequently applied to each region to extract its feature representations.

These extracted features are then passed to a linear classifier, typically a Support Vector Machine (SVM), to determine whether the region contains an object belonging to a specific class. To further enhance localization accuracy, a separate bounding box regression stage is employed to refine the coordinates of the proposed regions.

R-CNN achieved remarkable accuracy in object detection by leveraging pre-trained models on large-scale datasets (e.g., ImageNet) and fine-tuning them on smaller, task-specific datasets (e.g., PASCAL VOC). Its architecture is modular and extensible by design. By integrating region proposal methods with the representational power of convolutional networks, R-CNN laid the groundwork for a new generation of object detection algorithms [12].

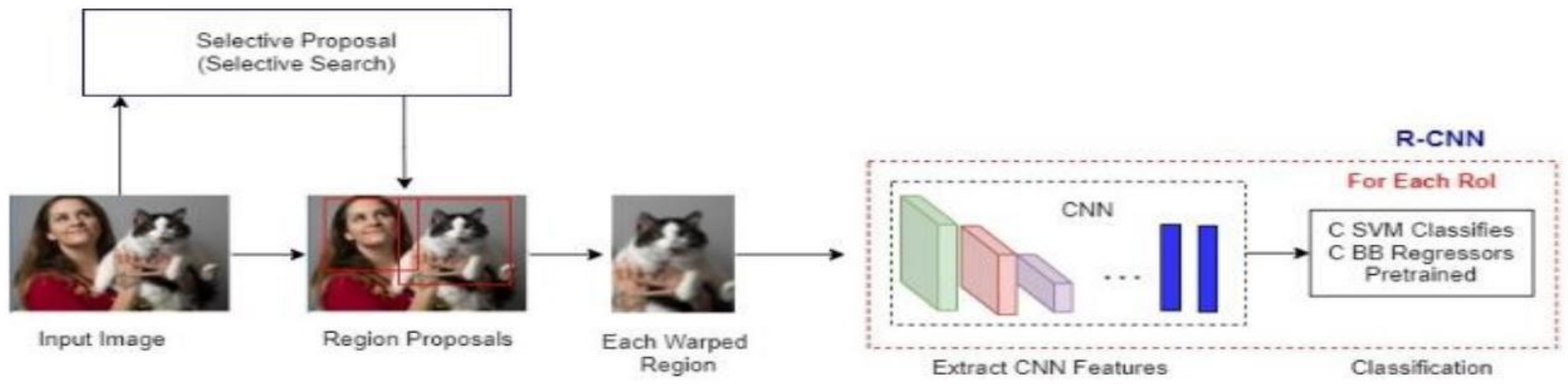

*Figure 5: R-CNN Framework [11].*

### *1.1.1.2 Fast R-CNN*

Fast R-CNN is a method for object detection that offers both high accuracy and efficient processing speed. It receives the entire input image along with a set of proposed regions—known as Regions of Interest (RoIs)—as input. Initially, the image is passed through convolutional layers and sampling operations to generate a shared feature map.

For each proposed region, a fixed-size feature vector is extracted using a Region of Interest (RoI) Pooling layer. This vector is then fed into fully connected layers whose output consists of two components: (1) the probability that the region belongs to each class, computed via a softmax function, and (2) the precise coordinates of the bounding box for each class, obtained through regression.

The model is trained in a single-stage, end-to-end manner, utilizing a multitask loss function to simultaneously learn both classification and localization of objects. This architecture enables all layers of the network to be trainable, eliminates the need to store intermediate features, and achieves relatively high speed and accuracy.

Unlike earlier methods that scaled or processed each region independently, Fast R-CNN leverages shared computation across proposed regions, significantly increasing test-time speed while maintaining high detection accuracy [13].

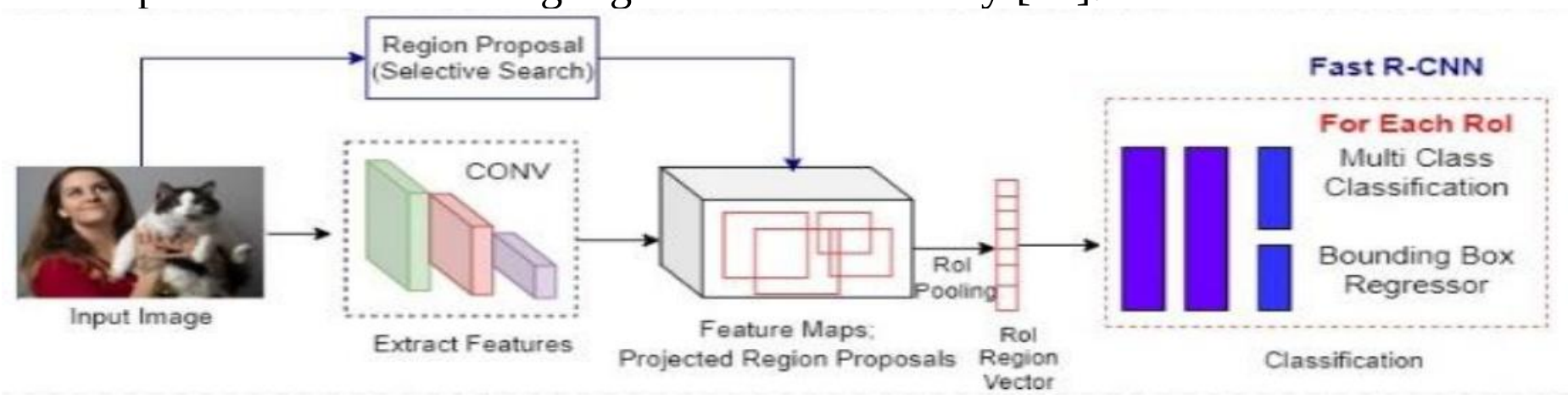

*Figure 6: Fast R-CNN Framework [11].*

### *1.1.1.3 Faster R-CNN*

Faster R-CNN is an advanced framework for object detection, designed as a unified deep neural network. This model comprises two primary components: a Region Proposal Network (RPN), which is responsible for generating candidate regions likely to contain objects, and an object detector based on Fast R-CNN, which classifies these proposed regions and accurately refines their spatial coordinates.

The RPN is a fully convolutional network applied directly to the feature map of the input image. For each spatial location on the feature map, it generates a set of region proposals—referred to as anchors—with various scales and aspect ratios. For each anchor, the network predicts both the likelihood of object presence and its precise bounding box coordinates.

A key feature of Faster R-CNN is the complete sharing of computations between the RPN and the final detector. The feature map is computed only once and utilized by both stages of the model. This architectural design significantly reduces computational time while enhancing detection accuracy.

The entire framework is end-to-end trainable and employs multi-task learning to simultaneously perform both classification and bounding box regression. Due to its high accuracy, ability to detect objects at various scales and aspect ratios, and its acceptable inference speed, Faster R-CNN is recognized as one of the standard methods in object detection [14].

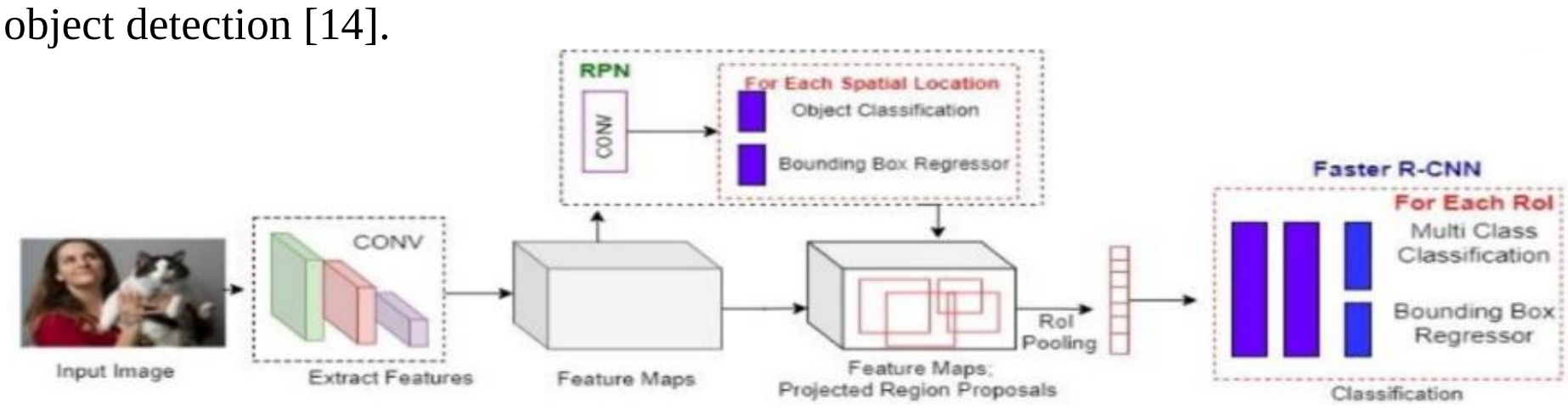

*Figure 7: Faster R-CNN Framework [11].*

### 1.1.2 Single-Stage Models

Single-stage models directly predict the location and class of objects in a single step, eliminating the separate region proposal stage. This architectural simplification enhances processing speed, making these models more suitable for real-time applications. However, their accuracy is generally slightly lower compared to two-stage models [10].

#### *1.1.2.1 YOLO (You Only Look Once)*

YOLO (You Only Look Once) is a unified and extremely fast method for object detection in images, which formulates detection as a direct regression problem from the input image to the bounding box coordinates and class probabilities. Unlike traditional approaches that involve multiple distinct stages—such as region proposal generation, feature extraction, and classification—YOLO identifies all objects in a single glance at the entire image.

The model divides the input image into a grid, with each cell in the grid responsible for predicting bounding boxes and the probability of the presence of object classes within its region. The YOLO architecture is designed such that its convolutional neural network performs all predictions globally and simultaneously. Due to its holistic view of the entire image, the model also benefits from contextual information.

This streamlined and cohesive design enables YOLO to operate with remarkable speed—achieving rates of up to 155 frames per second—making it well-suited for real-time applications such as computer vision, robotics, and autonomous vehicles.

Although YOLO demonstrates high accuracy in many scenarios, it faces certain limitations in precisely detecting small or closely spaced objects. Nevertheless, its ability to learn generalized and robust representations of objects allows it to perform effectively across a wide range of domains [15].

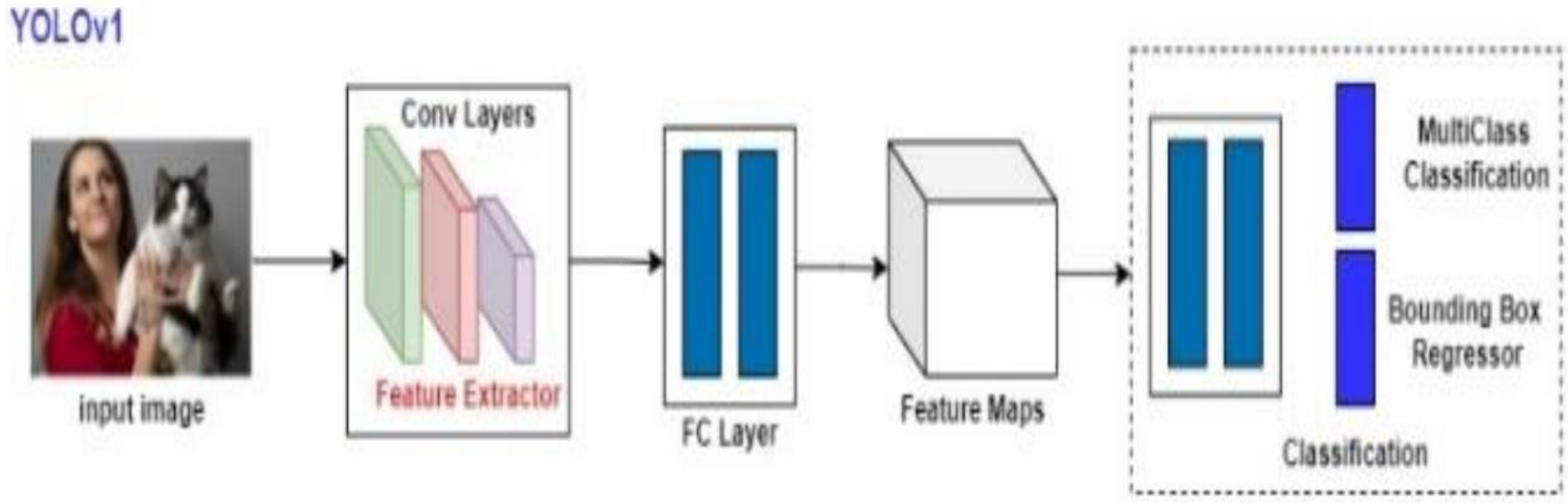

Figure 8: YOLOv1 Framework *[11]*.

*1.1.2.2 Single Shot MultiBox Detector (SSD)*

The Single Shot MultiBox Detector (SSD) is a fast and efficient method for object detection in images that performs all predictions in a single forward pass through a neural network, eliminating the need for separate stages such as region proposal generation or feature resampling. In this approach, a set of default boxes with various scales and aspect ratios is assigned to the cells of multiple feature maps at different levels of the network.

At each location within these feature maps, the network uses small convolutional filters to predict both the object class and the offsets of the bounding boxes relative to their corresponding default positions. The use of multiple layers with different resolutions enables SSD to effectively detect objects of varying sizes.

The training process involves matching the default boxes to the ground truth boxes based on Jaccard overlap (also known as the Intersection over Union, or IoU), and employing a combined loss function that accounts for both localization error and confidence (classification) loss. Due to its unified design, elimination of computationally expensive stages, and simultaneous utilization of multiple scales, SSD maintains high accuracy while offering excellent processing speed, making it well-suited for real-time applications such as machine vision and embedded systems [16].

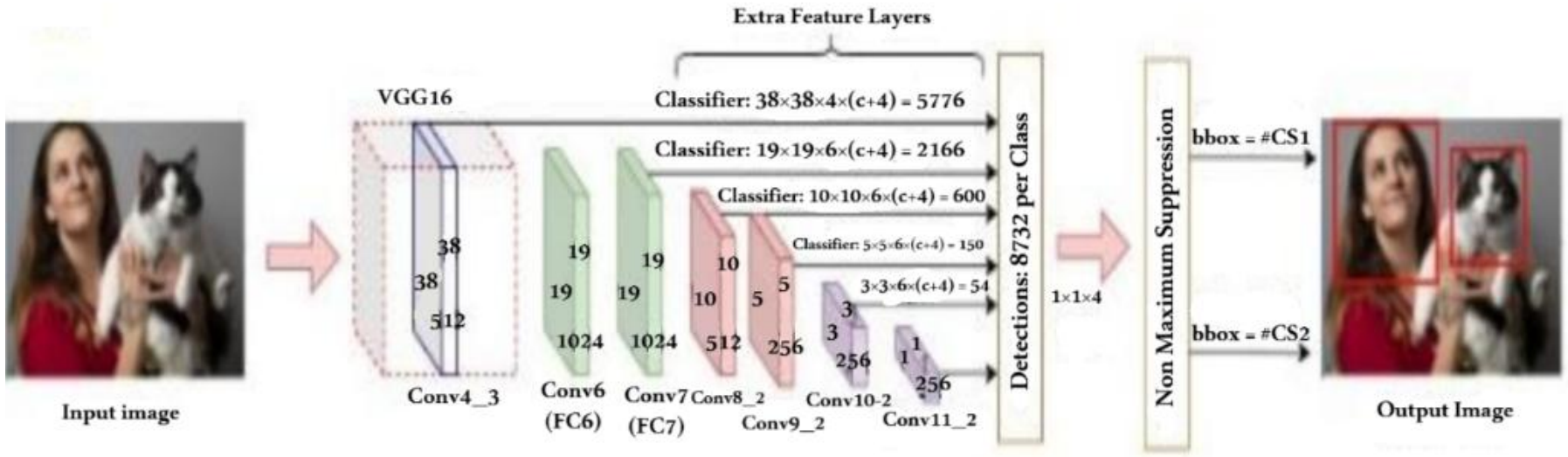

*Figure 9: SSD Framework [11].*

### *1.1.2.3 EfficientDet*

EfficientDet is an object detection architecture designed to simultaneously achieve high accuracy and computational efficiency. This model accomplishes its goals through the introduction of two key innovations:

I. Bi-directional Feature Pyramid Network (BiFPN)**:**

BiFPN facilitates effective and optimized multi-scale feature fusion by employing both top-down and bottom-up connections, along with learnable weights for each input feature. This design enables the network to prioritize more informative features during fusion, enhancing overall detection performance across various object scales.

II. Compound Scaling Method**:**

This technique uniformly scales the depth, width, and input resolution of all components in the model—including the backbone network, BiFPN, and the class and box prediction heads—using a single compound coefficient. This holistic scaling strategy ensures balanced model growth and performance improvements without disproportionately increasing computational cost.

EfficientDet utilizes EfficientNet as its backbone and employs depthwise separable convolutions, allowing the model to maintain high accuracy while reducing the number of parameters and computational operations. As a result, it offers a family of models (D0 to D7) that perform well across a wide range of resource constraints, from mobile devices to high-performance servers.

Due to its significantly lower parameter count and computational demands compared to previous models, while maintaining or even surpassing competitive accuracy on standard datasets such as COCO, EfficientDet is recognized as one of the most effective and scalable object detection models in the field [17].

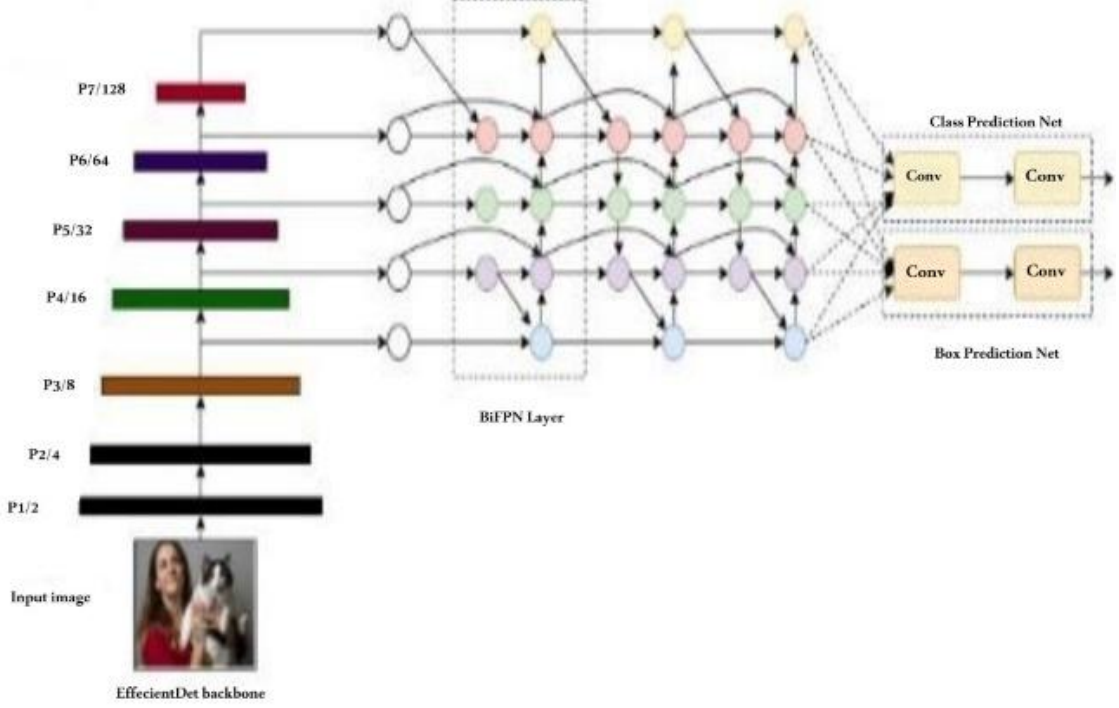

*Figure 10: EfficientDet Framwork [11].*

Each of these models is selected based on specific application requirements, such as the relative importance of accuracy, processing speed, and hardware constraints.

### 1.3 Types of XAI Methods in the Domain of Image Analysis

Numerous techniques have been proposed for visualizing and explaining the behavior of computer vision models after the training phase. Among these, saliency methods play a prominent role and have been primarily developed for image classification models. These classification models serve as the foundation for many more advanced architectures in computer vision, including object detection and image segmentation models. Consequently, explainability methods originally designed for classification models have, in many cases, provided the groundwork or inspiration for the development of explanatory approaches tailored to more complex computer vision tasks [18].

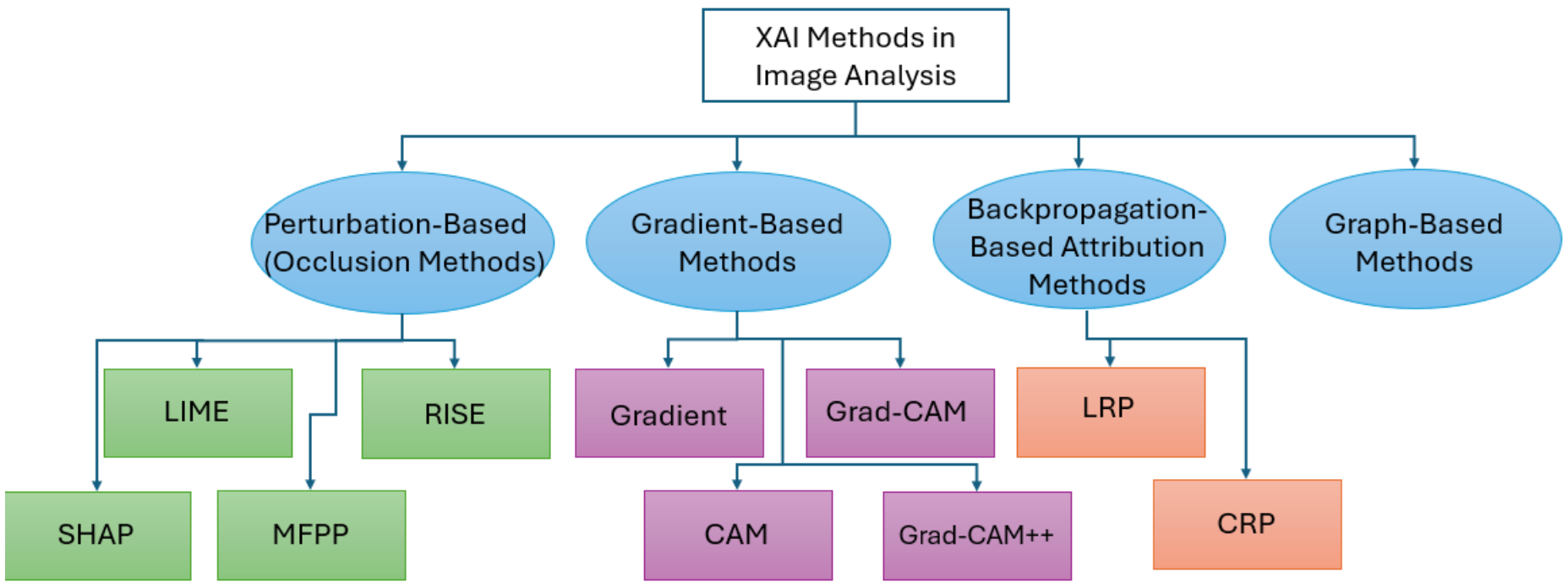

*Figure 11: Overall categorization of XAI methods in the domain of image analysis. These categories form the basis for the detailed review presented in the subsequent sections.*

In this study, the methods used to explain the decisions made by deep learning models can be broadly categorized into four groups, based on their underlying mechanisms:

#### 1.3.1 Perturbation-Based / Occlusion Methods

Perturbation-based methods, such as occlusion techniques, generate saliency maps by directly modifying the model's input and analyzing the effect of these changes on the output. These approaches evaluate the importance of each region in the input by removing or masking parts of the image and comparing the resulting output with the original output. This type of analysis is analogous to sensitivity analysis in control systems and is characterized by its simplicity, model-agnostic nature, and high explainability. Among the most well-known methods in this category are RISE and LIME. Although these methods offer high accuracy, their requirement for numerous forward passes renders them computationally expensive [19], [20].

Some of the most prominent methods within this category for classification tasks include the following:

##### *1.3.1.1 LIME Method*

The LIME algorithm (Local Interpretable Model-Agnostic Explanations) is an explainability technique for machine learning models that focuses on explaining the prediction for a specific instance, rather than explaining the entire model globally. LIME operates by generating a set of random samples in the vicinity of the input instance and evaluating the model's responses to these perturbed samples. It then fits a simple, interpretable model—typically a linear regression model—on the generated data. These samples are weighted based on their proximity to the original instance, thereby allowing the surrogate model to provide a more accurate local approximation. The output of LIME is a set of influential features that had the greatest impact on the model's decision, thus enabling a comprehensible explanation of the behavior of otherwise complex models [2].

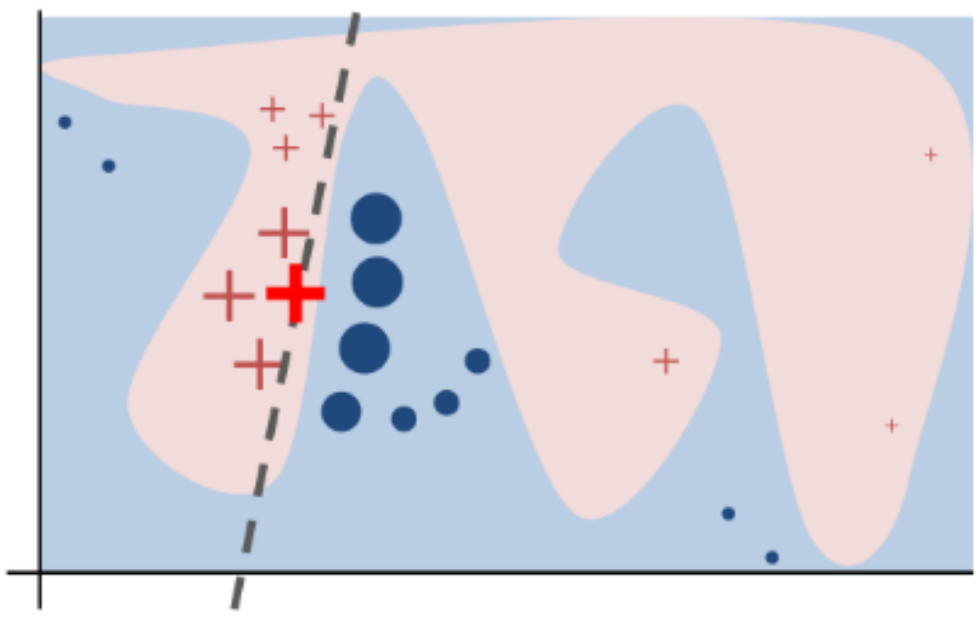

*Figure 12: Visualization of LIME's Functionality: The original model (represented by the background color) exhibits complex behavior that is not explainable at a global level. However, LIME addresses this challenge by sampling points in the vicinity of a specific instance (indicated by the red cross) and training a simple model on these perturbed samples. Through this process, LIME learns a local decision boundary (depicted by the dashed line) that is comprehensible and explainable. This local surrogate model provides a clear insight into the rationale behind the model's prediction [2].*

#### *1.3.1.2 SHAP Method*

The SHAP (SHapley Additive exPlanations) method is an explainability framework for machine learning models, grounded in cooperative game theory. It assigns an importance value to each feature for a specific prediction. By considering all possible permutations of feature inclusion in the model, SHAP calculates the average marginal contribution of each feature. This process results in a fair and explainable attribution of feature influence on the model's output.

SHAP guarantees three critical properties—local accuracy, missingness, and consistency—making it the only theoretically sound and uniquely determined solution for additive feature attribution within this framework [21].

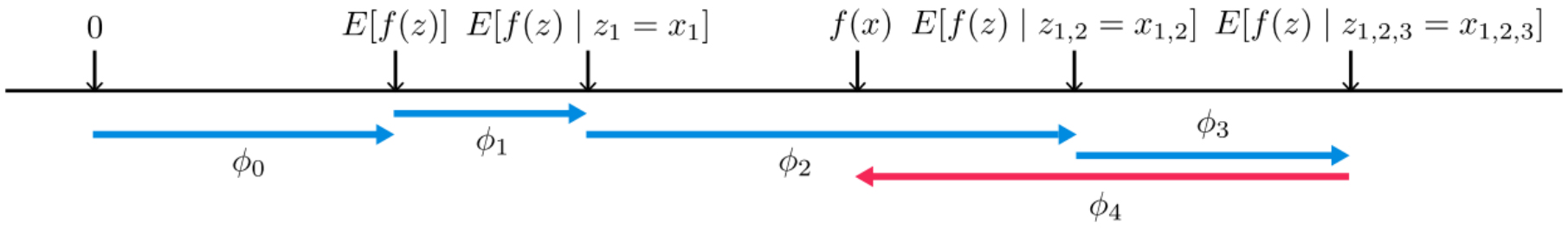

*Figure 13: Schematic Representation of SHAP Value Attribution to Features: In this illustration, the base value corresponds to the average prediction of the model when no input features are considered. Each SHAP value, denoted by $\varphi_i$, quantifies the expected change in the model's output when a specific feature is introduced, averaged over all possible combinations of feature inclusion. This approach is grounded in Shapley value theory from cooperative game theory and leverages the interpretation of conditional expectations to produce a mathematically rigorous and uniquely defined attribution for each feature. As a result, SHAP provides a precise and consistent explanation of individual feature contributions, making it a widely adopted technique in explainable AI [21].*

#### *1.3.1.3 RISE Method*

The RISE (Randomized Input Sampling for Explanation) method is an explainability technique for machine learning models that generates saliency maps for visual inputs without requiring access to the internal structure of the model. This approach operates by generating random masks that modify pixel values and multiplying them with the input image. The model's output is then computed for each masked image. Subsequently, the final saliency map is produced as a weighted linear combination of the masks, where each mask's weight corresponds to the model's confidence score for the respective masked image.

The primary advantage of RISE is that it does not require access to intermediate features or the internal architecture of the model, making it applicable to any vision-based model. Moreover, the use of soft masks results in smooth, noise-free saliency maps that are more consistent with human explanation [22].

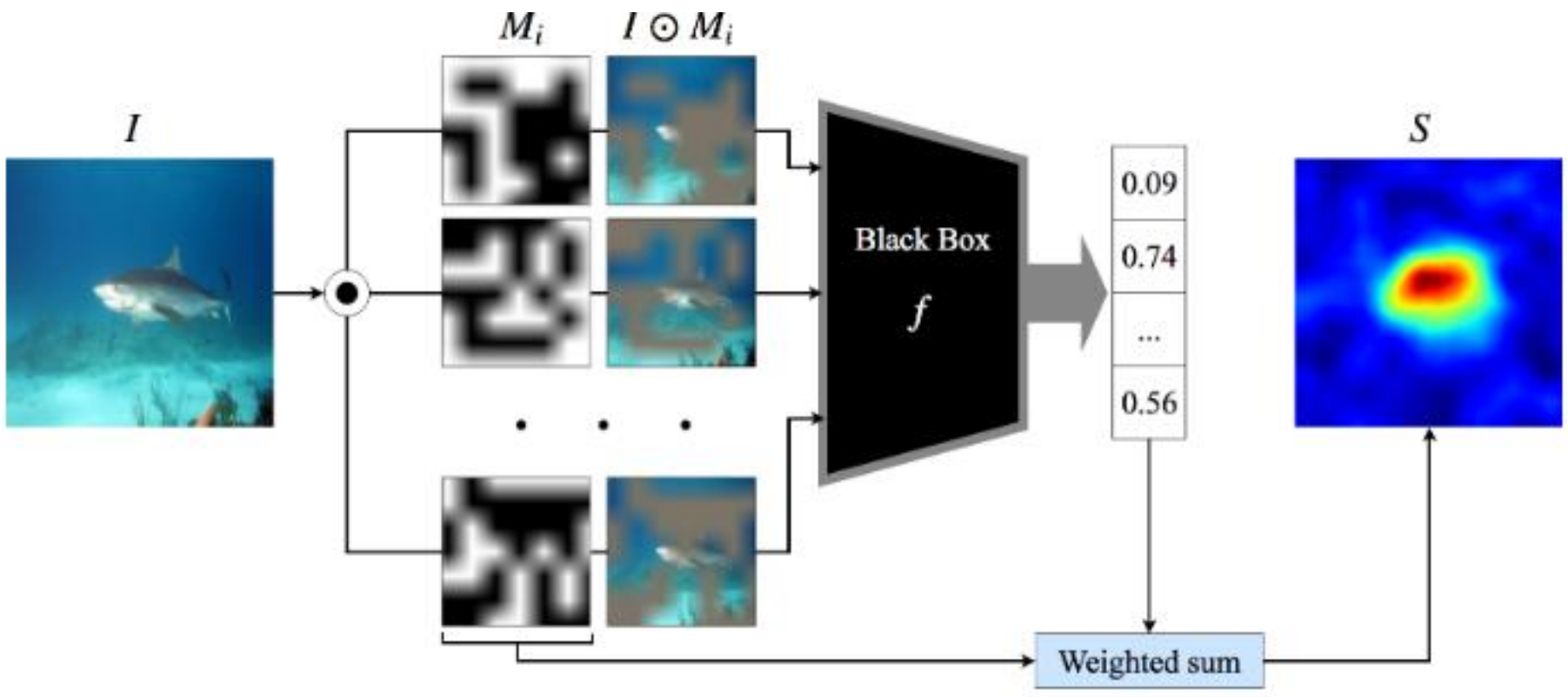

*Figure 14: Overview of the RISE Method: This figure provides an overview of the RISE method. In this approach, the input image is first element-wise multiplied with a set of randomly generated masks, producing multiple variations of the original image with certain regions occluded. These masked images are then fed into a black-box model, and for each image, the model's output (e.g., the probability of belonging to a target class) is computed. These outputs are interpreted as weights, reflecting the relative importance of each mask. Finally, a weighted average of the masks is computed to generate a saliency map, which highlights the regions of the image that had the greatest impact on the model's decision. This entire process is conducted without requiring access to the model's internal architecture or gradient information, making it fully applicable to black-box models [22].*

#### *1.3.1.4 MFPP Method*

MFPP (Morphological Fragmental Perturbation Pyramid) is a model-agnostic technique for interpreting deep learning models that leverages morphological analysis of images and the introduction of structural perturbations to generate high-precision saliency maps. In this method, the input image is segmented into fragments at multiple scales and then randomly masked to evaluate the impact of each fragment on the model's output.

By employing a pyramid-based architecture and incorporating shape, texture, and structural features across different levels of abstraction, MFPP establishes a stronger correlation between visual attributes and semantic concepts. This enables the method to detect object boundaries with higher accuracy.

MFPP utilizes statistical aggregation of the results from the masked inputs to produce explanations with enhanced visual and semantic quality. It is recognized as a lightweight and generalizable solution applicable to a wide range of image classification models [23].

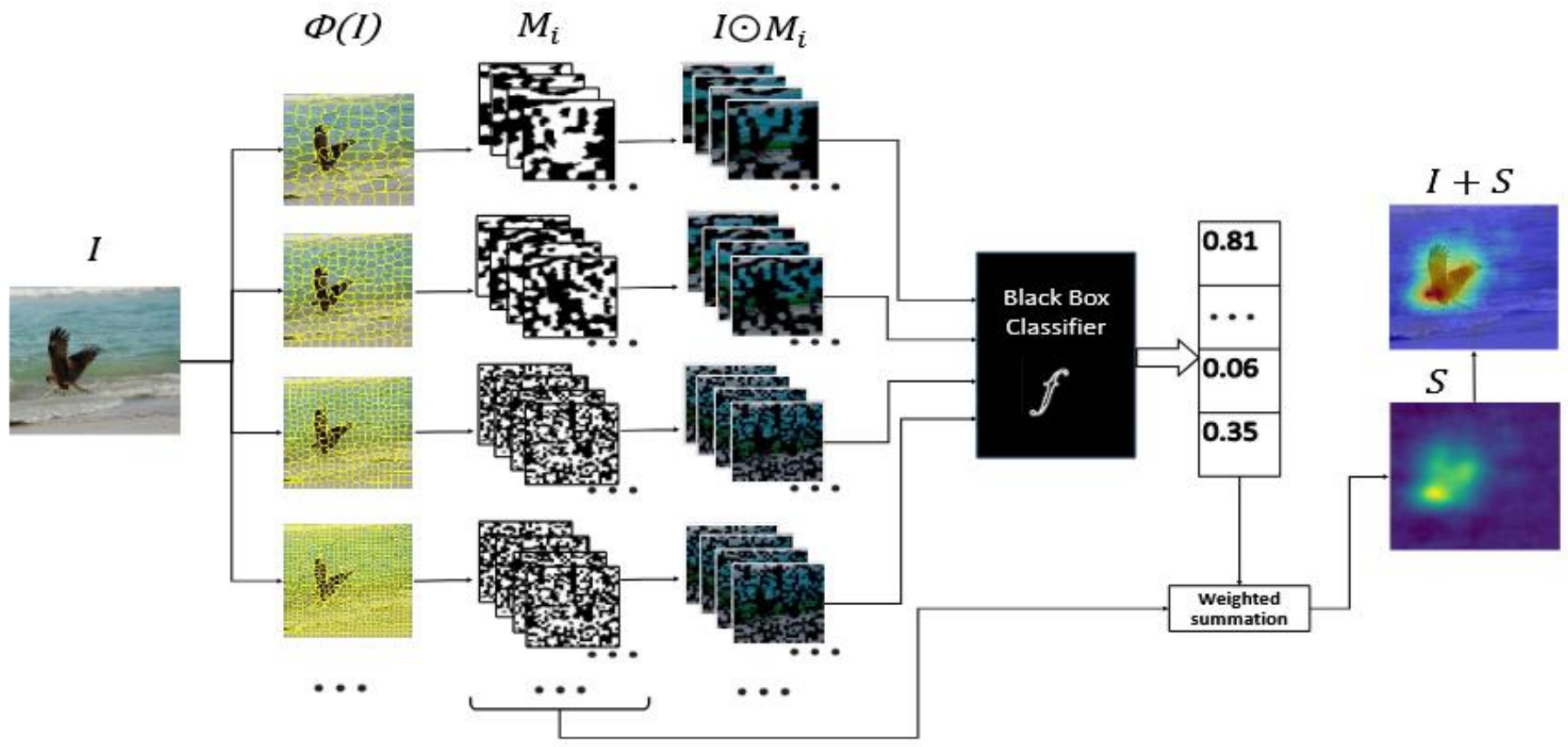

*Figure 15: General Procedure of the MFPP Method: The input image I is first segmented into fragments at multiple scales using the segmentation algorithm $\phi$ . For each scale, random masks $M_i$ are generated and applied element-wise to the original image to produce masked versions of the image. These masked images are then fed into the black-box model f, which computes an importance score for each region based on the model's prediction. Subsequently, these importance scores are aggregated in a weighted manner to produce the final saliency map S, which indicates the influence of each image segment on the model's output. The saliency map, when overlaid on the original image, serves to visually interpret the model's decision-making process [23].*

### 1.3.2 Gradient-Based Methods

In gradient-based methods, the importance of each region in the image is computed using the derivatives of the model's output with respect to the input. These methods generate sensitivity or saliency maps by performing a forward pass followed by a backward pass to calculate gradients. Notable examples include Gradient, Grad-CAM, and more advanced variants such as Grad-CAM++. These approaches are generally faster than occlusion-based methods and are particularly well-suited for convolutional neural networks. However, they may lack precise explainability compared to perturbation-based methods because they do not always reflect actual changes in the model's output [20].

Some of the most prominent methods in this category for classification tasks are as follows:

#### *1.3.2.1 Gradient Methods*

Gradient-based visualization is one of the effective and foundational methods for interpreting deep neural networks, particularly convolutional neural networks (CNNs). This approach calculates the gradient of the output score for a specific class with respect to the pixels of the input image by employing the backpropagation algorithm. The purpose is to identify which pixels have the greatest influence on the class prediction. These gradients are utilized as saliency maps to highlight the most important regions of the image. The core idea behind this method relies on a linear approximation of the class output score function around a particular input image. Owing to its requirement for no additional labeling and the need for only a single backward pass, this method is notably fast and computationally efficient [24].

### *1.3.2.2 CAM Methods*

The Class Activation Mapping (CAM) method is a fundamental technique for interpreting deep learning models, particularly in image recognition. This approach generates a heatmap by utilizing features extracted from the final convolutional layer along with the weights associated with the target class in the fully connected layer, thereby highlighting the important regions of the image that influence the model's prediction. To apply CAM, the model must include a Global Average Pooling (GAP) layer prior to the fully connected layer, allowing the class-specific weights to be applied to the spatial features of the last convolutional layer. The output of this method is a saliency map for each class, indicating the significant regions within the image. However, CAM is limited to specific network architectures and offers less flexibility compared to more advanced variants such as Grad-CAM [25].

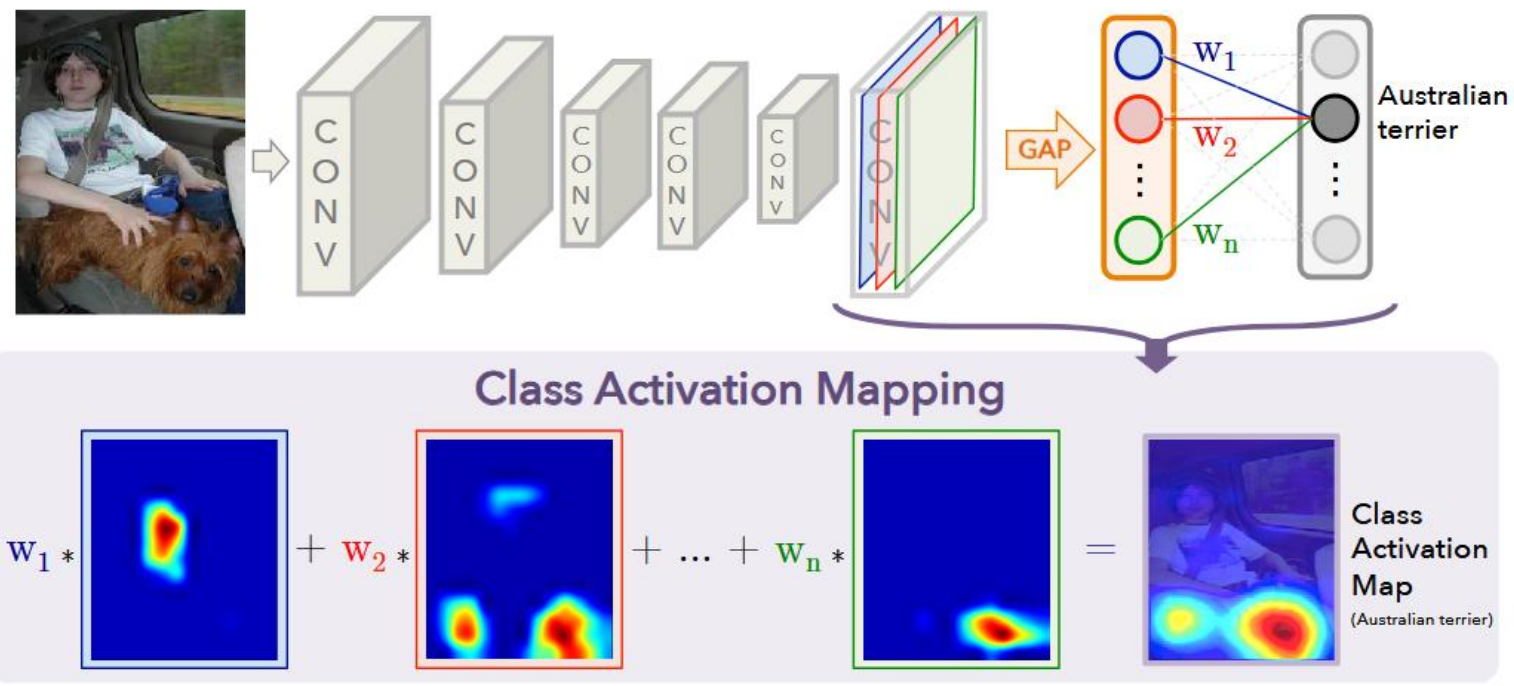

*Figure 16: Process of Generating Class Activation Maps (CAM): In this method, the input image first passes through multiple convolutional layers to extract spatial features. Subsequently, Global Average Pooling (GAP) is applied to the resulting feature maps, producing a feature vector. This vector is then combined with the weights corresponding to each class in the fully connected layer to compute the probability of the image belonging to each class. To generate the activation map for a specific class (e.g., an Australian Terrier dog), the weights of that class are applied to the feature maps of the last convolutional layer, and the weighted maps are summed to highlight the important regions of the image that have the greatest influence on the model's decision. The resulting map is presented as a heatmap overlaid on the input image, emphasizing the highly significant areas [25].*

### *1.3.2.3 Grad-CAM Method*

The Gradient-weighted Class Activation Mapping (Grad-CAM) method is designed to interpret the decisions of computer vision models, particularly convolutional neural networks. In this approach, the gradient of the model's output with respect to the feature maps of the final convolutional layer is first computed. These gradients are then averaged across each channel and used as weights to combine the convolutional layer's feature maps. The weighted combination produces a final heatmap that highlights the regions of the input image that had the greatest influence on the model's prediction.

Unlike the original CAM method, Grad-CAM does not require any modification to the model architecture and leverages gradient information to generate more flexible and generalizable explanations [26].

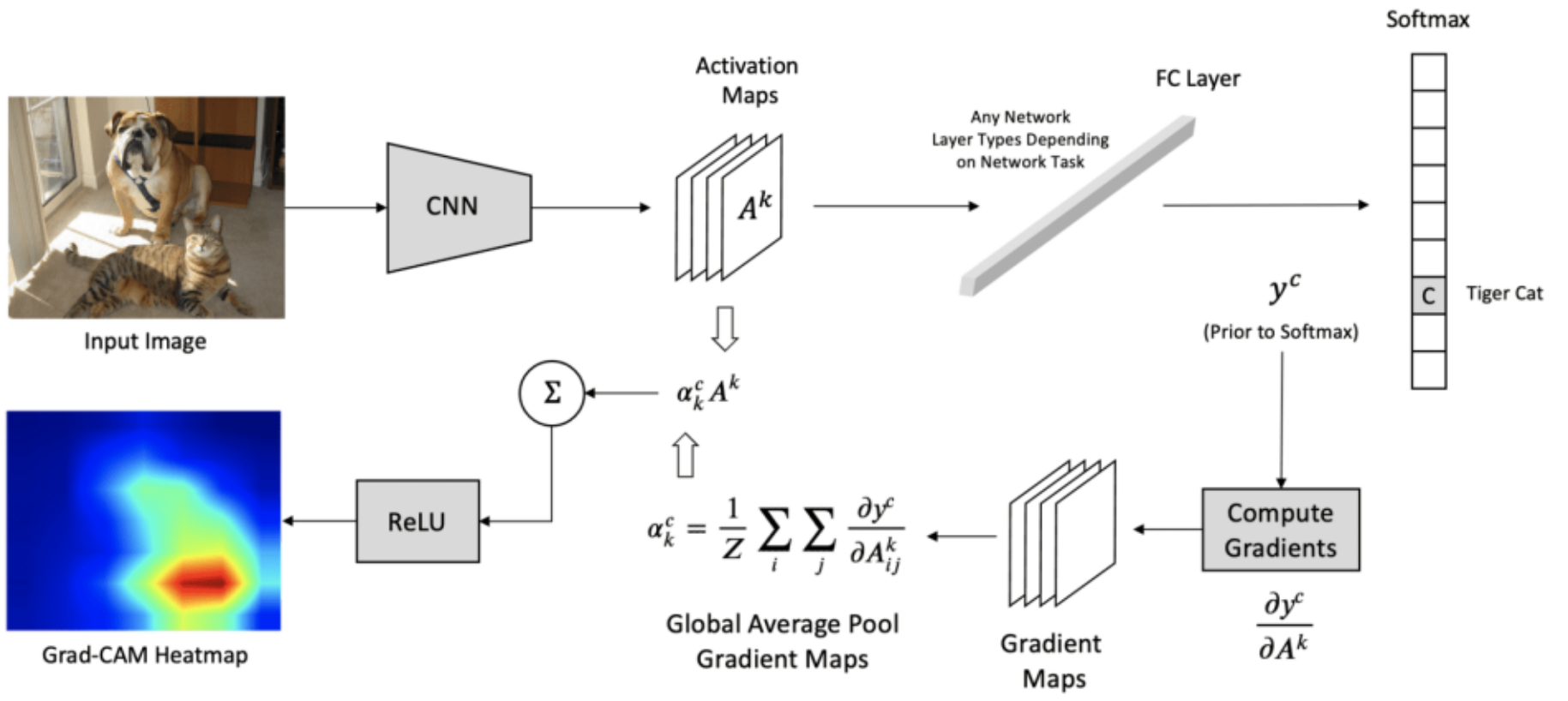

*Figure 17: Overall Architecture of Grad-CAM (Gradient-weighted Class Activation Mapping): This architecture outlines the process of generating an attention heatmap to interpret the decisions of convolutional neural network (CNN) models for a specific class (e.g., "tiger-cat"). Initially, the input image is processed through the CNN, and activation maps are extracted from a selected convolutional layer—typically the last convolutional layer before the fully connected layers. Simultaneously, the gradient of the output corresponding to the target class $y_c$ with respect to the activation maps $A_k$ is computed. Next, a Global Average Pooling operation is applied to these gradients to calculate the importance weights $\alpha_k^c$ for each feature map. These weights represent the relative significance of each channel in predicting the target class. Subsequently, a weighted linear combination of the activation maps and their corresponding weights is computed. Passing this result through a ReLU (Rectified Linear Unit) activation produces the Grad-CAM heatmap, which highlights the influential regions of the image in the model's decision. By preserving the spatial structure of the features. this method provides a visually explainable representation of the model's behavior and serves as an effective tool for explaining decision-making processes in deep learning models, particularly in the domain of computer vision [27].*

### *1.3.2.4 Grad-CAM++ Method*

Grad-CAM++ is an advanced version of the Grad-CAM (Gradient-weighted Class Activation Mapping) technique, developed to enhance the accuracy and explainability of saliency maps in convolutional neural network (CNN) models. Unlike the original Grad-CAM, which relies on simple averaging of gradients to weight feature maps, Grad-CAM++ employs a weighted combination of positive first-order gradients along with higher-order (second and third-order) gradients. This allows for a more precise estimation of the contribution of each spatial location in the convolutional layers with respect to a specific target class.

This refined weighting mechanism enables Grad-CAM++ to better identify multiple occurrences of the same object within an image and to provide more detailed and accurate visual explanations. The resulting saliency maps are capable of covering more complete regions and boundaries of objects, offering superior localization performance.

Notably, Grad-CAM++ is architecture-agnostic and can be applied to any CNN model without the need for model retraining, setting it apart from both Grad-CAM and the original Class Activation Mapping (CAM) method [28].

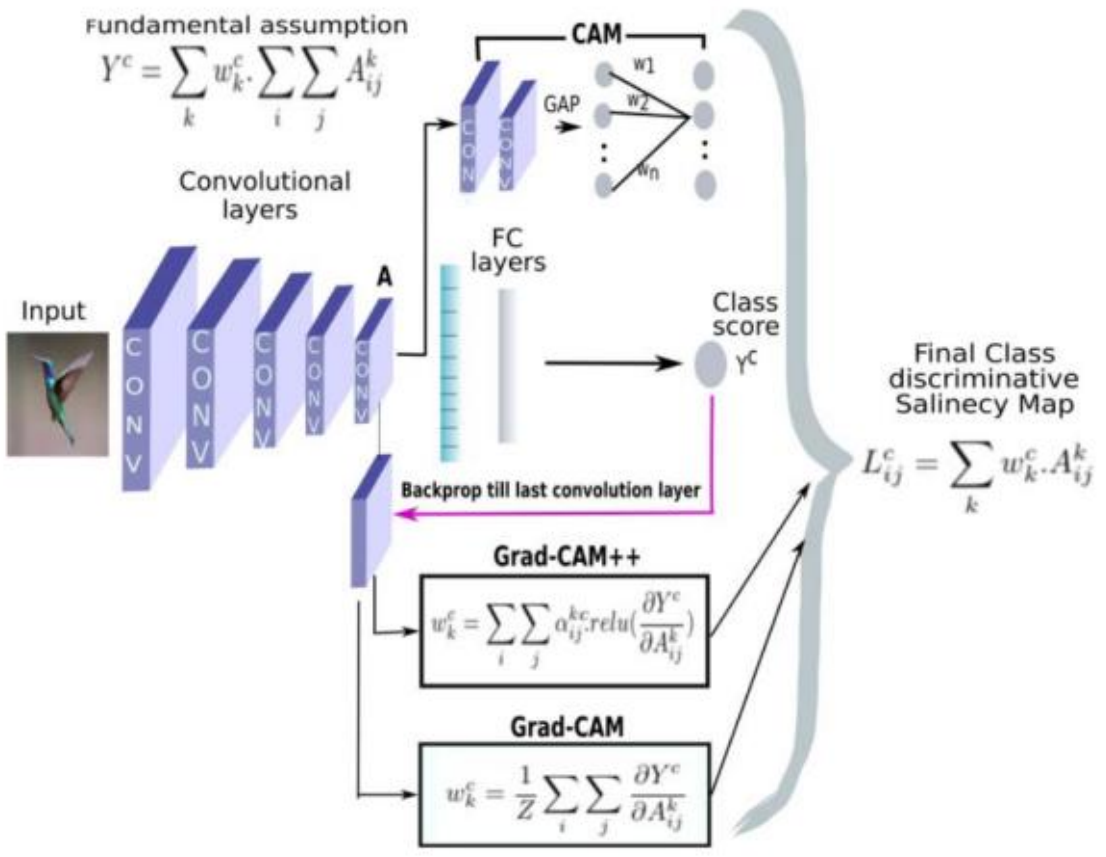

*Figure 18: Explainability Map Generation in Convolutional Neural Networks: This diagram illustrates the process of generating explainable maps for predicting a specific class $Y^c$ in convolutional neural networks (CNNs). The process begins by passing an input image through multiple convolutional (CONV) layers, which produces feature maps. In the Class Activation Mapping (CAM) method, feature maps are first reduced to vectors using Global Average Pooling (GAP). A linear layer is then employed for each class, learning weights $w_k^c$ to generate the class score $Y^c$. The final saliency map $L_{ij}^c$ is obtained through a linear combination of the feature maps and the learned weights. In Grad-CAM (Gradient-weighted Class Activation Mapping), instead of directly learning the weights, the gradients of the output score for class $Y^c$ with respect to the feature maps $A^k$are used. The weights $w_k^c$ are computed by averaging these gradients. This method is architecture-independent and can be applied to any CNN model without requiring structural modifications. The more advanced method, Grad-CAM++, replaces simple averaging with a weighted average of positive gradients. The corresponding coefficients $\alpha_{ij}^{kc}$αijkc are computed using second- and third-order derivatives. This enhancement facilitates more precise explainability and improves the identification of multiple instances of a given object class within the image. In all three methods, the final saliency map is formulated as: $L_{ij}^c = \sum_k w_k^c . A_{ij}^k$ This map serves as an explanation of the model's decision by highlighting the relative importance of each region in the image for class c [28].*

### 1.3.3 Backpropagation-Based Attribution Methods

This category of methods also employs the backpropagation process to assign importance to input features, albeit with certain modifications compared to raw gradients. Techniques such as Contrastive Relevance Propagation (CRP) and Layer-wise Relevance Propagation (LRP) fall within this group. These approaches aim to preserve the property of completeness, ensuring that the total sum of attribution values aligns with the difference between the model's actual output and the output for a baseline input.

One of the key advantages of these methods is their computational efficiency, as they do not require extensive masking or sampling. Nevertheless, the outputs of such methods can be noisy or unstable, particularly in variants that represent both positive and negative evidential contributions [29].

#### *1.3.3.1 LRP: Layer-wise Relevance Propagation*

The Layer-wise Relevance Propagation (LRP) method is designed to interpret the output of neural networks by recursively distributing the model's final prediction score from the output layer back to the input layers. This backward propagation aims to identify the contribution of each input feature to the model's decision-making process. The redistribution process follows specific conservation rules that ensure the relevance values are preserved across layers. These relevance scores are then visualized as a heatmap, highlighting the regions of the input—such as areas in an image—that had the greatest influence on the model's prediction.

In object detection models like YOLO (You Only Look Once), LRP can be integrated with the predicted bounding boxes to provide a detailed explanation of why certain objects were identified in specific locations within the image. This enhanced explainability facilitates a more thorough analysis of the model's performance and its potential errors [30].

#### *1.3.3.2 CRP: Contrastive Relevance Propagation*

The Contrastive Relevance Propagation (CRP) method is developed to enhance the explainability of outputs in object detection models by analyzing the prediction score in a contrastive manner relative to other classes. In contrast to Layer-wise Relevance Propagation (LRP), which attributes the overall contribution of input features to the model's prediction, CRP highlights the discriminative features of a specific class by computing the difference between the relevance assigned to that class and the average relevance of all other classes.

In networks such as SSD (Single Shot MultiBox Detector), which perform diverse predictions involving both class labels and object localization, CRP employs specialized propagation rules and a sign-based switching mechanism within the localization layers. This approach not only identifies the impact of each feature but also reveals the direction of influence—whether a feature contributes positively or negatively to the prediction.

As a result, CRP generates more precise and trustworthy heatmaps that improve model explainability and support more accurate error analysis [31].

### 1.3.4 Graph-Based Methods

This category of methods utilizes graph structures to model relationships among input components or extracted features. Due to their strong capability in representing complex structures, graphs enable the analysis of nonlinear interactions between image regions, output classes, or even internal layers of a neural network. Within this framework, *nodes* typically represent features, regions, or objects, while *edges* denote the degree of association or similarity among them.

Algorithms such as message passing, centrality metrics computation, and graph clustering are employed to identify salient and semantically significant regions within the image. Graph-based methods are particularly powerful in complex scenarios such as object interaction analysis, multimodal information integration, and the explanation of spatial-semantic relationships in visual data. These methods are capable of producing detailed and reliable explanations of model behavior.

In this study, we have classified graph-based approaches as a distinct category within the broader taxonomy of explainability methods.

Our research focuses on explainable AI models in the field of object detection; hence, the following chapter specifically reviews existing works and further studies conducted within this domain.

## 2 Explainability Methods in Object Detection

Although significant progress has been made in the field of explainability for deep learning models in recent years, our review reveals that existing studies focused on explaining object detection models remain limited, fragmented, and lacking in theoretical and empirical coherence.

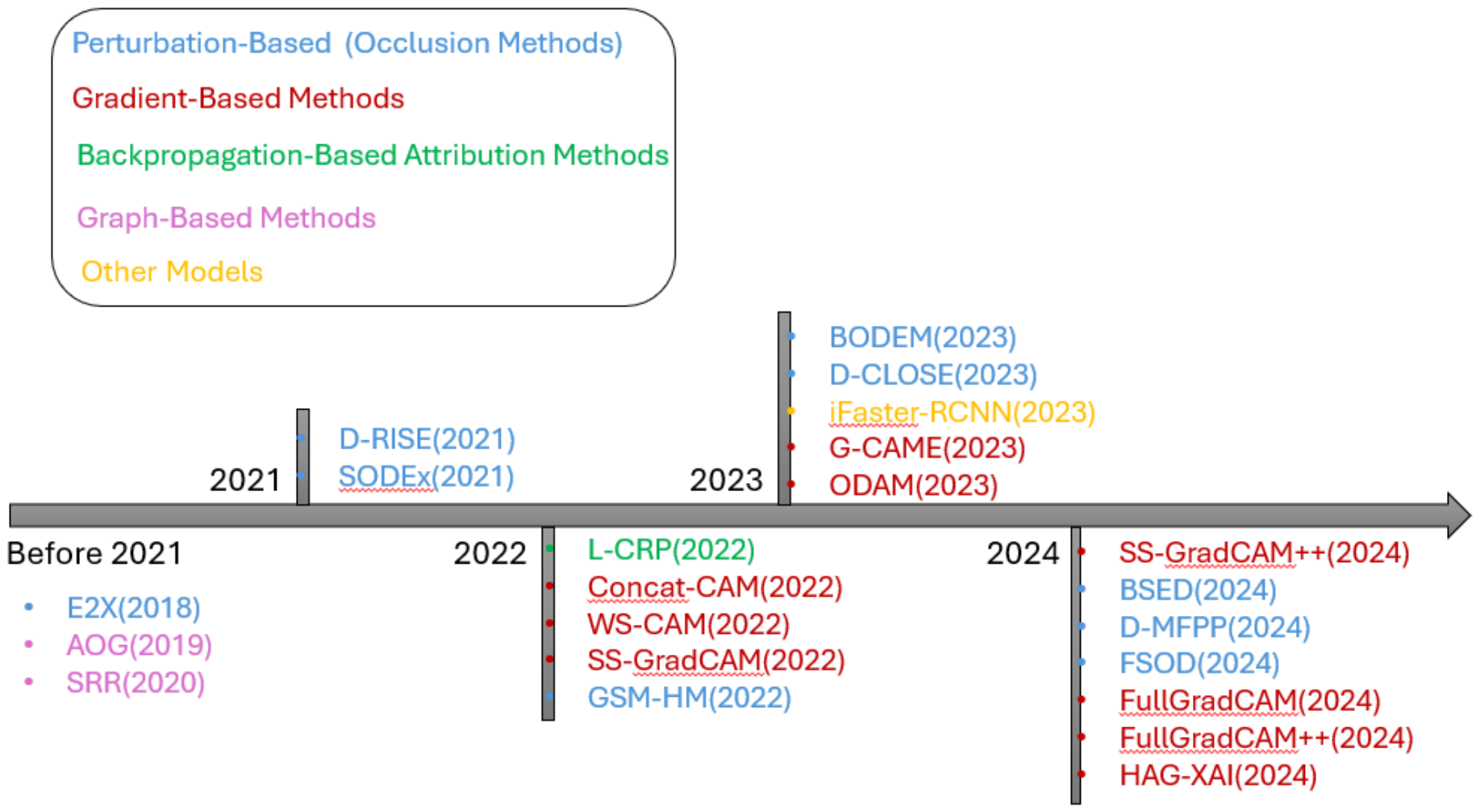

*Figure 19: Timeline of the development of explainability methods in object detection, categorized into perturbation-based, gradient-based, backpropagation-based attribution, graph-based, and other models, along with their respective publication years.*

A comprehensive examination of relevant academic literature indicates that—similar to the classification schemes established for image classification tasks—explainability methods in object detection can also be broadly categorized into four main groups:

1. Occlusion- or Perturbation-Based Methods
2. Gradient-Based Methods

3. Backpropagation-Based Methods
4. Graph-Based Methods (a newly emerging category)

These categories reflect the prevailing approaches adopted in current research, each offering distinct mechanisms for interpreting and visualizing the decision-making processes of object detection models.

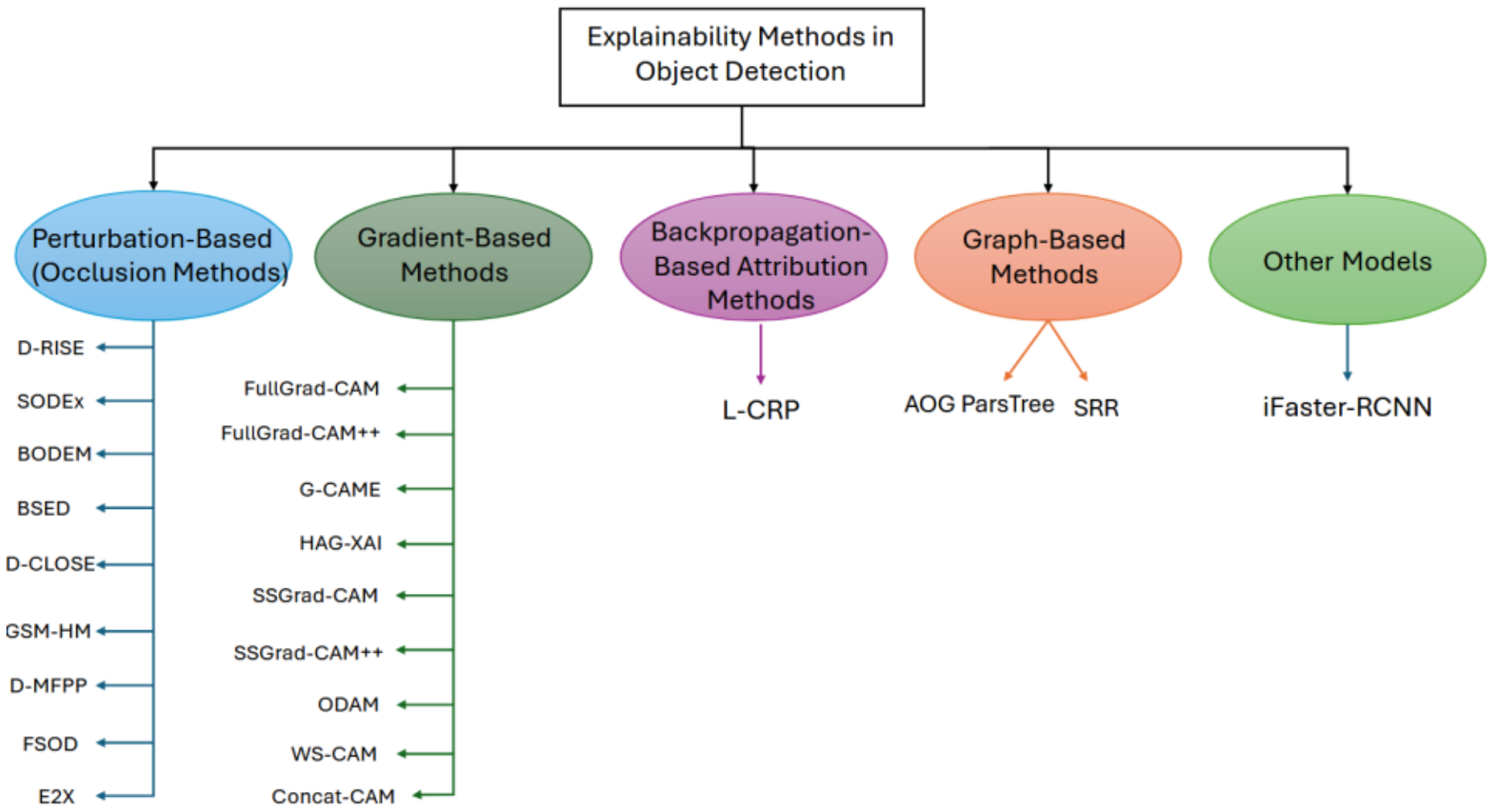

*Figure 20: Taxonomy of explainability methods in object detection, categorized into perturbation-based, gradient-based, backpropagation-based attribution, graph-based, and other approaches, along with representative techniques for each group.*

## 2.1 Perturbation-Based Methods

In perturbation-based methods, specific regions of the input image are intentionally altered or occluded in order to explain the decision-making process of object detection models. By analyzing how these modifications affect the model's predictions, researchers can infer which parts of the image are most influential in guiding the model's outputs.

### 2.1.1 D-RISE Method

The D-RISE method (short for Detector Randomized Input Sampling for Explanation) is a technique based on random sampling for explaining the behavior of object detection models. It operates without requiring knowledge of the internal structure of the model (i.e., it is a black-box approach). The goal of this method is to generate saliency maps that highlight which areas of an image have the most significant impact on the model's predictions regarding the location and class of objects.

To achieve this, a large number of random binary masks are first generated, which either blur or retain certain portions of the image randomly. These masks are then applied to the image, and the object detection model (such as YOLO or Faster R-CNN) is run on the altered versions of the image to obtain the corresponding outputs (object locations and classes).

For each masked image, the model's outputs are compared to the target detection (which can either come from the model itself or from ground truth data), and a similarity score is computed. This score is determined based on three key components: the Intersection over Union (IoU) of the bounding boxes, the cosine similarity of the class probability vectors, and, if available, the similarity in the objectness score. These similarity scores are then used as weights for the masks. By combining these weighted masks, the final saliency map is generated. This map clearly indicates which regions of the image the model relied on for detecting each object [32].

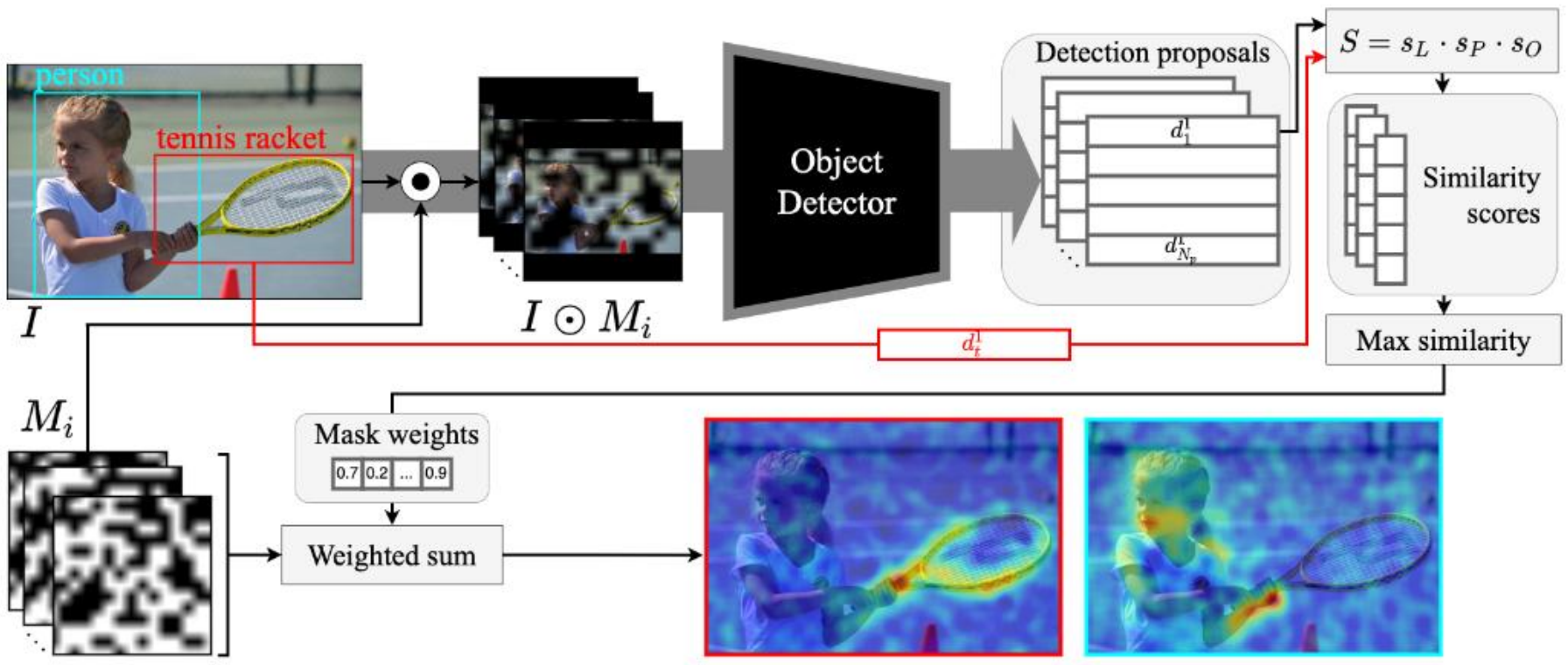

*Figure 21: Overview of the D-RISE Method for Explaining Object Detection Models: In this method, the input image I is multiplied by a set of random binary masks $M_i$ to generate incomplete masked images $I \odot M_i$ . Each masked image is then fed into the object detection model, producing a set of detection proposals $d_k$. For each masked image, the maximum similarity between its proposals and the target detection $d_t$ is calculated using a combination of three metrics: spatial overlap (IoU), cosine similarity of class vectors, and the objectness score. These similarity values are considered as weights for each mask in the final saliency combination. The final saliency map is obtained by summing the weighted masks, visually highlighting the areas of the image that had the most impact on the model's prediction [32].*

### 2.1.2 SODEx Method

The SODEx (Surrogate Object Detection Explainer) algorithm is a model-agnostic method for **explaining** the outputs of object detection models such as YOLOv4. Since object detection models (such as YOLO) directly predict multiple objects in an image with bounding box coordinates and confidence scores, rather than a fixed class, SODEx creates a surrogate binary classifier for each identified object. This surrogate classifier models the detection of the specific object in the image. The classifier is defined as a function that returns the model's confidence score if an object with the same class and in a similar position (based on the IoU metric) is present in the YOLO output; otherwise, it returns a score of zero. This enables the explanation and interpretation of each object independently.

To explain the output of this surrogate classifier, the LIME (Local Interpretable Model-agnostic Explanations) method is employed. LIME first divides the image into segments called superpixels and generates new images by either graying out or retaining these segments. By applying these modified images to the surrogate classifier, the influence of each superpixel on the confidence score is evaluated. The result is a simple linear model that highlights the importance of each segment of the image. This process allows us to understand which regions of the image (within or even outside the bounding box) the model relied on for its detection. In summary, SODEx combines YOLOv4 and LIME to provide intuitive and comprehensible explanations of why and how a model detects an object, without requiring access to the internal structure of the model [33].

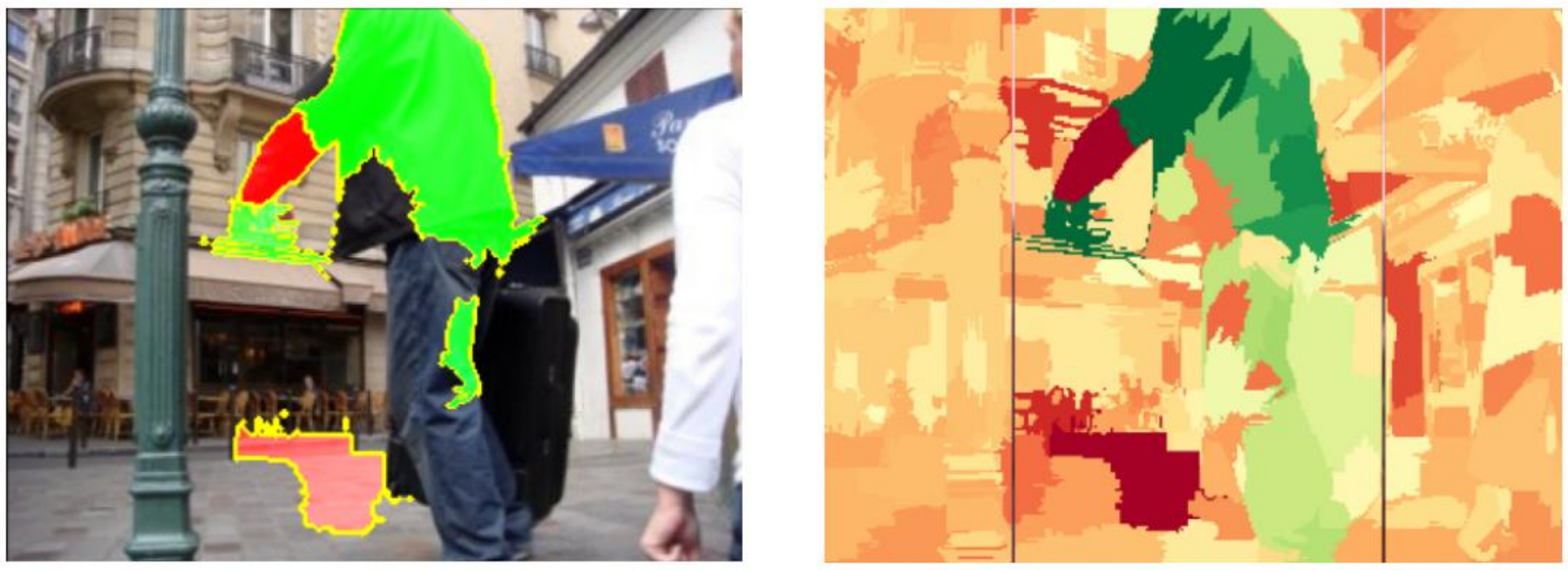

(**a**) Top-Bottom explanatory superpixels. (**b**) Superpixel explanation weights.

*Figure 22: Example of the SODEx Method Output for Explaining a Person's Detection by YOLOv4: In image (a), the green regions represent the superpixels that have had the most positive impact on the object detection, while the red regions have had the most negative influence. In image (b), the heatmap of the weights corresponding to the superpixels is shown, indicating the relative importance of each region in the model's decision-making process. In this example, with a YOLOv4 confidence score of 0.91, the model places the most emphasis on the upper body, visible hand, and legs of the person. In contrast, areas such as the forearm and around the legs may cause uncertainty in the model's detection [33].*

### 2.1.3 BODEM Method

The BODEM method (short for Black-box Object Detection Explanation by Masking) is a model-agnostic approach for explaining object detection models in a black-box manner. This method identifies the regions that influence the model's decision without requiring knowledge of the internal structure of the model or access to class scores and detection probabilities. It relies solely on the input image and the model's output (i.e., the coordinates of the detected bounding boxes).

The core mechanism of BODEM is based on hierarchical random masking, meaning that the input image is progressively masked at multiple levels, from large blocks to smaller regions. Initially, important areas are identified broadly using coarse masks, and then these areas are further refined and examined with finer masks at lower levels. At each stage, the model is applied to the masked image, and the new output is compared with the original output to measure the impact of the mask on object detection (using the Intersection over Union, or IOU, metric).

Ultimately, this process results in the generation of a saliency map for each detected object—a visual map that indicates the relative importance of each part of the image in the model's decision-making. A key advantage of BODEM over similar methods such as LIME and D-RISE is its higher accuracy and stability in generating saliency maps, as it focuses only on truly influential regions and prevents noise from being introduced into the maps. This feature makes BODEM particularly useful for sensitive applications such as software testing, medical imaging, or intelligent transportation systems, where it serves as a valuable tool for analyzing and validating the decisions made by object detection models [34].

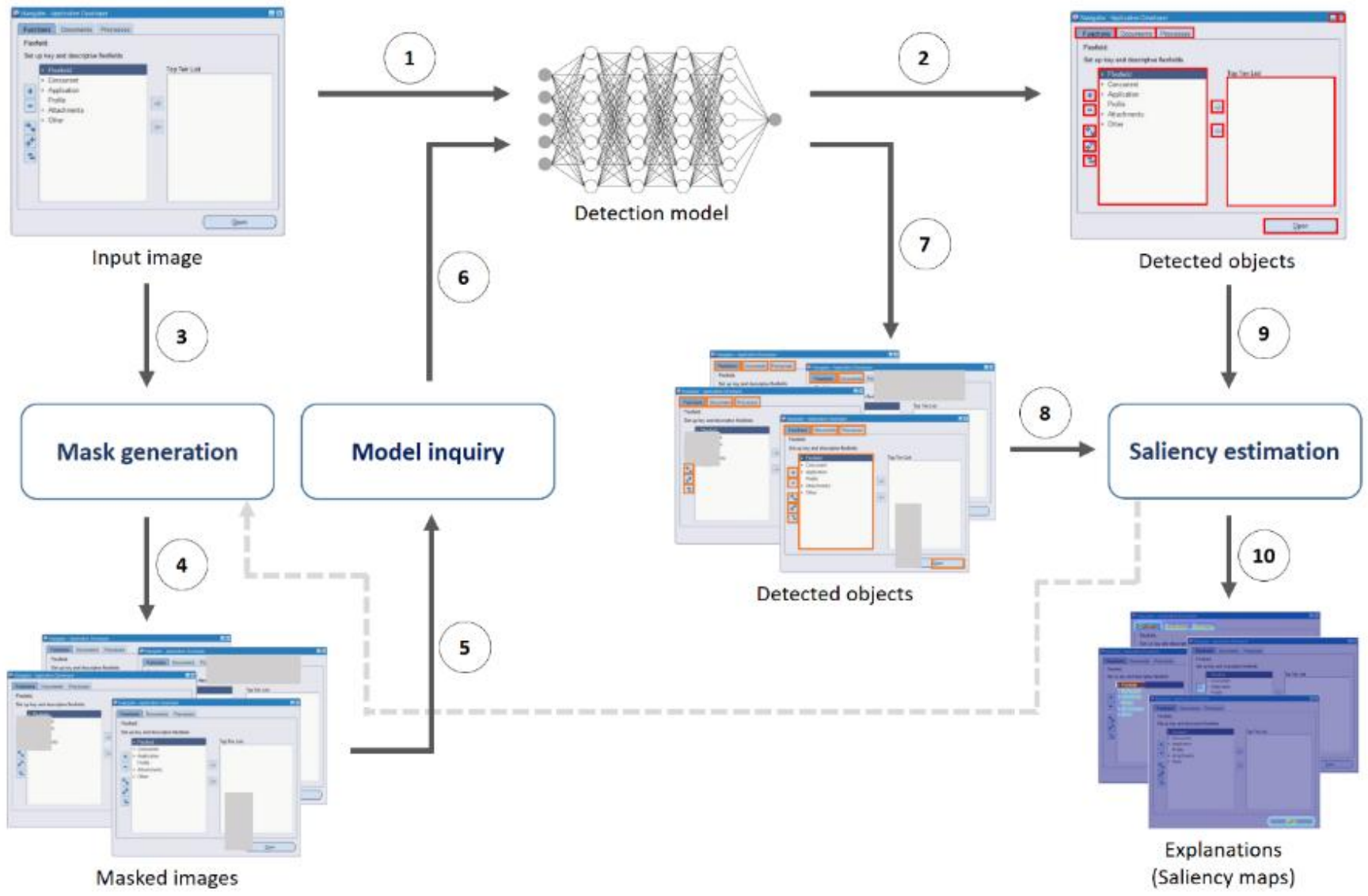

*Figure 23: General Architecture of the BODEM Method for Explaining Object Detection Models in Black-box Scenarios – The BODEM method generates explanations for object detection models in black-box settings through a sequence of ten steps: the input image is provided to the detection model, which predicts the objects and returns their bounding boxes. The mask generation module then receives the original image and produces masked versions of it using a hierarchical masking process. These masked images are sent to the model inquiry module and passed to the detection model, which generates new predictions for each masked image. The newly predicted bounding boxes, along with the original bounding boxes, are sent to the saliency estimation module. This module produces the final explanations in the form of heatmaps, highlighting the relative importance of image regions in the model's decisions, while also guiding the mask generation module during the iterative hierarchical masking process to refine the accuracy of the explanations[34].*

### 2.1.4 BSED Method

The BSED method (Baseline Shapley-based Explainable Detector) is an innovative technique in the field of explainability for object detection models, designed to provide accurate, reliable, and model-agnostic explanations. BSED is based on the mathematical concept of "Shapley value," which is used in game theory to assess the contribution of each participant to the final outcome. In this context, each pixel in the image is considered as a participant, and BSED aims to determine the positive or negative role of each pixel in the model's decision-making by calculating the impact of the presence or absence of each pixel on the model's output. The BSED method is model-agnostic (i.e., it does not require knowledge of the internal architecture of the model) and can be applied to any object detection model.

To implement this approach, BSED randomly generates masks on the input image, where parts of the image are hidden as patches. Then, by examining the changes in the model's output in response to these masks and using a multilayer approximation of the Shapley value, a map of the contribution of each region of the image is created. This map is displayed as a heatmap, with areas that are crucial for accurate detection highlighted in positive colors and misleading regions shown in negative colors. This feature makes BSED not only capable of identifying key regions of the image but also able to provide in-depth analysis of the reasons behind the model's errors or successes [35].

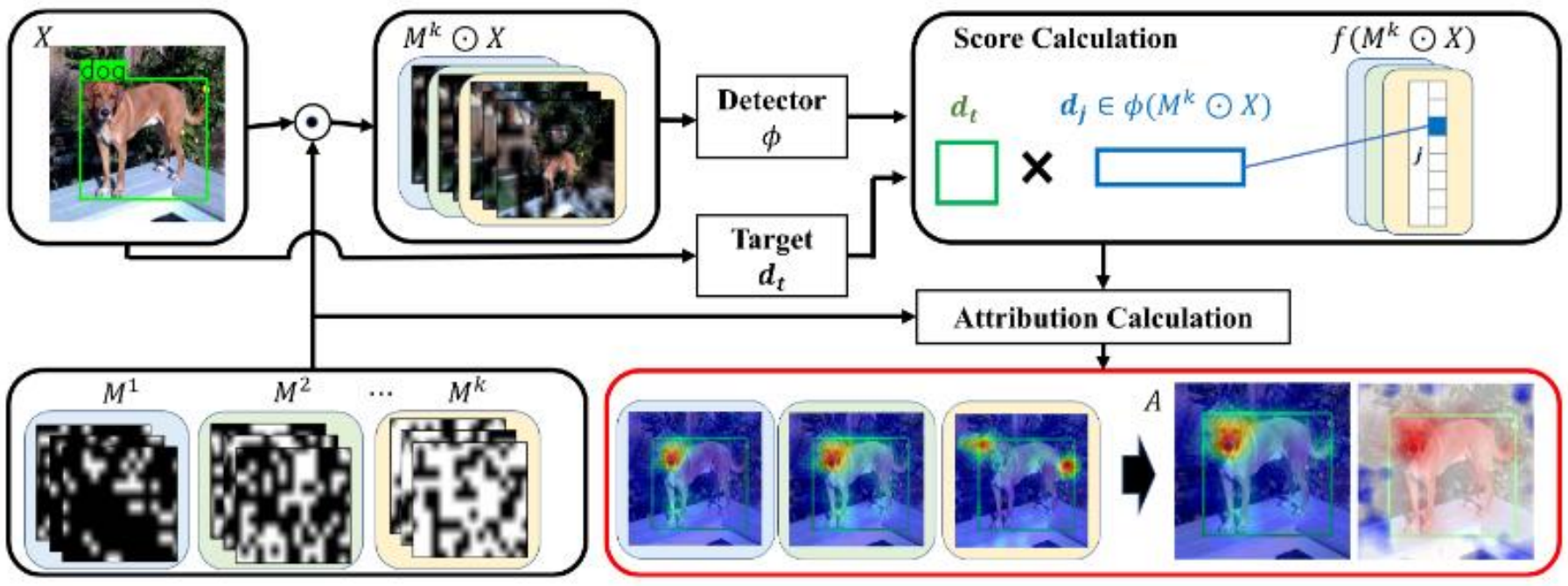

*Figure 24: Overview of the BSED Method for Feature Map Generation in Object Detection Models: The BSED method for generating feature maps in object detection models is outlined as follows. Initially, a target detection (such as an identified object) is selected from the input image. Using binary masks, different versions of the image are generated and fed into the model. The model's output for each masked image is then compared with the target detection, and the similarity (including location and class) is calculated. Subsequently, the contribution of each pixel to the model's decision is computed using a Shapley value approximation, ultimately producing a precise and quantitative map of the positive and negative importance of various regions within the image [35].*

### 2.1.5 D-CLOSE Method

The D-CLOSE method, which stands for Detector-Cascading multiple Levels of Segments to Explain, is a model-agnostic approach (i.e., independent of the model's architecture) designed to explain the decisions made by object detection models. In this method, the input image is first divided into smaller segments (super pixels) at multiple levels. These levels capture various details of the image, ranging from fine-grained to more general regions. Using these segments, random masks are generated, and different versions of the image are created by applying these masks. These masked versions are then fed into the model, and the model's output for each version is analyzed. D-CLOSE determines the contribution of each region to the model's decision by calculating the similarity score between the output of each masked image and the model's original prediction (considering bounding box coordinates, objectness score, and class probability).

Next, a concept called Density Map is employed to normalize the frequency of each pixel in the masks, reducing noise and enhancing the transparency of the explanation.

Finally, attention maps generated at different segmentation levels are hierarchically combined to create a final map that shows exactly which regions of the image the model focused on when detecting each object. The main advantages of D-CLOSE over methods like D-RISE include its ability to provide more precise explanations for objects of varying sizes and positions, reduce noise, eliminate the need for manual adjustments for each object, and offer higher computational efficiency. This method also demonstrates stable performance under challenging lighting or imaging conditions [36].

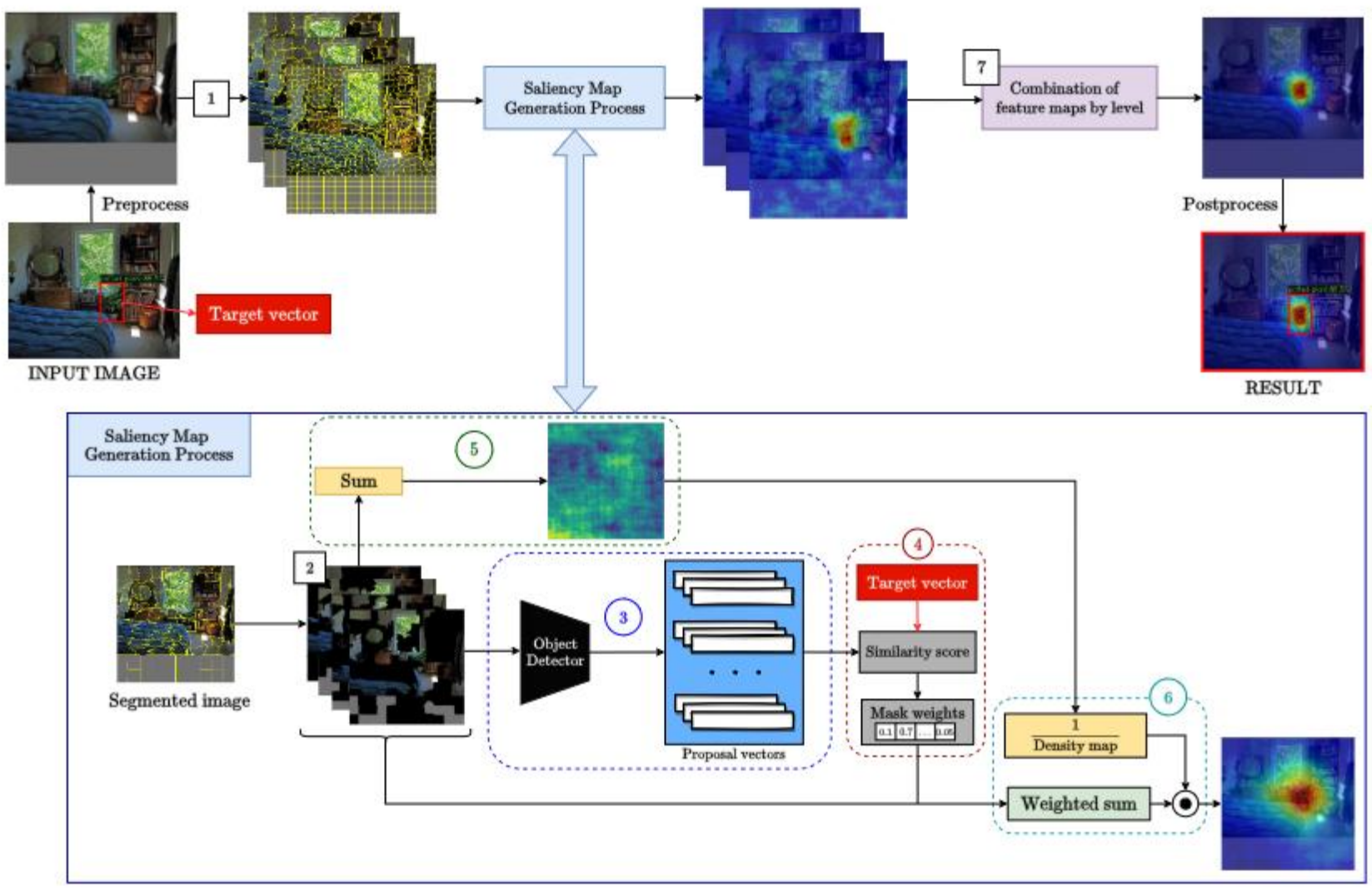

*Figure 25: Overview of the D-CLOSE Process: In the upper section, the overall structure of the algorithm is depicted, which includes the segmentation of the image into different levels of superpixels, the generation of random masks, passing the masks through the object detection model, and calculating similarity scores for each prediction. In the lower section, the process of combining saliency maps derived from various levels to create a final explainable map is illustrated. This approach, utilizing multi-scale fusion, provides higher quality and accuracy in explaining the model's predictions [36].*

### 2.1.6 GSM-HM Method

The GSM-HM method (Generating Saliency Maps for Hierarchical Masking-based Black-box Object Detection Models) is an innovative explanatory framework for object detection models that identifies important regions of an image without requiring access to the internal structure of the model. In this method, the image is divided hierarchically into cells of various sizes at multiple levels. Initially, at higher (coarser) levels, approximate regions that are likely important for the model are identified using random yet guided masking. Then, at lower (finer) levels, these regions are examined and refined more precisely. This process is conducted using the "nearest neighbor selection" algorithm, which generates masks that focus more on the important regions. At each step, the saliency score for each region is calculated by comparing the model's output for the original image and the masked image. This score is gradually updated at lower levels.

The advantage of GSM-HM over previous methods, such as D-RISE, is that it leverages the hierarchical structure to prevent the application of random, aimless masks, thereby reducing "noise" in the final map. This method can identify object-related regions with high accuracy and greater clarity, while irrelevant areas are either discarded or deemed of lesser importance [20].

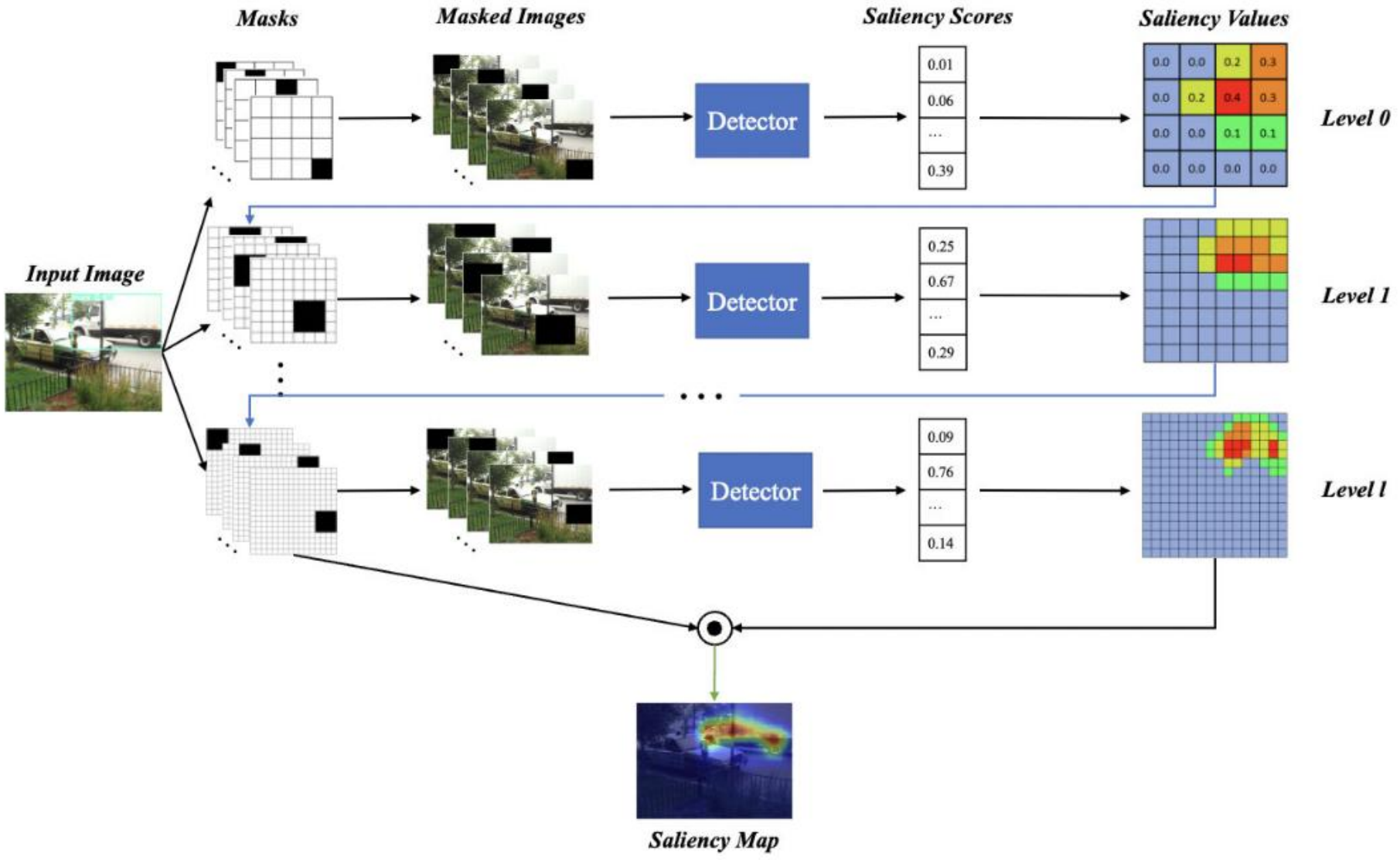

*Figure 26: Overview of the GSM-HM Hierarchical Framework for Generating Saliency Maps in Object Detection Models: Initially, masks are applied at a coarse level based on large cells to identify approximate salient regions. As the level increases and the cell sizes decrease, the salient regions are extracted and refined more precisely. Ultimately, a map is generated with a high focus on important regions while effectively removing irrelevant noise [20].*

### 2.1.7 D-MFPP Method

The D-MFPP (Developmental Morphological Fragmental Perturbation Pyramid) method is an explainability technique for black-box object detection models, independent of the model architecture. It is specifically designed to explain the outputs of complex models such as YOLOv8. Unlike methods like RISE, which use entirely random masks, D-MFPP employs masks based on image segmentation at multiple scales. This means that the input image is initially divided into meaningful segments (superpixels) of varying sizes. These segments are then randomly turned off or removed, and the model's output is evaluated for different versions of the image. The goal of this process is to examine the impact of removing each segment on the model's final prediction, helping to identify which regions of the image play the most significant role in object detection.

To calculate the importance of each region, D-MFPP utilizes a similarity score proposed in the D-RISE method. This score not only considers the similarity of the predicted class but also incorporates the overlap of the region (using Intersection over Union, or IoU) and the objectness score. With this approach, D-MFPP is capable of generating more accurate saliency maps that better explain both the position and meaning of each object in the image. Key advantages of this method include higher precision in detecting important regions and reduced noise in explanations, particularly in cases where the number of masks is limited or multiple objects of the same class are present in the image [37].

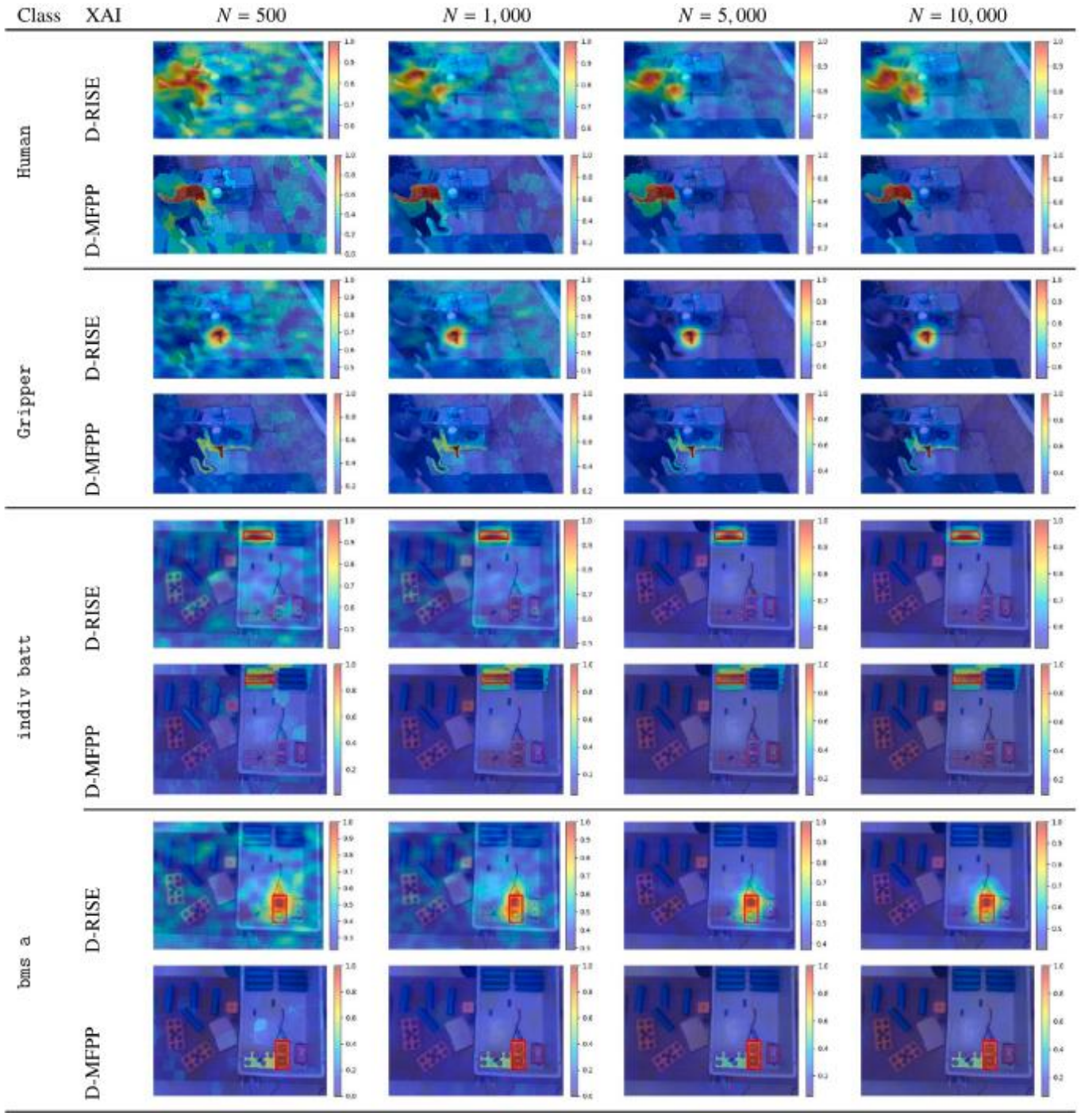

*Figure 27: An example of heatmaps generated by the D-RISE and D-MFPP methods with different numbers of masks, illustrating the impact of the number of masks on the quality and focus of the explanations. In scenarios where the number of masks is low, D-MFPP has been more effective at distinguishing important regions and reducing irrelevant noise. Conversely, as the number of masks increases, D-RISE performs better in preserving finer details [37].*

### 2.1.8 FSOD Method

The FSOD method (Fast Shapley Explanation for Object Detection) is designed to provide accurate, object-dependent, and fast explanations for object detection models. In traditional Shapley-based methods, high-precision saliency maps require heavy and time-consuming computations. To address this challenge, FSOD utilizes a learner model called the *explainer*, which is trained to estimate the Shapley value of each pixel for every object in the image. Once trained, this model can generate precise saliency maps in real-time during the testing phase, without the need for sampling or heavy computation.

The key difference between FSOD and previous methods is that it is specifically designed for object detection tasks, rather than simply image classification. In object detection, the object type and its spatial location (bounding box) are both important. To account for this, FSOD architecture uses two key inputs: first, a feature map extracted from the internal layers of the object detection model, which contains the spatial-conceptual information of the image; second, a map called the query map, which specifies the spatial location of the target object. These two inputs assist the *explainer* in producing precise, region-specific explanations for each object [9].

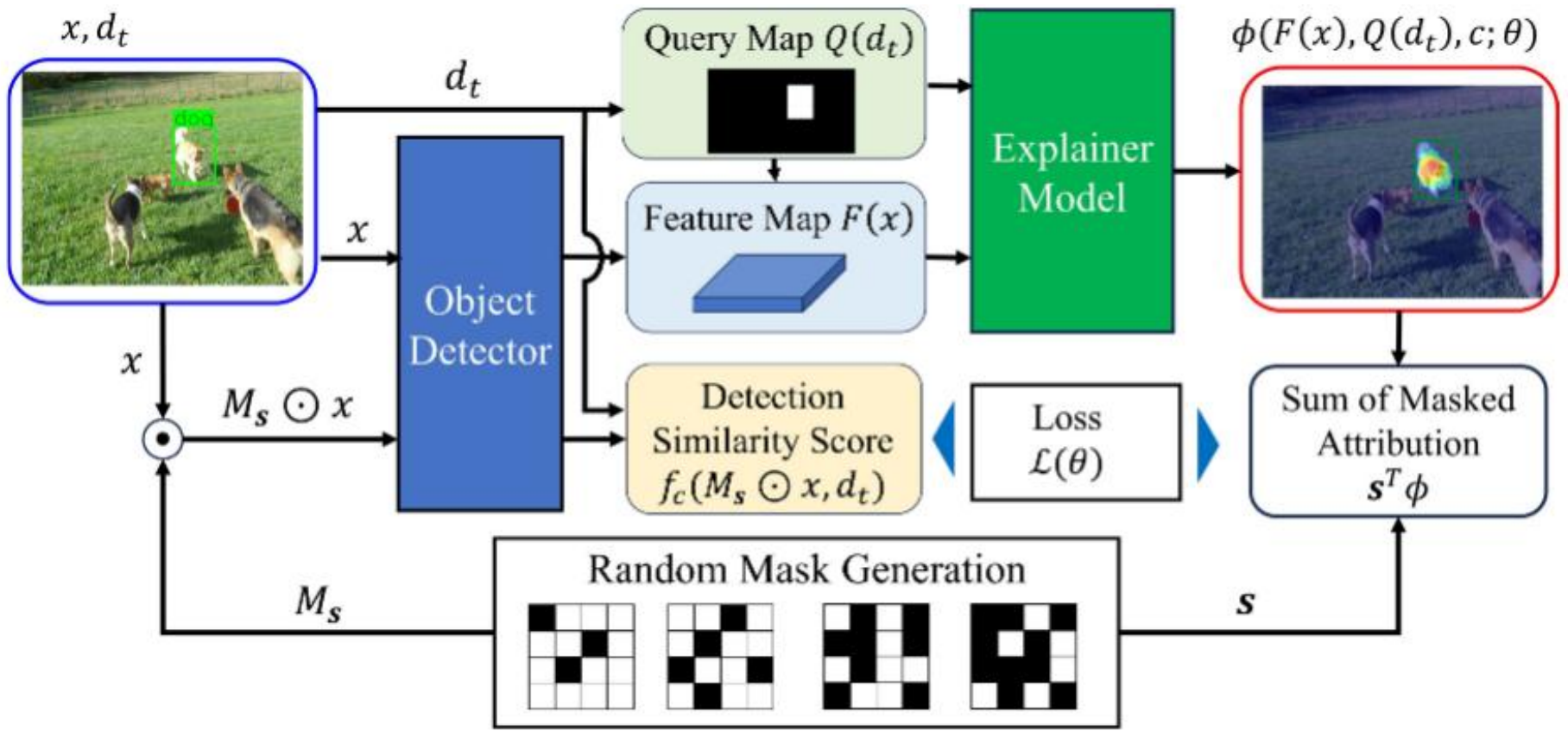

*Figure 28: Overview of the FSOD Framework as an Explainable Method for Object Detection: In this approach, an input image x and a target detection $d_t$ are fed into the model. The explainer model, utilizing the image feature map F(x), the query map extracted from the target detection Q(dt)‹, and the object class information, generates a spatial importance map or attribution map $\phi$, which highlights the regions influencing the object detection model's decision-making. To train the explainer model, random masks $M_s$ are applied to the image. A loss function L(θ) is then defined by measuring the similarity between the model's output on the masked image and the target detection. The goal of this loss function is to align the attribution map with the responses of the object detection model [9].*

#### 2.1.9 E2X Method

The E2X (Explain to Fix) method is an explainability framework for object detection models designed to identify and correct prediction errors. This method is based on the SHAP (Shapley Additive Explanations) theory and uses an optimized version of the Integrated Gradients (IG) method to calculate the importance of each region of the image (features). In the first step, the image is divided into a set of superpixels to transition from a pixel-level representation to a more semantically meaningful one. Then, the transition path from the reference image (e.g., a grayscale image) to the actual image is sampled with random weights, so that by maintaining the assumption of feature independence, the importance of each superpixel can be calculated more accurately and with less noise. Unlike traditional methods, this process does not require any changes to the structure or parameters of the neural network and is executed independently.

One of the key advantages of E2X is that, in addition to high accuracy in identifying effective features (both positive and negative), it is also very time-efficient (up to 10 times faster than methods such as LIME) and can be easily run on GPUs. This makes it suitable for large-scale analyses on datasets, such as images from surveillance cameras or autonomous vehicles. More importantly, E2X does not just generate attention maps; it also identifies lost or corrupted features that caused the error, enabling the correction of the model through retraining with improved data [38].

### 2.1.10 Summary of Perturbation-based Methods for Explaining Object Detection Models

In this category of methods, the impact of each section on the model's decision is evaluated by applying targeted modifications (perturbations) to regions of the image. These methods are primarily model-agnostic and are widely used for evaluating and analyzing model performance in real-world and complex conditions.

*Table 1: Summary of the methods mentioned in this review in the category of occlusion*

| Method Name | Main Idea | Advantages | Key Features |
|---|---|---|---|
| **D-RISE** | Applying random masks and measuring their effect on the model's output | Precise explanation of effective regions in prediction | Use of IoU (Intersection over Union) and class similarity to compute saliency |
| **SODEx** | Constructing an alternative classifier and using LIME to explain each object | Separate analysis of each object, understandable, no need for internal architecture | Combination of YOLO and LIME; displays the impact of each superpixel |
| **BODEM** | Hierarchical masking for precise analysis of key regions | High stability, noise reduction, suitable for sensitive applications | Multi-layer examination of masks from coarse to fine |
| **BSED** | Computing Shapley values for pixels via random masking | Model-agnostic, providing precise positive/negative maps | Use of Shapley value approximation; analysis of pixel roles |
| **D-CLOSE** | Combining multi-scale superpixels and weighting based on output similarity | Accurate explanations for objects with different scales and positions | Noise reduction, high speed, no need for manual adjustments |
| **GSM-HM** | Hierarchical guided masking with selection of important neighbors | Increased focus, noise elimination, improved saliency map accuracy | Smart selection of masks at different levels |
| **D-MFPP** | Removing meaningful parts of the image at multiple scales and analyzing output changes | High accuracy, suitable for complex or multi-object images | Utilization of superpixels and D-RISE scoring |
| **FSOD** | Training an explainer model to quickly estimate Shapley values | Real-time explanation, specific to object detection, no sampling required | Combination of feature map + query map for each object |
| **E2X** | Using an optimized version of IG + SHAP for | Fast, accurate, no need for model structure changes | Can be executed on GPUs, suitable for model retraining |
| | model analysis and improvement | | |

## 2.2 Gradient-based Methods

In this category, by calculating the gradient of the model's output with respect to the input, the regions of the image that have the greatest influence on the object prediction are identified and highlighted.

### 2.2.1 FullGrad-CAM Method

The FullGrad-CAM method was introduced as an extension of the Grad-CAM method to address its limitations when applied to object detection models. While Grad-CAM was primarily designed for image classification models and generates class-specific saliency maps, FullGrad-CAM overcomes this by eliminating the spatial averaging operation on the gradients and directly using an element-wise combination of the gradients and activation maps. This allows for the generation of object-specific saliency maps.

Since object detection models may contain multiple objects with different classes in an image, FullGrad-CAM facilitates better model explainability by calculating separate saliency maps for each object and normalizing them individually. This method, without introducing any trainable parameters, enables its application to more complex architectures such as Yolo and Faster-RCNN simply by modifying the calculation structure. A key feature of this model is its ability to preserve more accurate spatial information in the gradients, thus improving the spatial precision of the saliency maps [39].

### 2.2.2 FullGrad-CAM++ Method

The FullGrad-CAM++ method is an extension of FullGrad-CAM that utilizes the ReLU function on the gradients to focus on more positive and meaningful features. The application of ReLU not only reduces noise but also enhances the focus of saliency maps on more influential areas, especially in scenes with dense or overlapping objects.

By combining positive gradients and activation maps, and without employing global pooling, FullGrad-CAM++ offers greater spatial accuracy compared to previous methods. This characteristic makes it particularly effective in real-world environments, such as road images in autonomous driving scenarios, where the model's attention regions align well with the objects present. A prominent feature of this model is the generation of high-resolution explanatory maps, better object distinguishability, and greater alignment with human-expected regions [39].

### 2.2.3 G-CAME Method

The G-CAME (Gaussian-Class Activation Mapping Explainer) method is an advanced technique for explaining the decisions of object detection models. Built on the Grad-CAM framework, it incorporates key innovations that overcome the limitations of previous methods. Unlike Grad-CAM, which is only suitable for classification models, G-CAME is designed for object detection models and can identify the effective region in an image for each predicted bounding box.

This method first utilizes gradients to pinpoint the precise location of the target object in the feature map. Then, using gradient-based weighting, it calculates the importance of each channel in the feature map. Subsequently, to focus more on the object's core region, a Gaussian mask is applied around the center of that region.

By combining gradient weighting and spatial filtering through a normal distribution (Gaussian), G-CAME generates an accurate, smooth, and low-noise map of the region relevant to the model's decision. This ensures that only the area related to the target object is highlighted in the image. G-CAME is applicable not only to single-stage detection models, such as YOLOX, but also to two-stage models like Faster-RCNN.

Key advantages of G-CAME over similar methods include its high speed, reduced noise in saliency maps, and precise identification of the target object's location without interference from nearby objects. These features make it an efficient tool for explaining the decisions of deep learning models in computer vision [40].

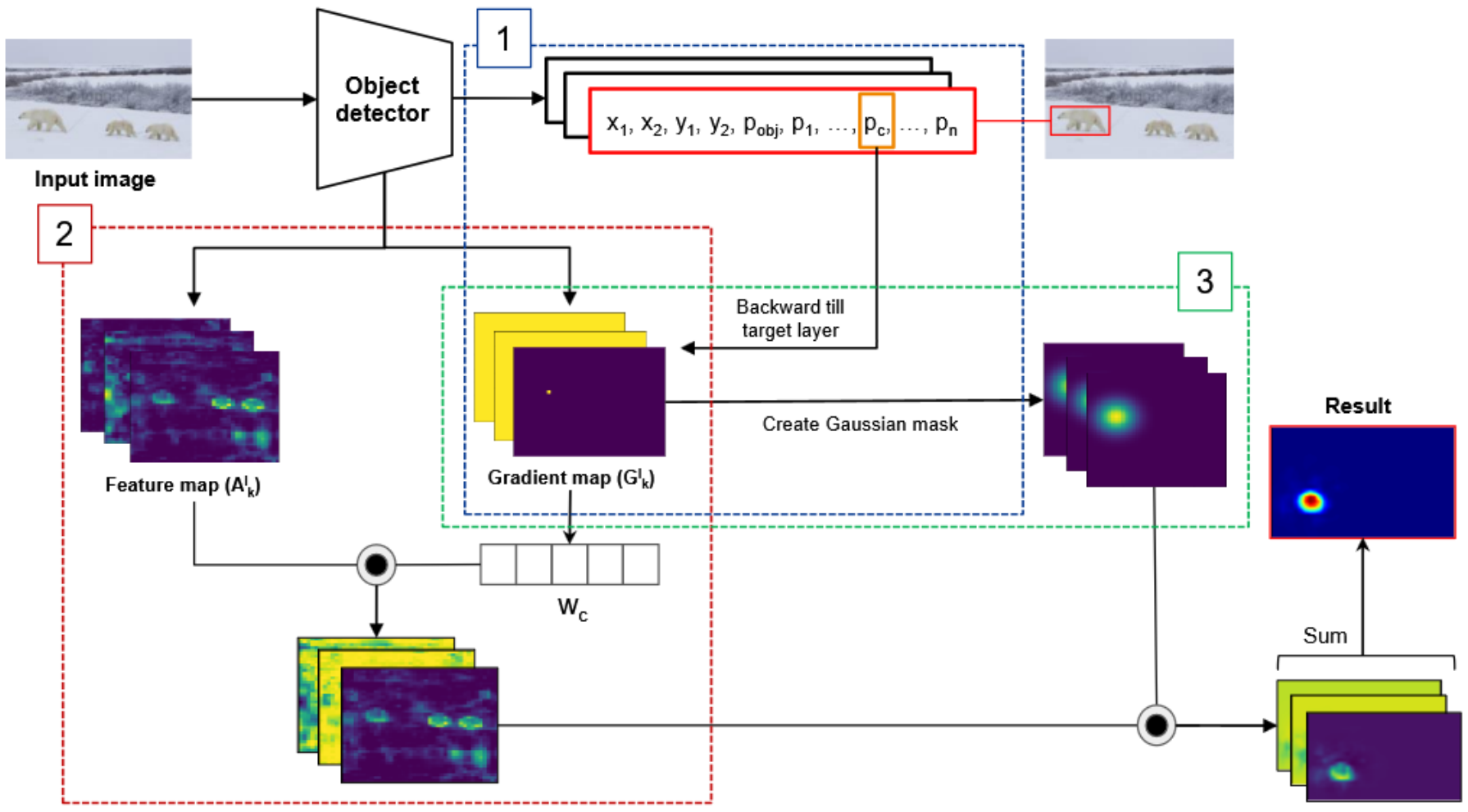

*Figure 28: Step-by-Step Process of G-CAME: First, the input image (e.g., containing several polar bears) is fed into an object detection model (such as YOLOX or Faster-RCNN). The model generates the predicted bounding box coordinates ($x_1$, $y_1$, $x_2$, $y_2$), the object existence score ($p_{obj}$), and the probability for each class ($p_1$, ..., $p_n$). Feature maps are then extracted from a target layer, and the gradients corresponding to the prediction are calculated to determine the influence of each region on the model's decision. Next, using the predicted bounding box coordinates, a Gaussian mask is created and applied to the feature maps. This serves to amplify the relevant regions while suppressing the irrelevant ones. Finally, the weighted feature maps are combined, producing a clear saliency map focused on the target object [40].*

### 2.2.4 HAG-XAI Method

The HAG-XAI method (Human Attention-based Explainability) is an innovative explainability framework for computer vision models, developed with the goal of generating more explainable and accurate saliency maps. Unlike conventional XAI methods that operate solely based on gradients or input variations, HAG-XAI is trained using real human attention data (e.g., eye movement maps when viewing an image) to learn an optimal combination of activation maps and gradients. This combination is achieved through learnable activation functions (which distinguish the positivity and negativity of the signals) and Gaussian smoothing kernels, ensuring that the final saliency map aligns as closely as possible with the human attention map in terms of smoothness and focus.

The training process of HAG-XAI is designed to maximize the similarity between the generated maps and the human attention maps (using metrics such as Pearson's correlation coefficient and RMSE error). In contrast to many deep learning models, all the parameters of this system are explainable, allowing us to identify which features contribute to the enhanced similarity and accuracy [39].

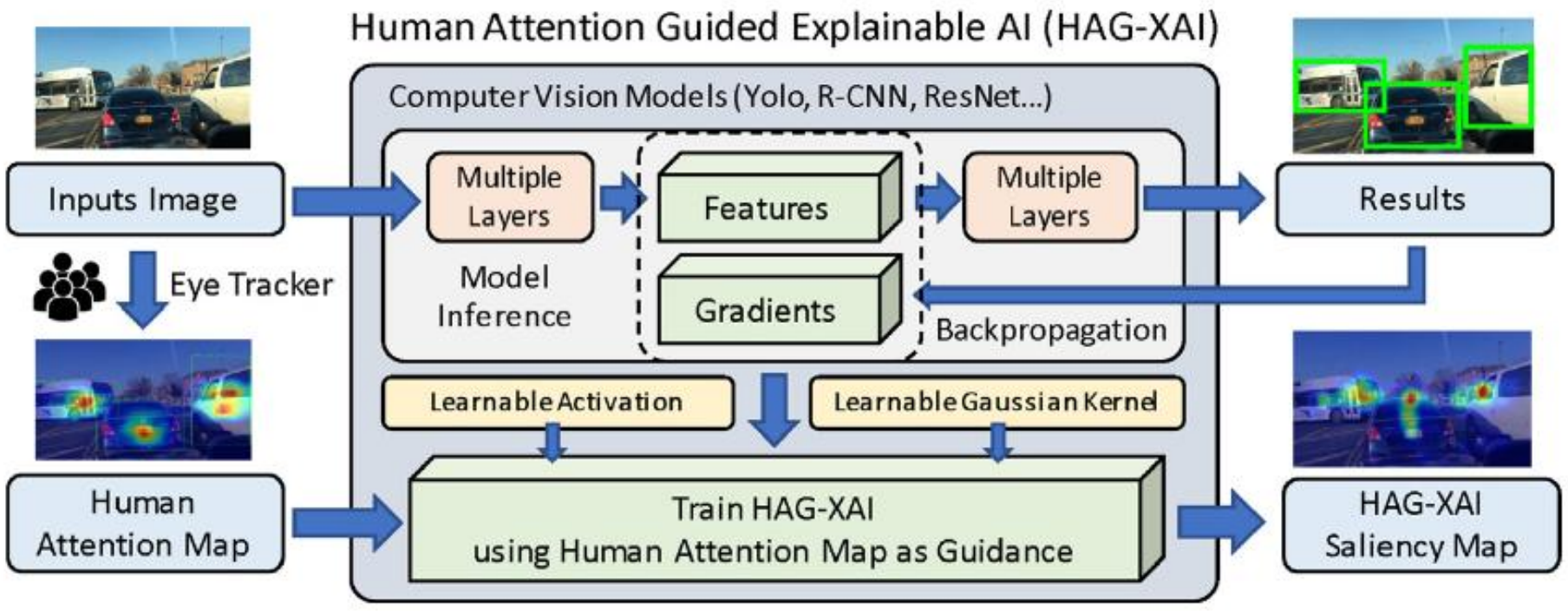

*Figure ۳۰: HAG-XAI Methodology: In the HAG-XAI method, an input image is first presented to the machine vision models. Simultaneously, human attention data (such as the eye movement path tracked using eye-tracking devices) for the same image is collected. The vision model is then executed on the image, and feature maps are extracted from the intermediate layers, while gradients are computed using backpropagation to measure the model's sensitivity to the features. Subsequently, two types of operations are performed on the feature maps and gradients: learnable activation functions are applied to recalibrate the importance of features, and smoothing is performed using learnable Gaussian kernels to align the structure of the maps with human attention patterns. The processed maps are then compared to the human attention maps, and the HAG-XAI model is trained so that the generated saliency maps have the highest possible similarity to human attention while maintaining fidelity to the original model's decisions. The final output of this process consists of two components: the model results, such as object detection boxes or classification predictions, and the HAG-XAI saliency map, which represents the regions of focus that the model places on the image in a human-understandable manner [39].*

### 2.2.5 SS-GradCAM Method

SS-GradCAM (Spatial Sensitive Grad-CAM) is a visual explanation method for object detection models designed to address the primary limitation of Grad-CAM. While Grad-CAM focuses on feature importance, it lacks spatial sensitivity. As a result, in many cases, irrelevant regions not related to the detected object are also highlighted in its heatmaps. To overcome this limitation, SS-GradCAM incorporates spatial sensitivity into the process, in addition to feature importance. This is achieved by calculating spatial maps, which are derived by normalizing the absolute gradients with respect to the feature maps. These spatial maps indicate which areas of the image have the most significant impact on the model’s decision, considering the spatial location of each pixel.

In SS-GradCAM, a separate heatmap is generated for each detected object. This heatmap is created by element-wise multiplication of the extracted weights (α) and feature maps, followed by a combination with the spatial maps and the final application of the ReLU function. Unlike Grad-CAM, this method is capable of highlighting only the region corresponding to the specific object, even when similar objects are present in the image, or the model has used surrounding areas for decision-making. Since this method does not require changes to the network architecture, it can easily be applied to models such as SSD or SSD+FPN, resulting in more accurate, focused, and reliable heatmaps that enhance the explainability of the model's behavior [41].

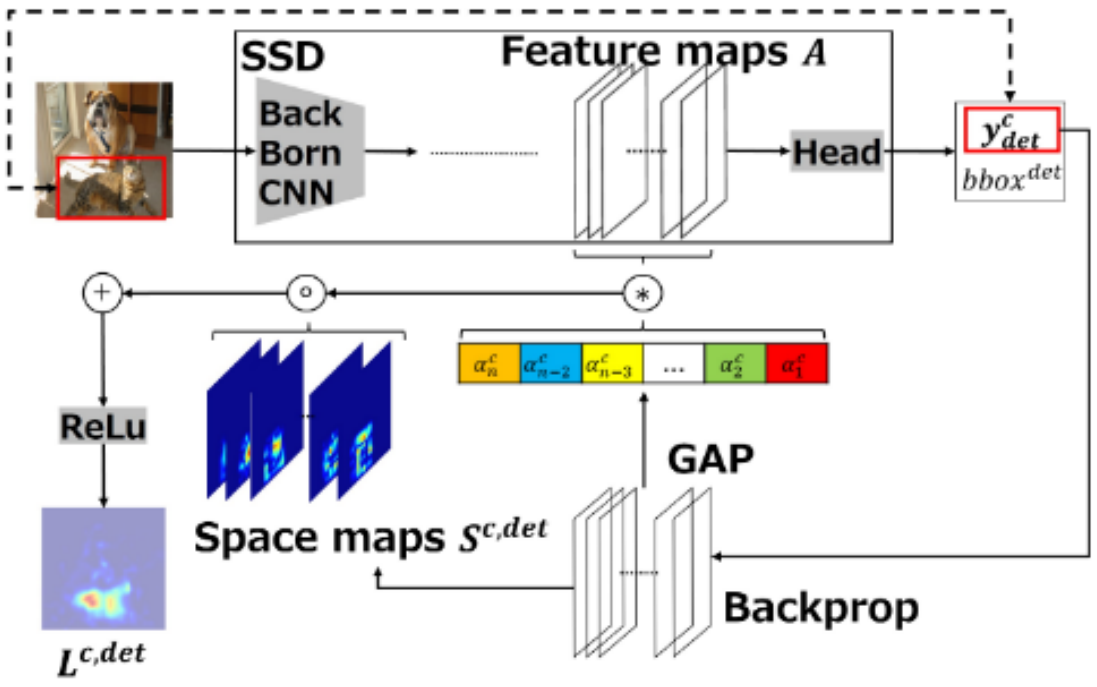

*Figure 29: Schematic Representation of the SSGrad-CAM Architecture: SSGrad-CAM (Spatial Sensitive Grad-CAM) is a method developed to enhance the spatial explainability of object detection models. This method first follows the standard Grad-CAM process, where the feature importance weight $\alpha^c_k$ is extracted using the gradient of the class score $y^c_{det}$ with respect to the feature map $A^k$ at the intermediate layers of the network. These weights are then multiplied by the corresponding feature map $A^k$, generating the initial attention map. In the next step, a spatial map $S^{c,det}$ is computed, which includes the normalization of the absolute gradients at the pixel level; this map reflects the model’s spatial sensitivity to different regions of the image. The product of Grad-CAM is then combined with this spatial map, and after passing through the ReLU function, the final spatial attention map $L^{c,det}$ is obtained. This map provides more precise information about the locations the model relied on to make predictions for the target object. The main advantage of SSGrad-CAM over traditional Grad-CAM is its improved spatial accuracy and better focus on the actual object regions in the image, without requiring any changes to the underlying model architecture [41].*

### 2.2.6 SS-GradCAM++ Method

The SS-GradCAM++ method is an enhanced version of SS-Grad-CAM, designed to produce more accurate visual explanations in object detection models. In the original

SS-Grad-CAM approach, the importance of each feature (feature map) is determined using a simple averaging of gradients with respect to the model's output. This importance is then combined with a spatial map—reflecting the influence of different regions in the image—to generate the final heatmap. However, due to its reliance on mere averaging, SS-Grad-CAM may highlight only a subset of truly relevant regions, potentially overlooking critical information necessary for explaining the model's behavior.

To address this limitation, SS-GradCAM++ improves the feature importance computation by drawing inspiration from Grad-CAM++. Instead of relying on simple averaging, it employs a weighted combination of gradients. These weights (denoted as α) are calculated using the second- and third-order derivatives of the model's output with respect to the features, allowing for more precise attribution of importance at the pixel level. These computed weights are then integrated with a spatial map—calculated similarly to SS-Grad-CAM—to generate the final heatmap.

The outcome of this process is heatmaps with enhanced spatial resolution and improved ability to comprehensively highlight regions relevant to the identified object. This significantly contributes to a deeper explanation of how object detection models operate [42].

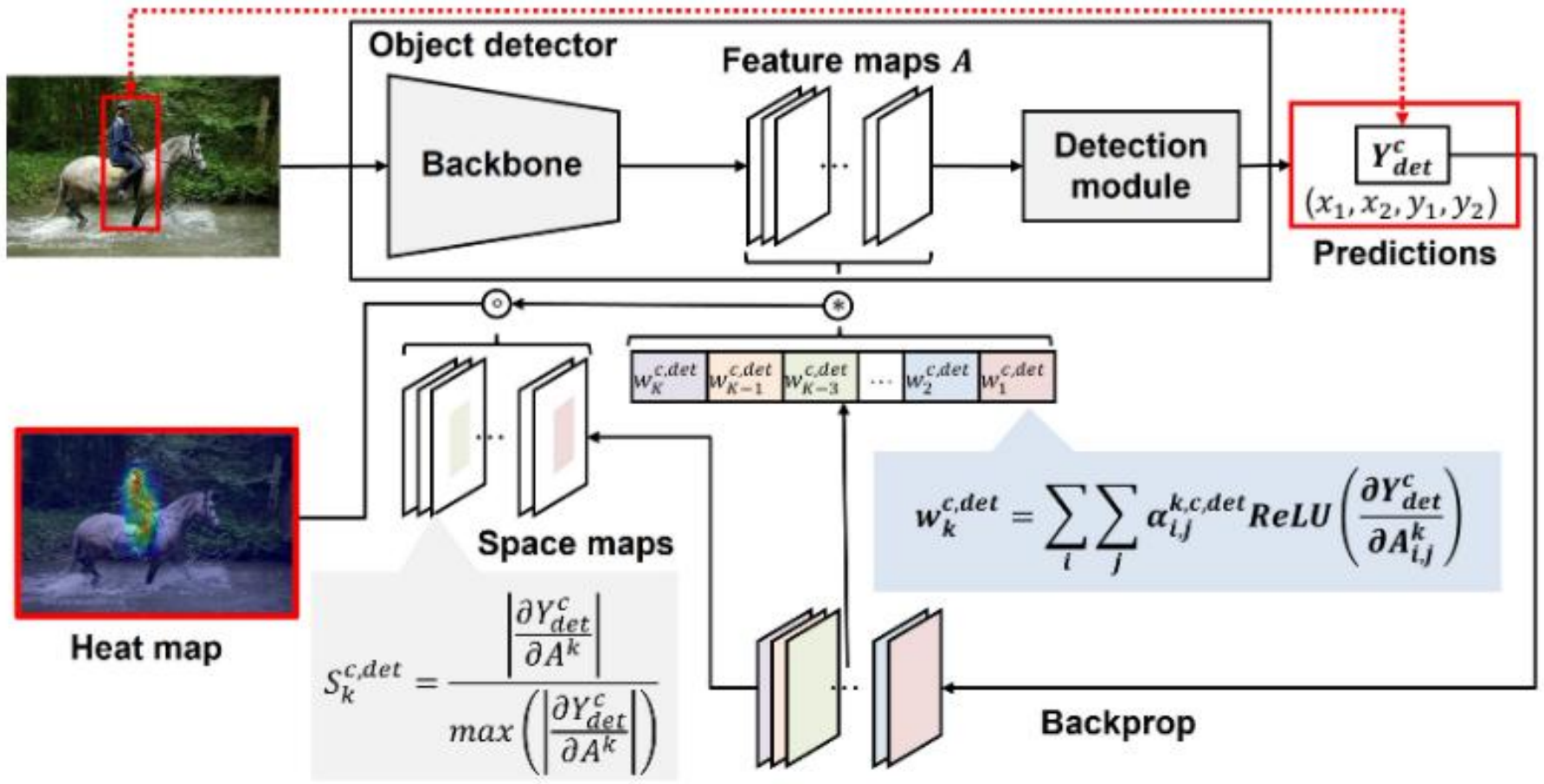

*Figure 30: Detection Model Architecture and Heatmap Generation*
*The object detection model comprises two main components: a feature extraction module (Backbone) and a detection module, which together produce the final output. For each detected object, two key components are computed: 1. Feature Weights $w_k^{c,det}$: These weights represent the importance of the k-th channel in the feature maps and are obtained using a weighted combination of gradients, inspired by Grad-CAM++. This allows for the channel-wise attribution of significance based on the model's response to a specific object class c .2. Spatial Sensitivity Map $S_k^{c,det}$: This map highlights the importance of each spatial region within a given channel. It is computed through gradient normalization, following a procedure similar to that of SSGrad-CAM, thereby capturing spatially localized relevance. Finally, the resulting heatmap is generated by combining these two components—feature importance and spatial relevance—providing an object-specific and spatially precise explanation of the model's focus. By leveraging both gradients and higher-order derivatives, this approach offers enhanced explainability and spatial accuracy compared to earlier methods such as Grad-CAM and SSGrad-CAM [42].*

### 2.2.7 ODAM Method

The ODAM (Object Detector Activation Maps) method is a novel and effective technique for generating visual explanatory heatmaps in object detection models. In contrast to classical approaches such as Grad-CAM, which produce class-based heatmaps and tend to highlight all objects of the same category in a similar manner, ODAM generates a distinct and individualized heatmap for each detected object within an image.

These heatmaps are derived by computing the gradients of the model's outputs with respect to intermediate feature maps in the neural network. Through a process of local and uniform aggregation—facilitated by a smoothing operation such as a Gaussian filter—ODAM identifies the precise regions of the image that contributed to the model's decision for that specific object. This property enables ODAM to provide more precise and object-specific explanations of the model's internal decision-making processes.

To enhance the accuracy and explainability of the generated heatmaps, an auxiliary training strategy termed ODAM-Train is introduced. This training approach incorporates two types of loss functions:

- a consistency loss, which encourages coherence within the heatmaps corresponding to a single object, and
- a separation loss, which promotes greater distinction between heatmaps of different objects.

This dual-loss framework significantly improves the quality of visual explanations, particularly in cluttered or crowded scenes.

Building on ODAM's outputs, the authors further propose the ODAM-NMS algorithm for the Non-Maximum Suppression (NMS) phase. Unlike traditional NMS, which relies solely on the overlap between bounding boxes (Intersection over Union, IoU), ODAM-NMS also incorporates the similarity of heatmaps. This additional criterion helps prevent the erroneous suppression of distinct objects in dense scenes.

The integration of these three components—object-specific heatmap generation, enhanced training via ODAM-Train, and improved post-processing through ODAM-NMS—positions ODAM as a fast, accurate, and reliable method for improving both the performance and explainability of object detection models [43].

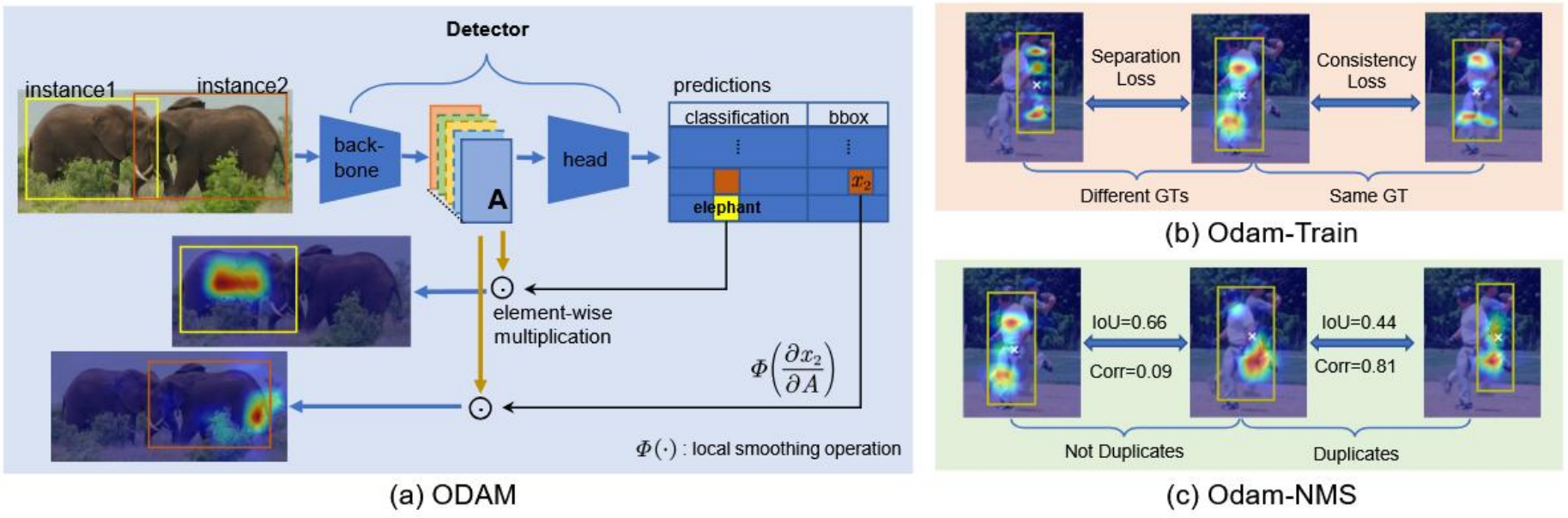

*Figure 31: Proposed ODAM Framework for Explainability in Object Detection – (a) The ODAM module generates instance-specific heatmaps by leveraging smoothed gradients with respect to the network's intermediate feature maps. These heatmaps distinctly highlight the regions that contributed to the prediction of each individual object. (b) In the ODAM-Train module, two auxiliary loss functions are defined: a consistency loss, which reinforces the similarity*

*among heatmaps corresponding to different predictions of the same object, and a separation loss, which enhances the differentiation of heatmaps associated with predictions of different objects. This training process leads to improved spatial separation in the explanatory heatmaps. (c) In the ODAM-NMS component, for the purpose of eliminating redundant predictions, the Intersection over Union (IoU) of bounding boxes is combined with the correlation between heatmaps. This approach ensures that predictions with overlapping boxes but distinct heatmaps are retained, while predictions with highly similar heatmaps—despite having slightly different boxes—are considered duplicates and removed[43].*

### 2.2.8 WS-Grad Method

The WS-Grad (Weighted Sum of Gradients) method is an innovative and hybrid technique in the field of Explainable Artificial Intelligence (XAI), introduced in the referenced article to enhance the explainability of object detection model decisions—particularly in the context of small object recognition. The method aims to generate more accurate heatmaps by integrating three prominent gradient-based techniques:

**Saliency Map**: Identifies influential regions in the image by computing the derivative of the output with respect to the input pixels.

**Guided Backpropagation (GBP)**: Filters gradients during backpropagation to allow only positive and significant features to pass through, thereby emphasizing relevant structures.

**Integrated Gradients (IG)**: Estimates the contribution of each pixel by integrating gradients along the path from a baseline image to the actual input image, providing a continuous measure of importance.

In the WS-Grad method, these three heatmaps are first computed independently and then combined using a weighted summation. The weights can be fixed or tuned based on specific requirements. This linear combination yields a final heatmap that offers both high spatial resolution and enhanced informational richness, surpassing what each individual technique can provide in isolation.

The key advantage of WS-Grad lies in its ability to leverage the complementary strengths of its constituent methods. For instance, it integrates the high edge precision of GBP, the broad pixel-level sensitivity of IG, and the overall coverage of Saliency Maps into a single, coherent output. This composite heatmap enables both users and developers to gain a clearer understanding of which regions the model focused on and the rationale behind specific decisions. Such explainability is particularly critical in applications involving the detection of small objects with limited visual detail [44].

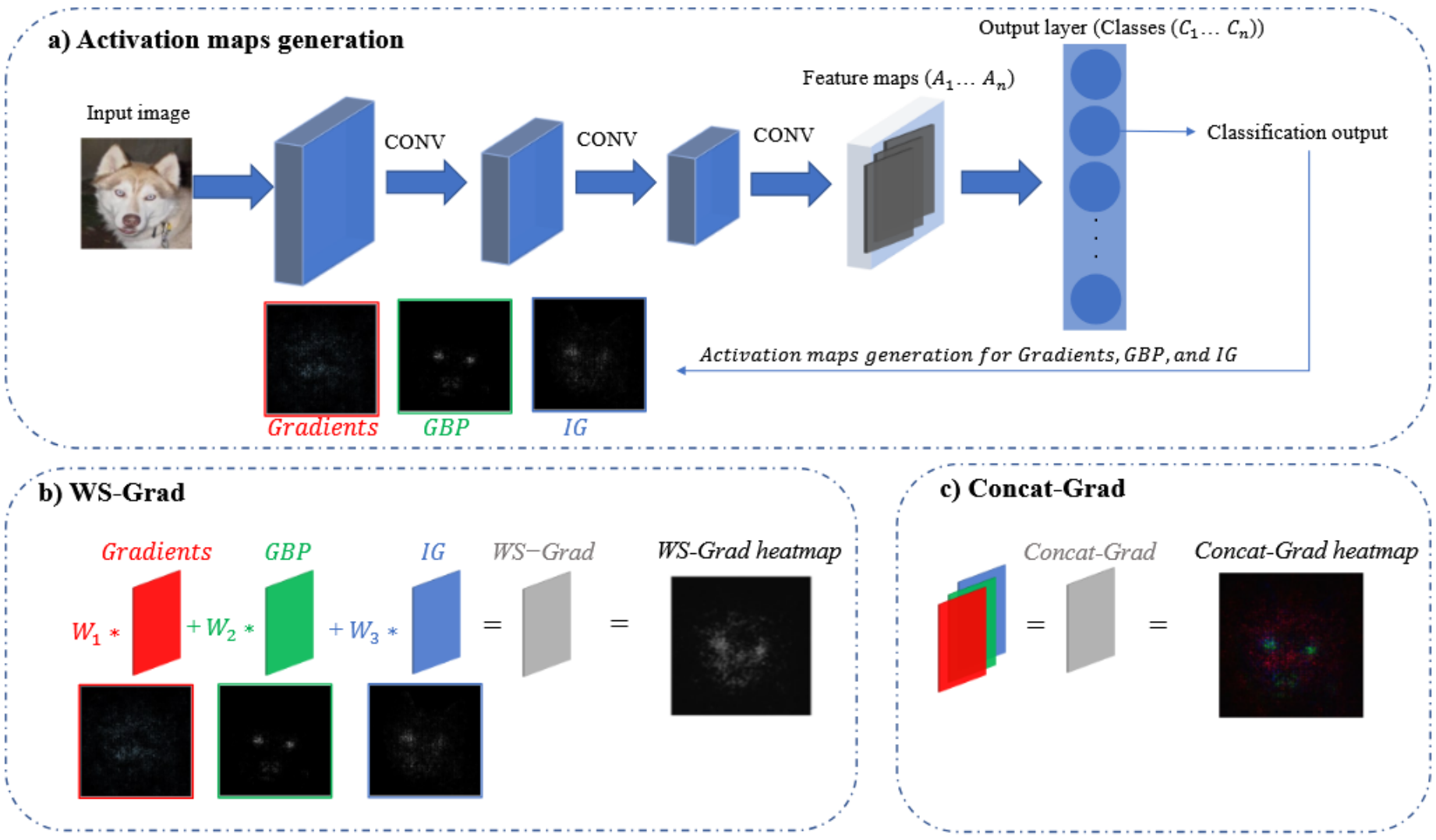

*Figure 32: Visualization and Comparison of Explanation Methods for Deep Learning Models in Image Classification: This figure presents three approaches for enhancing the explainability of deep learning models in the context of image classification: Figure (a) illustrates the process of generating activation maps. This involves feature extraction through convolutional layers, followed by the application of three explanation techniques: Gradients, Guided Bac propagation (GBP), and Integrated Gradients (IG). Figure (b) introduces the WS-Grad method, in which activation maps generated by the three aforementioned techniques are combined using different weights to produce a final heatmap. Figure (c) depicts the Concat-Grad method, where the activation maps are merged via direct concatenation, resulting in a unified heatmap. Overall, this figure offers a conceptual and comparative overview of the processes and outcomes of the two proposed explanation methods for computer vision models alongside baseline techniques. It effectively conveys the conceptual framework of XAI (Explainable Artificial Intelligence) methods and is particularly well-suited for illustrating their underlying principles [44].*

### 2.2.9 Concat-Grad Method

The Concat-Grad method is an approach developed based on the integration of three widely used gradient-based explanation techniques: Simple Gradients (Saliency Map), Guided Backpropagation, and Integrated Gradients. Rather than performing numerical combination or weighted aggregation of the resulting maps, Concat-Grad assigns each technique's output to a distinct color channel in the final image: the red channel is allocated to Saliency Maps, the green channel to Guided Backpropagation, and the blue channel to Integrated Gradients. This arrangement produces a high-resolution, color-coded heatmap that simultaneously displays the contribution and influence of each individual method across different regions of the image.

The key feature that distinguishes Concat-Grad from other explanation methods is its ability not only to highlight important regions of the image as perceived by the model, but also to indicate which technique identified each region's significance. In other words, this method provides a visually intuitive representation of why and how each part of the image influenced the model's prediction, thereby enhancing explainability in a multi-faceted and interpretable manner [44].

The architecture of the Concat-Grad method was illustrated in the preceding section.

#### 2.2.10 Summary of Gradient-Based Methods for Explaining Object Detection Models

The following table provides an overview of gradient-based methods employed in the field of explainability for object detection models. These methods identify the regions that influence the model's decisions by analyzing the gradients of the model's output with respect to internal features or the input image. The results are typically visualized as high-resolution heatmaps, enabling precise spatial localization of the influential regions.

*Table 2: Summary of the methods mentioned in this review in the gradient-based category*

| Method Name | Core Idea | Advantages | Key Features |
|---|---|---|---|
| **FullGrad-CAM** | Direct combination of gradients and activation maps without averaging | Preserves fine-grained spatial information; suitable for multi-object scenarios | An enhanced Grad-CAM that generates object-specific saliency maps for object detection without spatial gradient pooling. |
| **FullGrad-CAM++** | Incorporation of ReLU to enhance positive regions in gradients | Reduced noise; sharper focus; improved spatial precision | An extension of Grad-CAM++ that retains spatial detail and applies ReLU for clearer visualization of multiple objects. |
| **G-CAME** | Application of Gaussian masks on features for precise localization | Produces soft and accurate saliency maps; effective object separation | Combines activation maps with Gaussian kernels to generate precise, low-noise saliency maps for both one- and two-stage object detectors. |
| **HAG-XAI** | Aligning saliency maps with human attention via learning | More human-aligned interpretations; interpretable parameters; high fidelity | Human-Attention-Guided XAI using learnable activation functions and smoothing kernels to generate human-aligned saliency maps. |
| **SSGrad-CAM** | Combining spatial sensitivity with Grad-CAM | Improved spatial accuracy; focuses on specific objects | An improved Grad-CAM with spatial sensitivity to generate more precise heatmaps focused on detected object regions. |
| **SSGrad-CAM++** | Use of higher-order derivatives to enhance SSGrad-CAM | Produces more detailed maps with improved emphasis | Accurately computes feature importance via weighted gradient combination and spatial sensitivity for object detection. |
| **ODAM** | Generates separate heatmaps per object + learns discriminative loss | High accuracy; prevents object overlap; ODAM-NMS for duplicate removal | compatible with all detectors, faster and cleaner than perturbation methods, enhanced object discrimination and NMS via auxiliary training. |
| **WS-Grad** | Weighted combination of Saliency, GBP, and Integrated Gradients | Leverages strengths of each method; effective for small objects | Combines multiple gradient-based maps via weighted sum to produce high-resolution, expressive explanations. |
| **Concat-Grad** | Combines three gradient-based methods using RGB channels | Simultaneously and distinctly visualizes source of importance for each region | Concatenates saliency, GBP, and IG maps into RGB channels to visualize diverse features in a single heatmap. |

## 2.3 Backpropagation-Based Methods

This category of methods leverages the backpropagation mechanism—originally employed during neural network training to update model weights—to explain model decisions. In these approaches, the model's prediction score (e.g., the probability of belonging to a particular class) is propagated backward through the layers of the neural network. The objective of this process is to determine the contribution of each region or pixel in the input image to the model's final output. These contributions are typically visualized as heatmaps or saliency maps, which highlight the areas of the image that most significantly influenced the model's decision, thereby enhancing the explainability of the model's behavior.

### 2.3.1 L-CRP Method

The L-CRP method (short for Localized Concept Relevance Propagation) is an advanced and comprehensive technique in the field of explainable artificial intelligence (XAI). It has been developed to enhance the explainability of complex deep learning models, particularly in object detection and image segmentation tasks. Unlike conventional approaches—such as Grad-CAM or LRP—that merely highlight important image regions via heatmaps, L-CRP adopts a hybrid "glocal" strategy (i.e., combining global conceptual and local spatial reasoning) to identify not only which concepts (e.g., giraffe skin, ocean waves, or fence bars) played a pivotal role in the model's prediction, but also where in the image these concepts appeared and how significantly they contributed to the final output.

In this method, the importance of latent features within intermediate layers of the model is first computed using the Layer-wise Relevance Propagation (LRP) technique. Then, by imposing constraints on these features—such as selecting a specific filter or channel—only the relevance associated with a particular concept is propagated backward through the network. The result of this process is a "conceptual heatmap", which precisely localizes the targeted concept within the input image.

This capability is particularly valuable for uncovering contextual biases within models. For instance, a model trained to detect a frisbee may rely more on the frequent co-occurrence of a dog beside it in the training data, rather than on the visual features of the frisbee itself. By computing context scores for various concepts, L-CRP can expose such undesirable dependencies on contextual elements, thereby facilitating model improvement, dataset refinement, or the development of more ethically responsible AI systems [45].

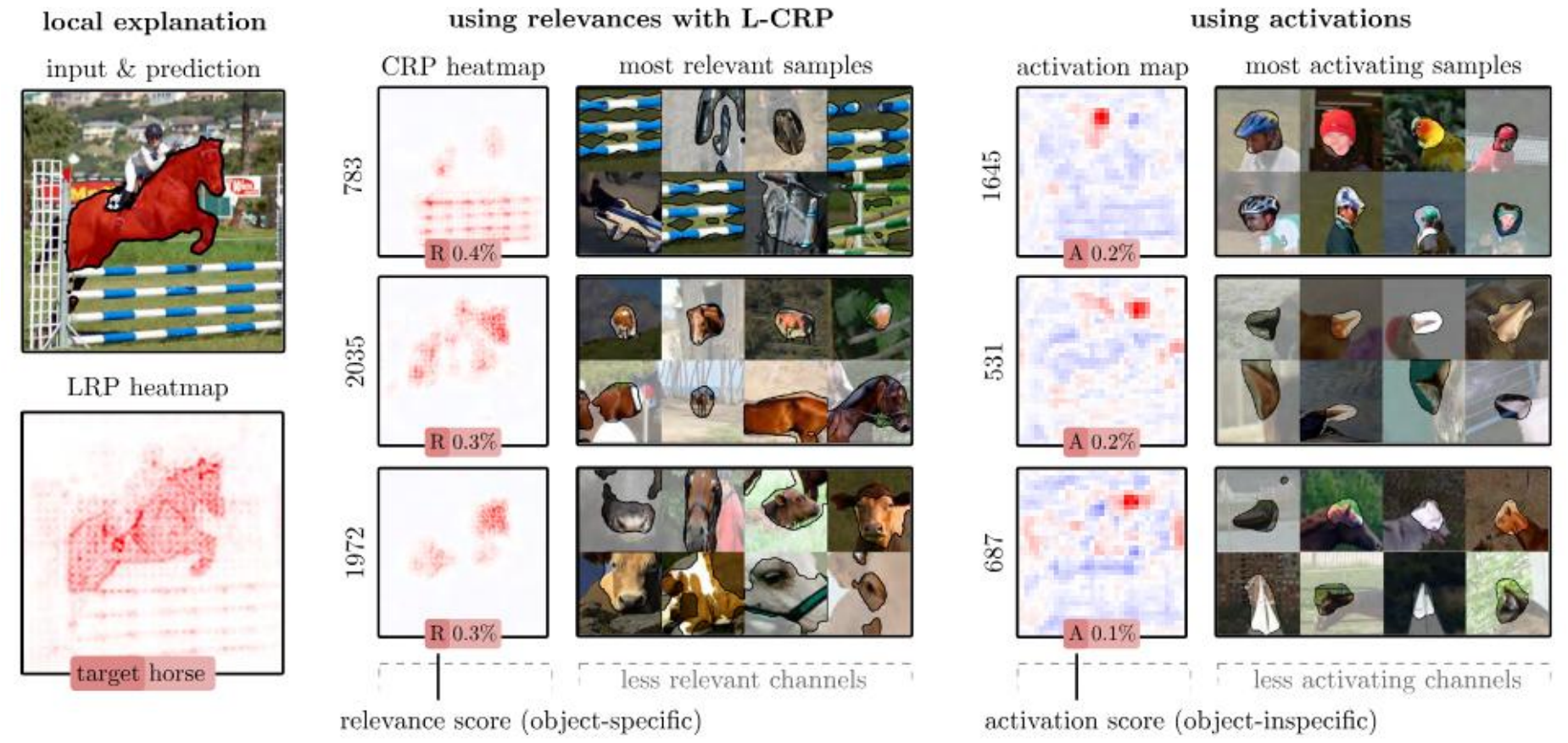

*Figure 33: This section compares three different methods for explaining deep learning models in the task of object detection: Localized Heatmaps (left), Concept-based Explanation using L-CRP (center), and Activation Analysis (right). In the left section, the heatmap generated using the LRP method demonstrates which regions of the image the model focused on to identify the horse. However, this method only indicates the pixel-level relevance without referencing higher-level concepts. In the middle section, L-CRP illustrates how specific latent concepts (e.g., vertical bars blocking a jump or regions of the horse's body) quantitatively and precisely contributed to the prediction of the "horse" class. Here, for each latent channel, a concept-based heatmap is provided, alongside examples of images most strongly associated with that concept. The percentage R-values also indicate the degree of relevance of each specific concept to the model's decision. This explanation not only identifies the location of concepts within the image but also quantitatively measures each concept's contribution to the model's decision-making process. In contrast, the right section uses activation maps, which show which latent channels exhibited the strongest responses. However, these channels may be related to different classes or might not have been used in the final decision-making of the model at all. As a result, the comparison between L-CRP and traditional activation methods reveals that L-CRP is capable of accurately extracting class-specific concepts and pinpointing their location within the image—an essential feature for analyzing multi-object models [45].*

### 2.3.2 Summary of Backpropagation-Based Methods for Object Detection Model Explainability

This table provides an overview of methods for explaining object detection model decisions using backpropagation mechanisms. These methods offer precise, class-oriented, and concept-oriented explanations of model performance by tracing relevance backward from the output to the input, with a focus on hidden concepts.

*Table 3: Summary of the methods mentioned in this review in the backpropagation -based category*

| Method Name | Main Idea | Advantages | Key Features |
|---|---|---|---|
| **L-CRP (Localized Concept Relevance Propagation)** | Combines relevance propagation with hidden concepts to generate conceptual heatmaps. | Extracts the role of specific concepts (e.g., skin, obstacles, textures) in model decisions. | Reveals contextual biases, class-oriented concept analysis, preserves exact location of concepts. |

## 2.4 Graph-Based Method

We propose a complementary approach based on graphs. This method represents the model's output in the form of a graph composed of active regions or key features, where the nodes represent meaningful parts of the image or intermediate features of the network, and the edges denote spatial, conceptual, or hierarchical relationships between them. Unlike attention maps, which merely indicate the distribution of importance within an image, the graph-based explainability approach provides a richer and more interactive structure of features, allowing the analysis of how the components combine to reach the model's final decision.

### 2.4.1 AOG Parstree Method

The AOG (AND-OR Graph) method is an approach for explainability in object detection systems such as Faster R-CNN. Instead of relying solely on the features extracted by the Fully Connected layers in a flat, incomprehensible structure, this method introduces a hierarchical and compositional grammatical model into the detection process. In this model, the proposed image regions (RoIs) are decomposed into a set of latent components using a directed acyclic graph (DAG) of the AND-OR type. The AND nodes represent the combination of components, the OR nodes allow for selection between different options, and the Terminal nodes represent the final parts of the image. This graph automatically generates a parse tree for each region at runtime, which describes the internal structure of the object without requiring manual labeling of object components in the training data. The AOG Parsing operation is designed as a new, integrable component with the RoIPooling/RoIAlign layers, allowing for the discovery of hidden structures while the model is being trained. A sub-network is also used to evaluate the importance (value) of each terminal node, helping the model select more meaningful paths in the graph. This process is carried out in a weakly supervised manner, eliminating the need for labeling the internal components of objects, while still providing an intuitive explanation of the reasons behind object detections. The primary advantage of this method is that it adds more transparency to the model's outputs without compromising accuracy, representing a significant step toward explainable AI (XAI) in computer vision [46].

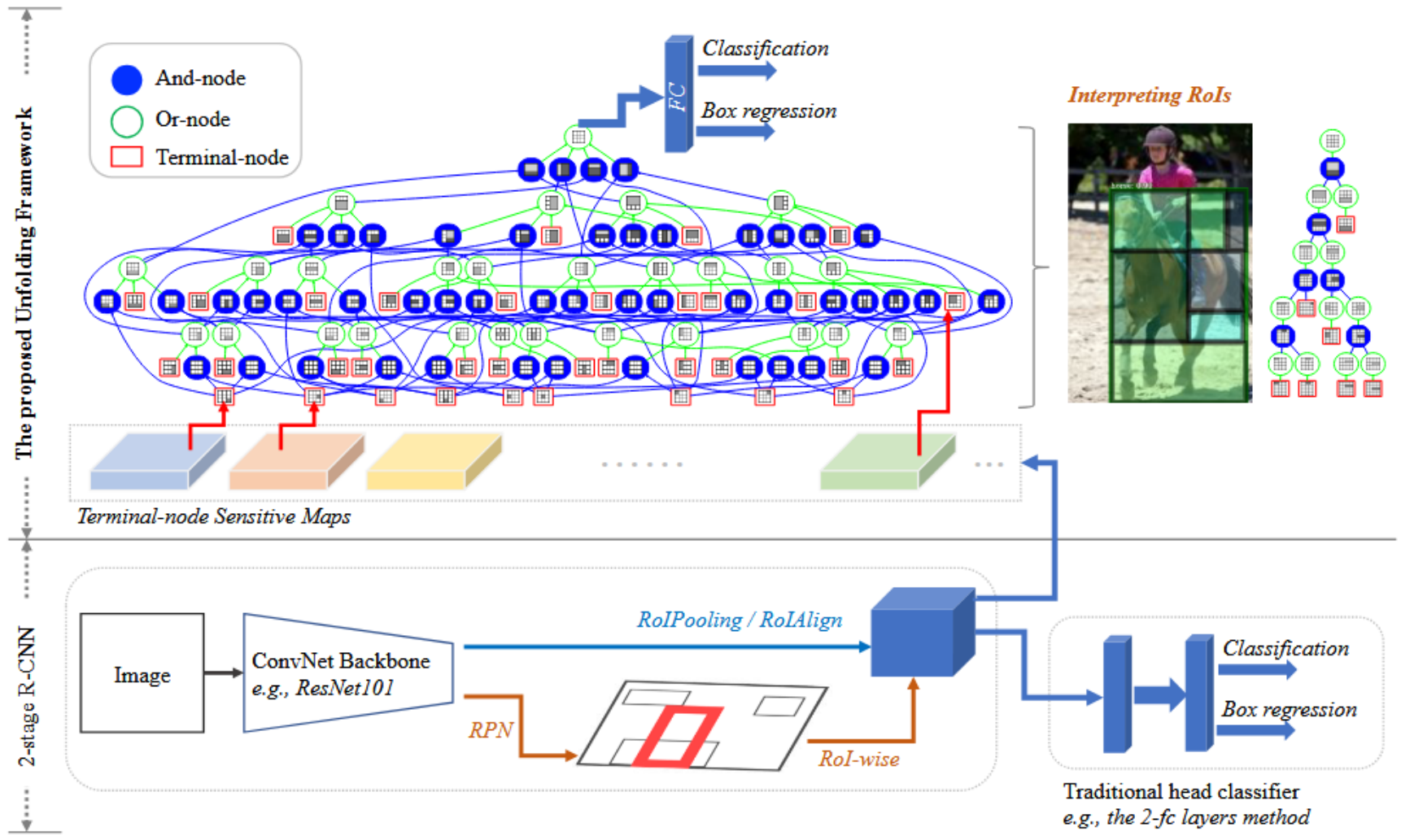

*Figure 34: Overview of the Proposed Method Based on the Faster R-CNN Model, Utilizing the AND-OR Graph (AOG) for Enhanced Explainability in Object Detection: This framework employs a top-down grammar model to represent the hidden structures of Regions of Interest (RoI) as a directed, acyclic AND-OR graph. In this graph, the AND nodes (blue) represent the combination of smaller components, the OR nodes (green) provide options for alternative decompositions, and the Terminal nodes (red) correspond to the final components linked to image features. Instead of using the traditional two fully connected layers in the network's head, the features of the proposed region are hierarchically decomposed via the AOG graph, and the optimal parse tree is selected. The final output includes explainable features for classification and bounding box regression. This approach provides greater transparency in the decision-making process while maintaining the model's accuracy [46].*

### 2.4.2 SRR Method

The SRR (Spatial Relation Reasoning) framework is an innovative approach designed to enhance the explainability of deep neural networks in object detection tasks, inspired by the human visual system. When the human brain identifies objects, it first focuses on key components such as the "head" and "body" and analyzes the relationship between them. Accordingly, SRR seeks to reconstruct this human-centered way of thinking within deep learning networks. The framework consists of two main components: the Spatial Feature Encoder (SFE) and the Graph-based Spatial Relation Encoder (GSRE).

Initially, the SFE extracts the features of the Regions of Interest (RoI) from the image. Then, the GSRE utilizes Graph Convolutional Networks (GCN) to categorize different regions into groups, such as "components of an object," based on their spatial similarity. In this process, the relationships between these components are defined using an adjacency matrix based on cosine similarity to ensure that important relationships are preserved while irrelevant ones are eliminated. Finally, the output features from both components are combined and used for final object detection and localization. The result is that the model not only improves in accuracy for object recognition but also exhibits human-like visual processing, becoming more explainable and reliable [47].

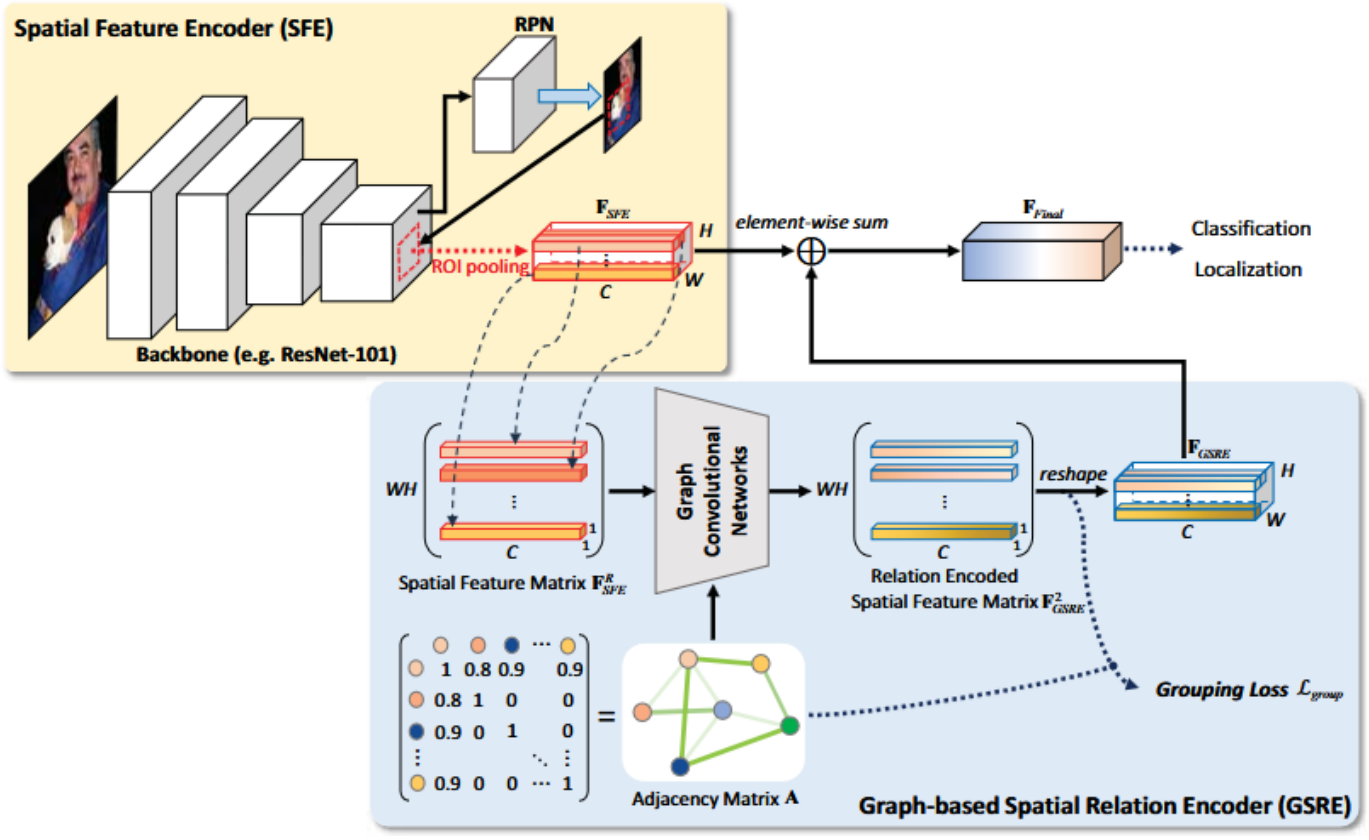

*Figure 35: The architecture diagram of the proposed framework for human-centered visual explainability in object detection is shown. This model consists of two main modules: the Spatial Feature Encoder (SFE), which extracts the spatial features of the image, and the Graph-based Spatial Relation Encoder (GSRE), which learns the spatial relationships between the features using graphs. The final output of this model is the combination of the features extracted from both modules, which are then used in subsequent layers for object recognition and localization tasks[47].*

### 2.4.3 Summary of Graph-Based Methods for Explaining Object Detection Models

In these methods, the structure and internal relationships between features or components of the image are modeled as a graph. This graph-based structure helps the model to focus not only on the intensity of features but also on their composition and interrelationships, similar to the human analysis of visual perception.

*Table 4: Summary of the methods mentioned in this review in the Graph -based category*

| Method Name | Main Idea | Advantages | Key Features |
|---|---|---|---|
| **AOG ParsTree (AND-OR Graph)** | Using hierarchical grammar in the form of a graph to parse object structure | No need for manual labeling of internal components, structural and intuitive explanation | AND/OR combination for representing component composition and selection, integration with RoIAlign in Faster R-CNN |
| **SRR (Spatial Relation Reasoning)** | Modeling spatial relationships between object components using learning graphs | Mimicking the human visual process, improved accuracy and explainability | Includes two modules: SFE and GSRE, uses GCN and adjacency matrix for analyzing component relationships |

## 2.5 Other Models

In addition to the models categorized in the previously mentioned classifications, several other approaches have been introduced in the field of enhancing the explainability of object detection models that do not fall into the previous categories. These models, utilizing creative ideas such as spatial reasoning, expanding latent structures, or combining classical and modern techniques, have aimed to strike a balance between performance accuracy and the transparency of decisions made by neural networks. Below is a brief overview of some of these methods.

### 2.5.1 iFaster-RCNN Method

The iFaster-RCNN model introduced in this paper is a combination of two well-known deep learning architectures, Faster-RCNN and ProtoPNet, designed with the aim of increasing transparency and explainability in object detection. In this approach, the Faster-RCNN model first uses a Region Proposal Network (RPN) to identify potential object regions in an image. These regions of interest (RoIs) are then sent to ProtoPNet, a model based on "prototypes." Unlike traditional models that merely predict class labels, ProtoPNet clarifies its decision-making process by comparing the features of the input region with learned prototypes from training images, which represent characteristic features of each class.

Specifically, ProtoPNet extracts a set of representative prototypes for each class (e.g., a car wheel or a human foot for the relevant classes) during training. When making predictions, the model examines which part of the input region most closely resembles these prototypes, and this comparison forms the basis for the model's decision. Ultimately, this combination enables iFaster-RCNN to not only detect objects with reasonable accuracy but also to make its decisions explainable for humans, both visually (via activation maps) and numerically (through similarity scores), without requiring post-processing methods [48].

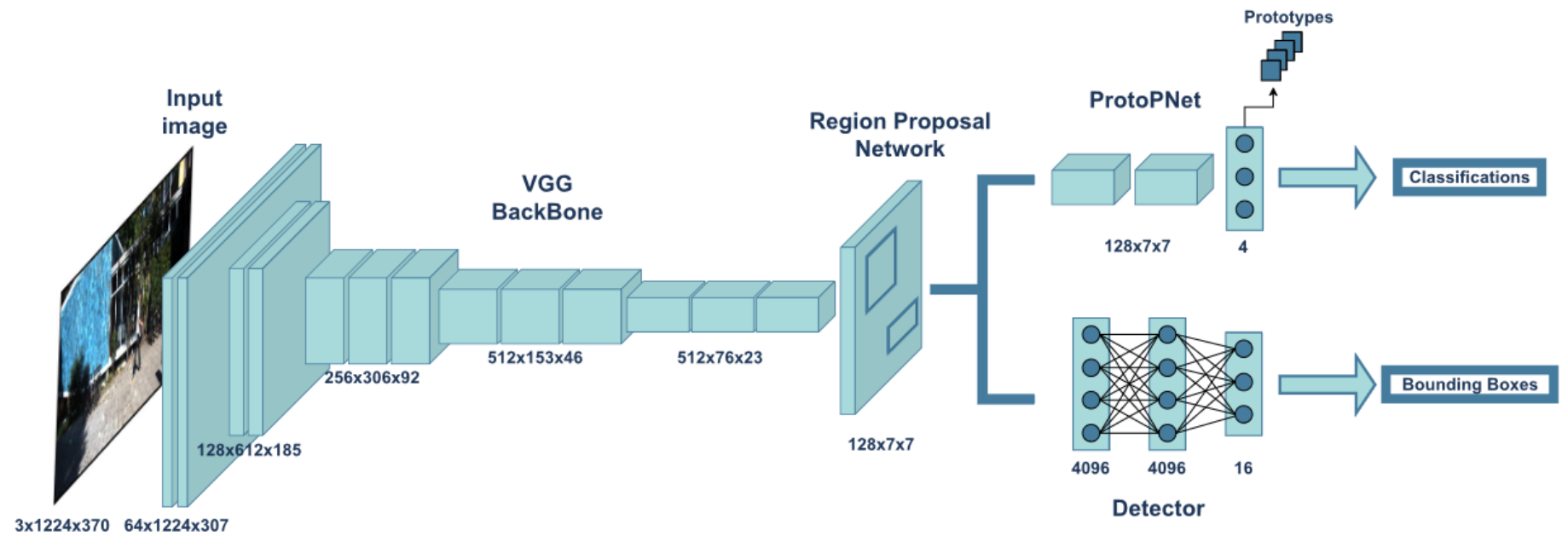

*Figure 36: Architecture of the Proposed Model: The above figure illustrates the complete architecture of the proposed model, which integrates the object detection network Faster-RCNN with the explainable ProtoPNet network. The input image is first processed by the VGG convolutional neural network (CNN), extracting low- to high-level features. Then, the Region Proposal Network (RPN) uses these features to extract potential object regions (RoIs). The output of this section, which consists of 7×7×128 feature maps for each region, is fed in parallel into two separate pathways. In the first pathway, ProtoPNet is responsible for classifying the regions and*

*providing explainability for the model's decisions, outputting the probability of each region belonging to one of the classes. In the second pathway, the detection network, derived from Faster-RCNN, predicts the exact coordinates of the bounding boxes using fully connected layers. This hybrid design enables the model not only to achieve high performance in object detection but also to offer reliable explainability of its decision-making process [48].*

### 2.5.2 A review of other methods to improve explainability in object recognition models

This method leverages an innovative and hybrid structure to increase the transparency of decisions made by deep learning models without significantly compromising accuracy. It operates outside traditional Explainable AI (XAI) frameworks and is considered complementary to existing methods.

*Table 5: A summary of the methods mentioned in this review in the other category of models*

| Method Name | Core Idea | Advantages | Key Features |
|---|---|---|---|
| **iFaster-RCNN** | Combination of the Faster-RCNN object detection network with the interpretable ProtoPNet network based on prototype samples. | Provides both visual and numerical explanations without the need for post-processing methods. | Displays regions related to class prototypes, enhances transparency in decision-making, and maintains model accuracy. |

# 3 Datasets

Based on the reviewed articles in the field of explainable artificial intelligence (XAI) for object detection, the following datasets have been utilized:

## 3.1 MS-COCO (Microsoft Common Objects in Context)

The Microsoft COCO (Common Objects in Context) dataset is one of the most comprehensive and reputable datasets for object detection and segmentation in natural contexts. This dataset includes over 328,000 images and 2.5 million labeled samples from 91 common object categories, with 82 of these categories containing more than 5,000 samples. Unlike iconic images, the images in this dataset depict real, complex scenes where objects appear from unusual angles, with partial occlusion, among multiple objects, and within natural settings. COCO has undergone three labeling stages through the Amazon Mechanical Turk platform and the design of intelligent user interfaces. These stages include identifying object categories, marking the location of objects, and performing pixel-by-pixel segmentation. The dataset focuses particularly on distinguishing different instances of the same category, offering not only bounding boxes but also precise segmentation masks. With an average of 7.7 objects per image, COCO provides an excellent resource for training deep models capable of scene understanding and object detection in complex conditions. COCO is also used as a benchmark for evaluating model performance on tasks such as detection, segmentation, and captioning [49].

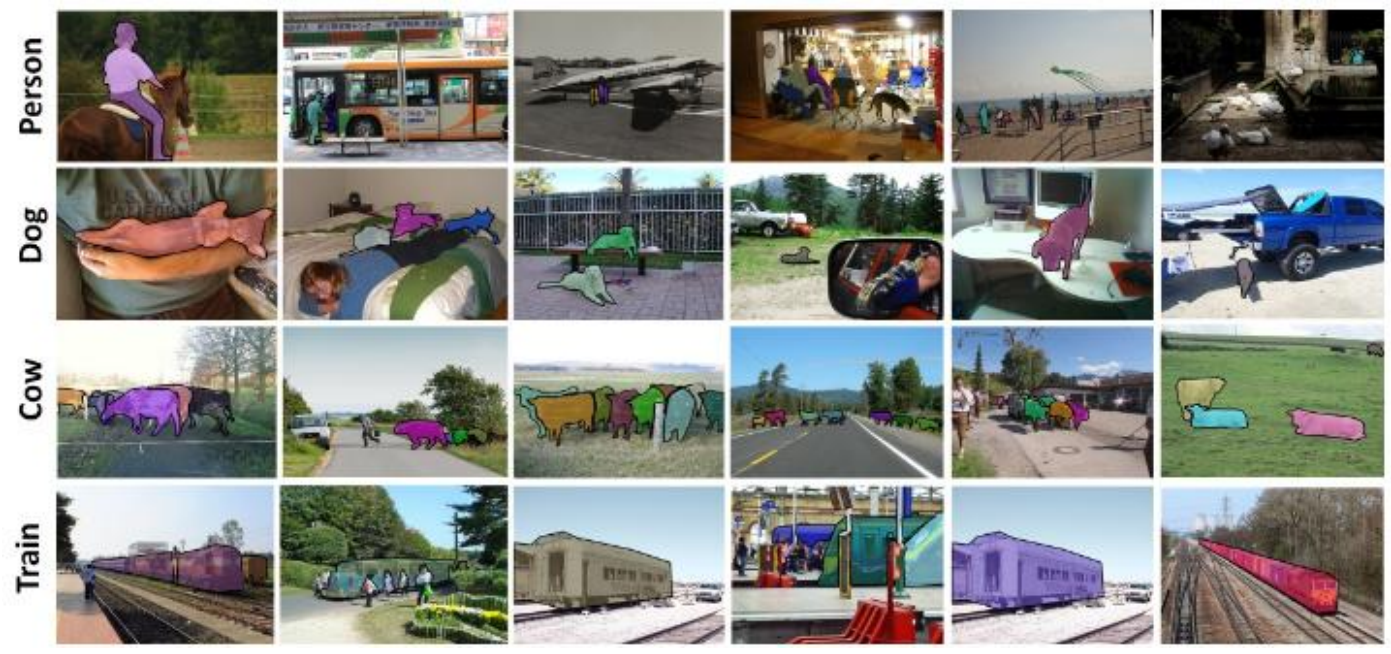

*Figure 37: Examples of Annotated Images in the MS-COCO Dataset [49].*

### 3.2 PASCAL VOC

The PASCAL VOC (Visual Object Classes) dataset is one of the most important reference datasets in the field of computer vision and object detection. Introduced in 2005 as part of the annual VOC challenge, the VOC2007 version includes 9,963 images with 24,640 labeled object instances across 20 different classes, such as humans, animals (e.g., dogs and cats), vehicles (e.g., cars, airplanes, motorcycles), and household items (e.g., tables and chairs). The images were collected from the Flickr website and feature high variability in terms of size, viewpoint, lighting, occlusion, and composition. The primary goal of this dataset is to provide a standard platform for training and evaluating object detection and image classification models.

PASCAL VOC includes several challenging tasks, such as object detection, classification, pixel-level segmentation, and human body part detection (Person Layout). All data have been carefully and manually annotated according to specified guidelines. For evaluation, metrics such as mean Average Precision (mAP) and Intersection over Union (IoU) are employed. Due to its precise structure, diverse images, and standardized evaluation process, this dataset is considered one of the foundational resources for developing and evaluating machine vision algorithms [50].

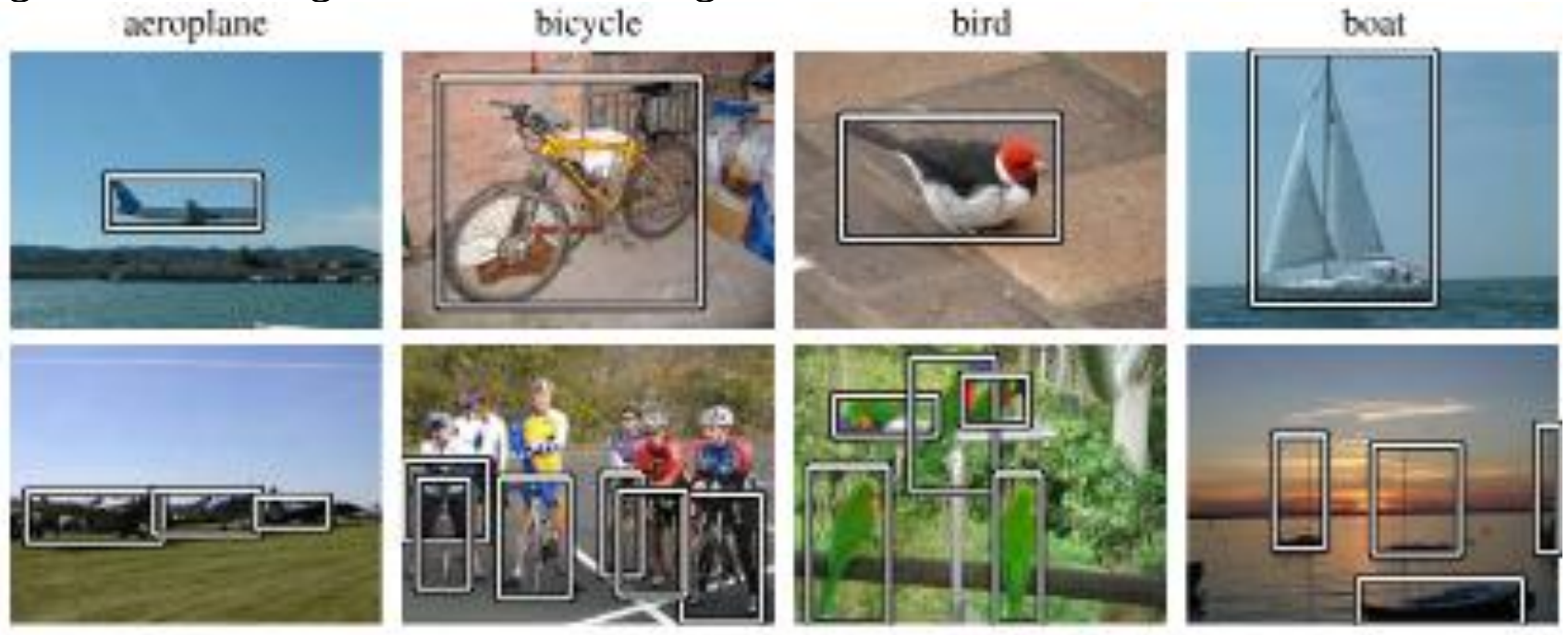

*Figure 38: Sample Images from the VOC2007 Dataset: For each of the 20 classes available, two examples are shown [50].*

### 3.3 CrowdHuman dataset

The CrowdHuman dataset is a benchmark and highly challenging dataset specifically designed for human detection in crowded and densely populated environments. It was developed to address the challenges posed by occlusion, which frequently impairs detection accuracy in such settings. The dataset comprises approximately 25,000 images containing more than 470,000 human instances, with an average of 22.6 persons per image—significantly higher than earlier datasets such as Caltech, CityPersons, and COCO.

For each human instance, three types of bounding boxes are meticulously annotated: 1)Head. 2)Visible region of the body. 3)Full body

These annotations are provided separately and with high precision, enabling detailed analysis and robust model training. The images in this dataset represent a broad diversity of scenes, positions, and viewpoints and are primarily sourced through extensive keyword-based searches on the internet.

Due to its comprehensive coverage of various occlusion levels—ranging from mild to severe—and its large data volume, CrowdHuman is not only suitable for the accurate training and evaluation of person detection models, but it also demonstrates strong cross-dataset generalization capabilities. This makes it a powerful pretraining resource for a wide range of human detection tasks, showing effective transferability to other datasets such as Caltech, CityPersons, COCOPersons, and Brainwash [51].

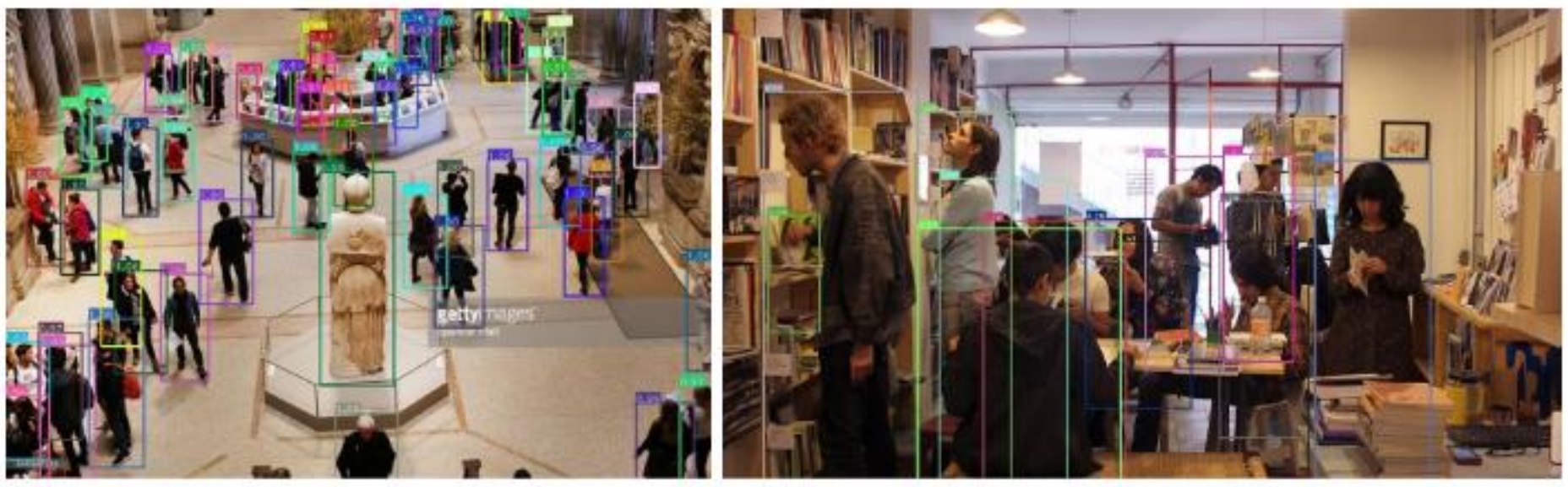

*Figure 39: Examples of Annotated Images in the CrowdHuman Dataset [51].*

### 3.4 BDD-100K dataset (Berkeley DeepDrive)

BDD100K is a large-scale and diverse dataset designed for autonomous driving applications. It contains 100,000 high-resolution (720p) driving videos, each approximately 40 seconds in duration, recorded at a frame rate of 30 frames per second. The dataset has been collected from over 50,000 distinct driving trips across various regions of the United States, including New York and the San Francisco Bay Area. It encompasses a wide range of geographic locations, lighting conditions (daytime, nighttime, sunrise/sunset), and weather scenarios (clear, rainy, snowy, foggy, etc.).

BDD100K provides a rich set of annotations tailored for ten distinct computer vision tasks, including: Object detection, Lane marking recognition, Drivable area segmentation, Semantic and instance segmentation, Multi-object tracking, Imitation learning, Domain adaptation.

This diversity of tasks and data types makes BDD100K a comprehensive reference for heterogeneous multitask learning, enabling rigorous evaluation of model performance under complex and realistic driving conditions.

Furthermore, the inclusion of GPS/IMU data and temporal video structure enhances its suitability for driver behavior modeling and imitation learning applications. These features allow researchers to develop and assess systems that more closely mirror the dynamic and context-sensitive nature of real-world autonomous driving [52].

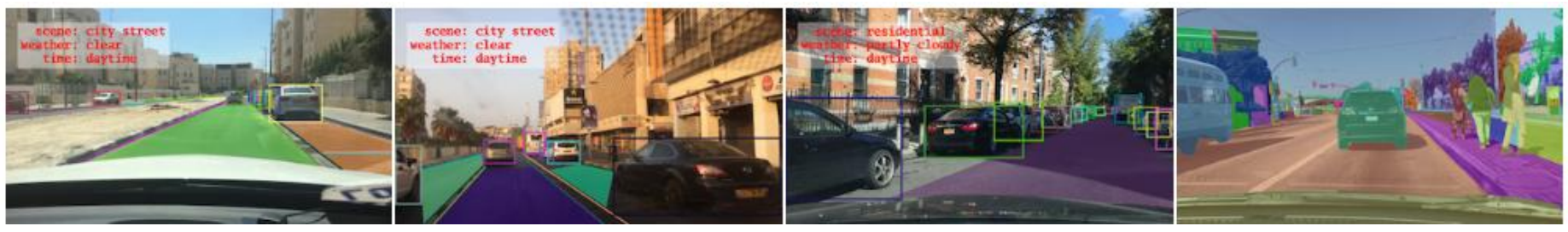

*Figure 40: Examples of Annotated Images in the BDD-100K Dataset [52].*

## 3.5 KITTI dataset

The KITTI dataset is one of the most well-known and widely used benchmarks in the fields of computer vision and mobile robotics, particularly in autonomous driving research. The dataset was recorded using a Volkswagen Passat equipped with multiple sensors, including stereo color and grayscale cameras, a Velodyne 3D LiDAR scanner, and a high-precision GPS/IMU navigation system.

The recorded scenarios cover real-world driving environments, such as highways, urban areas, residential neighborhoods, and university campuses. In total, KITTI provides approximately 6 hours of sensor data, recorded at frequencies ranging from 10 to 100 Hz. All data are fully calibrated, synchronized, and annotated, and include: Raw and rectified stereo images ,3D point clouds, GPS/IMU measurements, Object labels in the form of 3D tracklets (i.e., temporally tracked 3D bounding boxes)

The labeled object classes include cars, trucks, cyclists, pedestrians, and others.

In addition to the dataset itself, a development kit is available for both MATLAB and C++, facilitating data processing and algorithm implementation. The KITTI website also hosts multiple evaluation benchmarks for key vision and robotics tasks such as: Object detection, Depth estimation, Optical flow, SLAM (Simultaneous Localization and Mapping)

These benchmarks have established KITTI as a foundational resource for developing and benchmarking algorithms under realistic autonomous driving conditions [53].

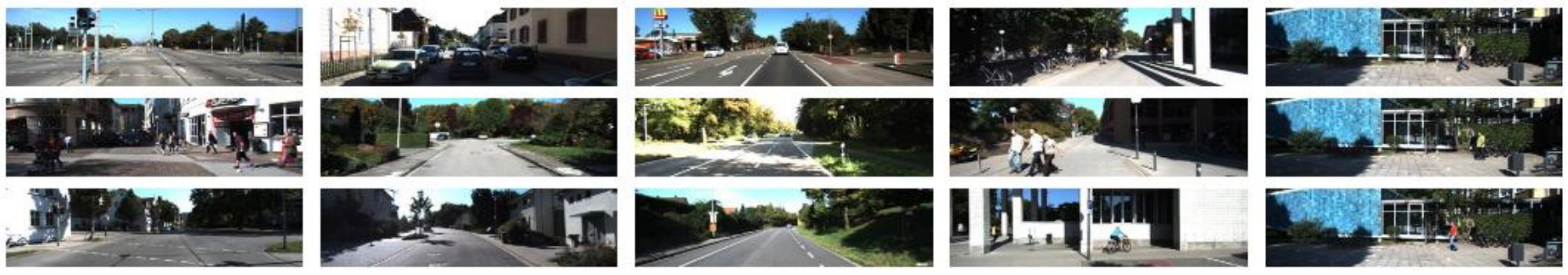

*Figure 41: Examples of Annotated Images in the KITTI Dataset [53].*

### 3.6 GRAZPEDWRI-DX dataset

The GRAZPEDWRI-DX dataset is a comprehensive, high-quality collection of digital wrist radiographs of pediatric patients with traumatic injuries, compiled at the University Hospital of Graz (Austria) between 2008 and 2018. This dataset comprises 20,327 images from 6,091 patients, ranging in age from 0.2 to 19 years. All images are stored in 16-bit grayscale format and have been meticulously annotated by pediatric radiologists using specialized tools.

The annotations include 67,771 objects—such as fractures, periosteal reactions, metallic implants, and skeletal abnormalities—labeled using bounding boxes, lines, and polygons. Additionally, the dataset features 74,459 image-level labels. It is released in standard machine learning formats, including YOLOv5 and Pascal VOC, making it highly suitable for computer vision tasks, particularly automatic fracture detection in the domain of object detection.

Preliminary evaluations using the YOLOv5 model demonstrated that the model could identify fractures with an accuracy of 91.7% and a mean Average Precision (mAP) of 0.933. These results underscore the dataset's significance for research in medical imaging and deep learning [54].

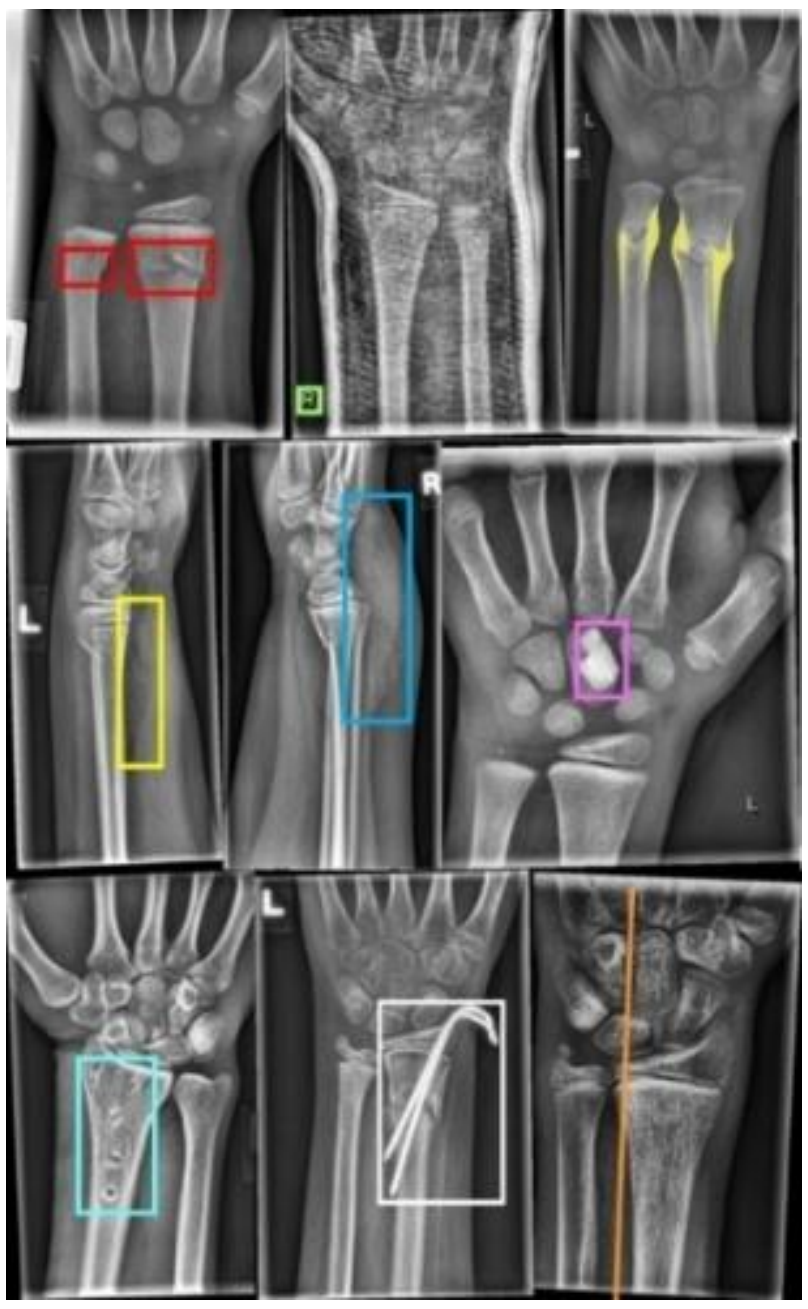

*Figure 42: Examples of Annotated Images in the GRAZPEDWRI-DX Dataset [54].*

### 3.7 Summary of the Datasets Used in This Review

Below is a summary of the datasets used in the reviewed articles:

*Table 6: Summary of datasets used in the reviewed articles*

| Dataset Name | Number of Images | Data Type and Classes | Key Applications and Features |
|---|---|---|---|
| **MS-COCO** | Approximately 328,000 images | Includes 91 common object classes (e.g., human, animals, vehicles, household items, etc.) with approximately 2.5 million labeled samples; precise instance-level labeling including bounding boxes and mask segmentation. | Suitable for object detection, instance segmentation, scene understanding, and image captioning. Contains natural, non-iconic images with an average of 7.7 objects per image. High-accuracy labeling using crowdsourcing and multi-stage annotation processes. |
| **PASCAL VOC 2007** | 9,963 images in the final set | Includes 20 common object classes such as person, cat, dog, car, airplane, bicycle, table, chair, train, etc. Each object is labeled with a bounding box and, in some cases, additional labels such as viewpoint, occlusion, and difficulty. | A classic standard for early object detection algorithms. |
| **CrowdHuman** | 25,000 images (15,000 train, 4,370 val, 5,000 test) | Three bounding boxes per person: head, visible body, and full body | Human detection in crowded scenes with high occlusion; high density (avg. 22.6 persons/image); diverse scenes, viewpoints, and cities; ideal for pretraining detectors. |
| **BDD-100K** | 100,000 video clips (each ~40 seconds | High-resolution images and videos annotated for 10 tasks including: object detection (10 classes), semantic & instance segmentation (40 classes), multi-object tracking, lane marking, drivable area, scene tagging, imitation learning, etc. | Heterogeneous multitask learning for autonomous driving; rich diversity in location, time, weather; supports domain adaptation and driver behavior analysis with GPS/IMU data. |
| **KITTI** | Over 150,000 images | Color & grayscale images, 3D point clouds, GPS/IMU data, 3D bounding boxes<br>Classes: Car, Van, Truck, Pedestrian, Cyclist, Sitting Person, Tram, Misc | 3D Object Detection, Tracking, Stereo Depth Estimation, Optical Flow, SLAM<br>Features: Calibrated & synchronized sensors, real-world driving scenarios, diverse environments (urban, residential, highway), public benchmarks available. |
| **GRAZPEDWRI-DX** | 20,327 wrist radiographs | Pediatric wrist X-ray images (16-bit PNG), annotated with classes such as fracture, periosteal reaction, bone anomaly, metal, soft tissue swelling, etc.; 67,771 labeled objects and 74,459 image tags. | Automated bone injury detection (e.g., fractures) via machine learning and computer vision; suitable for supervised learning in *Object Detection* tasks; provided in YOLOv5, Pascal VOC, and Supervisely formats; de-identified, expert-reviewed dataset |

# 4 Evaluation Criteria

There are various criteria for evaluating and comparing explainable artificial intelligence (XAI) methods. Some of the most important of these are summarized in Table 7.

*Table 7: Some criteria for evaluating explainable AI [55].*

| Features | Concept |
|---|---|
| **Algorithmic Transparency** | The degree of confidence that the learning algorithm behaves in a clear and understandable manner. |
| **Causality** | The ability of the explanation method to establish clear and transparent links between model inputs and outputs. |
| **Completeness** | The extent to which the explanation is comprehensive and flawless. |
| **Informative-ness** | The quality and depth of the explanation in effectively conveying concepts. |
| **Cognitive Relief** | The ability of the explanation method to reduce ambiguities and enhance user understanding of the model. |
| **Modifiability** | The ability of the explanation method to adapt and improve models based on feedback and changes. |
| **Effectiveness** | Assisting in better decision-making by providing accurate and practical explanations. |
| **Efficiency** | The model's ability to facilitate faster decision-making for end users through clear explanations. |
| **Clarity** | The speed and clarity with which an explanation method can describe a model. |
| **Fidelity** | The degree of alignment between the explanation and the actual relevant features of the model, selecting them precisely. |
| **Understanda-bility** | The ability of the explanation method to provide understandable and comprehensible explanations, fostering greater understanding. |
| **Comprehensi-bility** | The explanation method should be able to generate complex concepts in a manner that is accessible to various users. |
| **Interactivity** | The ability to respond to user questions and provide active, interactive explanations of the model. |
| **Engagement** | Creating engagement in the explanation by offering appealing explanations and considering the user's needs. |
| **Instructiveness** | The ability to provide useful and comprehensive information to users through model explanations. |
| **Persuasiveness** | Convincing users to make specific decisions through effective explanations. |
| **Robustness** | The ability of the explanation method to withstand input changes while maintaining the accuracy of the explanation output under varying conditions. |
| **Satisfaction** | Enhancing user satisfaction with the model by providing appropriate and practical explanations. |
| **Fault Diagnosis** | The ability of the explanation method to detect flaws and deficiencies in the model and offer corrective feedback. |
| **Simplicity or Selectivity** | The ability to choose key and important information from among diverse data to simplify the model's explanation. |
| **Sensitivity** | The ability of the explanation method to demonstrate the model's sensitivity to different inputs and changes in its underlying layers. |
| **Accuracy** | The degree of precision and correctness in the explanations provided by the method, reflecting the true behavior of the model. |
| **Stability** | The ability to maintain a consistent explanation when similar inputs are received by the model, ensuring explanation stability. |
| **Speed** | The generation of explanations in the shortest time possible to facilitate quick decision-making by users. |

Ultimately, explainable models are preferred over more complex models due to their transparency, ability to clarify causal relationships, and increased user trust.

One of the key challenges in the field of Explainable Artificial Intelligence (XAI), especially in the context of object detection, is the development and implementation of precise, valid, and comprehensive evaluation metrics for assessing the quality of model explanations. In contrast to the image classification field, where extensive and systematic studies have been conducted on evaluating explanation methods, research on evaluating explainable methods in object detection remains limited and fragmented. As a result, many studies are forced to use evaluation criteria and methods developed for image classification, with only a few studies specifically introducing and implementing metrics tailored to the object detection domain.

A review of reputable articles and resources in the XAI field, particularly in object detection, has led to the extraction of a set of evaluation criteria, each focusing on a specific aspect of the quality and effectiveness of explainable methods. These criteria are categorized into five main groups based on the evaluation objectives and the features being assessed.

## 4.1 Fidelity

Alignment of Explanation Maps with the Actual Behavior of the Model. Fidelity assesses whether an explanation genuinely reflects the features utilized by the model during decision-making.

### 4.1.1 Deletion

One of the most important and widely used evaluation metrics in Explainable AI (XAI), the Deletion method involves systematically removing image pixels based on their saliency ranking in the saliency map, from most to least salient. The model's performance is then compared between the original and the altered images, and the drop in prediction accuracy is quantified. This difference is measured using a metric denoted as $DDR_{I,\acute{I}}$, which combines the intersection-over-union (IoU) of detected regions and the model's confidence in both images. If the highlighted areas indeed represent critical regions related to the object, their removal should result in a steep performance drop. Therefore, a sharp decline in $DDR_{I,\acute{I}}$ indicates a high-quality saliency map and the strong explanatory power of the model [20].

$$DDR_{I,\acute{I}} = \frac{IoU_{I,\acute{I}} \times CS_{\acute{I}}}{CS_{I}}$$

### 4.1.2 Insertion

This method operates in contrast to *Deletion*. Starting with an empty image, salient pixels are gradually inserted in order from most to least important. The model's performance is assessed as these regions are introduced. A steeper increase in $DDR_{I,\acute{I}}$ indicates a stronger correlation between the highlighted regions and the target object [20].

### 4.1.3 i-AUC and d-AUC

The Insertion and Deletion processes used for evaluating explanation fidelity rely on the generated saliency map to identify important image regions. Both processes are performed incrementally over 100 discrete steps. In each step, 1% of the total area of the bounding boxes corresponding to objects detected by the model is either inserted into or removed from the image. At each stage, the model's confidence in its predictions is measured and recorded. The resulting data forms insertion and deletion curves, and the area under these curves—denoted by i-AUC and d-AUC, respectively—serves as a quantitative fidelity metric. A high i-AUC indicates that the highlighted regions have a strong positive effect on the model's decision-making, while a low d-AUC suggests that their removal significantly undermines the model's confidence. Thus, the combination of these two indices reflects greater alignment between the saliency map and the model's decision-making process, thereby indicating higher quality of the explainability method [39].

### 4.1.4 Minimum Subset

This metric is designed to assess the accuracy and fidelity of saliency maps. It identifies the minimum number of important pixels that must be removed to cause the model to change its initial prediction. In other words, the *Minimum Subset* is the smallest critical subset of salient regions whose removal results in a shift in the model's predicted class. Like Deletion, pixel removal follows the importance ranking derived from the saliency map. However, instead of monitoring gradual performance changes, this method aims to pinpoint the threshold at which the model can no longer correctly identify the target object. The fewer the pixels that need to be removed to reach this threshold, the higher the accuracy and effectiveness of the saliency map in identifying crucial regions [37].

### 4.1.5 D-Deletion

Since the Deletion metric is tailored for classification tasks, it may not suffice in object detection scenarios where multiple instances of the same class may appear in a single image. The D-Deletion metric is proposed to address this limitation and provide a more nuanced assessment of explanation fidelity in object detection models. Unlike class-based Deletion, D-Deletion focuses on a specific target bounding box and considers only those regions that match the target class and exhibit a high degree of spatial overlap (IoU) with it. By progressively removing important pixels and monitoring the reduction in the objectness *score* for the target box, this metric evaluates the performance of the saliency map more precisely under conditions involving multiple similar objects. A rapid score decline in this process indicates high fidelity of the saliency map in explaining the behavior of the object detection model [37].

### 4.1.6 D-Minimum Subset

In alignment with the established relationship between the Minimum Subset and Deletion, the D-Minimum Subset metric is considered an extension of D-Deletion. Its objective is to identify the minimum number of pixels whose removal not only alters the predicted class for the target bounding box but also reduces the box's Intersection over Union (IoU) below a predefined threshold. Essentially, this metric imposes two simultaneous conditions:
(1) the model no longer recognizes the target class, or (2) the spatial location of the detected object is altered to the extent that it significantly deviates from the ground truth bounding box.

As such, the D-Minimum Subset incorporates both semantic information (i.e., class) and spatial information (i.e., position) into its analysis. It serves as an effective measure for evaluating the accuracy of saliency maps in identifying critical regions of a specific object, particularly in complex object detection tasks [37].

#### 4.1.7 Pointing Game (PG)

The Pointing Game is a human-centric evaluation method used to assess the accuracy of saliency maps in localizing the target object within an image. In this method, if the point on the saliency map with the highest activation falls within the human-annotated bounding box, it is considered a "hit"; otherwise, it is marked as a "miss."

PG accuracy is calculated as the ratio of the number of hits to the total number of hits and misses. The final reported value is the average accuracy across all categories in the dataset. Owing to its simplicity and reliability, PG has been widely adopted as a qualitative benchmark in many explainability studies [37].

#### 4.1.8 Energy-Based Pointing Game (EBPG)

The Energy-Based Pointing Game (EBPG) is a quantitative and energy-driven variant of the traditional PG. Instead of relying on a single point of maximum saliency, EBPG evaluates the overall concentration of activation within the region of interest. Specifically, it computes the ratio of the sum of activation values inside the bounding box to the total activation across the entire saliency map.

A higher EBPG value indicates a greater model focus on the relevant object region, reflecting higher fidelity in the explanation process. By analyzing the energy distribution over the saliency map, EBPG provides a more fine-grained metric for assessing the quality of explanation localization and is frequently employed in recent studies for comparing explainability algorithms [37].

#### 4.1.9 Visual Explanation Accuracy (VEA)

The VEA metric is introduced as a method for assessing the fidelity and correctness of visual explanations in deep learning models. It evaluates how well high-importance regions in the saliency map align with the actual object regions in the image. To calculate this metric, the saliency map is first binarized at various threshold levels. At each level, the Intersection over Union (IoU) between the resulting binary region and the ground truth segmentation mask of the object is computed. Subsequently, an IoU-versus-threshold curve is plotted, and the area under this curve (AUC) is reported as the final VEA value. A higher VEA score indicates a greater alignment between the saliency map and the true object region, reflecting a more accurate and faithful explanation by the model [9].

### 4.1.10 Pearson Correlation Coefficient (PCC)

The Pearson Correlation Coefficient is a widely used relative metric for measuring similarity between two saliency maps, particularly in evaluating the plausibility of explainability methods for deep learning models. This coefficient ranges from -1 to 1, where values closer to 1 represent strong positive correlation and high similarity in the spatial patterns of the two maps. In this method, saliency maps are first normalized and transformed into vectors. The correlation between these vectors is then computed to assess the structural similarity and trend consistency of important regions in the image. PCC is particularly valuable in explainability analysis due to its sensitivity to relative spatial trends and pattern variations [39].

### 4.1.11 Root Mean Square Error (RMSE)

RMSE is used as an absolute metric to quantify the numerical difference between two saliency maps. It reflects the average distance between corresponding values in the explanatory map and human attention data. This metric is computed by taking the square root of the mean of squared differences between each corresponding point in the two maps. The resulting value is non-negative, with lower values indicating higher numerical similarity and a closer match between model predictions and human focus. Due to its sensitivity to absolute differences, RMSE complements relative metrics such as PCC and plays a significant role in evaluating the precision of model explanations [39].

### 4.1.12 In–Out Importance Ratio (iratio)

The in–out importance ratio measures the relative influence of pixels inside a bounding box compared to those outside it on the model's decision. It is calculated by dividing the average absolute weight of pixels inside the box by the average absolute weight of pixels outside the box. A higher ratio indicates that the model focuses more on the features within the object's bounding box, suggesting that these internal regions are key to object recognition. This metric effectively captures the model's attention on object-specific features and serves as an indicator of explanation quality in computer vision tasks [33].

### 4.1.13 In–Out Weight Difference (wdiff)

The in–out weight difference assesses the difference in average saliency weights between pixels inside and outside the bounding box. It reveals whether the model relies more on internal object features or the surrounding context in its decision-making. A positive value indicates higher average weights within the bounding box, suggesting that the model primarily depends on internal object information to boost prediction confidence. This metric aids in understanding how the model differentiates between essential and contextual information, a crucial factor in enhancing explainability and trust in model explanations [33].

#### 4.1.14 Dummy Feature Principle

The dummy feature principle refers to the assignment of zero importance to features that do not influence the model's output. In the context of object detection explainability, this principle asserts that if a feature (e.g., a pixel or image region) does not significantly affect the model's output when altered, the explanation map should assign a value close to zero to that feature. This is evaluated by masking random image patches and observing changes in the model's output scores. If the average attribution in patches that cause minimal output change is also near zero, the method adheres to the dummy principle. Compliance with this principle indicates that the explanation method accurately filters out irrelevant information and focuses on genuinely influential features in the model's decision-making process [35].

### 4.2 Stability

The consistency of explanations across repeated executions or similar inputs.

#### 4.2.1 Convergence

The convergence criterion is employed to assess the stability and reproducibility of saliency maps in explainable models. In this approach, for a fixed input, the saliency map generation process is repeated three times, and the Euclidean distance between each pair of resulting maps is calculated. The average of these distances is then defined as the convergence value. A lower convergence value indicates a higher degree of model stability in producing similar saliency maps for a given input. This feature is of considerable importance, as the stability of saliency maps reflects the reliability and repeatability of explanations produced by machine learning models[20].

### 4.3 Cognitive Relief

Cognitive relief refers to the capacity of an explainability method to reduce the user's mental burden and uncertainty, thereby facilitating a clearer understanding of the model's behavior through the provision of focused, intelligible, and coherent explanations.

#### 4.3.1 Sparsity

The sparsity metric assesses the extent to which saliency maps are concentrated on specific regions of an image, playing a crucial role in the visual explainability of explainable methods. This criterion measures the ratio of the maximum to the mean of the normalized values in a saliency map, indicating how restricted and precise the active regions are. A high sparsity value suggests a strong focus on a limited number of pixels, which enhances explanatory clarity and ease of comprehension for the human user. Therefore, this characteristic can significantly influence the user's understanding of and trust in the model's outputs. While sparsity is secondarily related to concepts such as simplicity, selectivity, and comprehensibility, its strongest alignment is with the metric of cognitive relief. This is because visual focus and reduced informational load in the saliency map directly lead to lower ambiguity and make the explanation process more intuitive for users [39].

## 4.4 Algorithmic Transparency

Algorithmic transparency refers to the ability to comprehend a machine learning model's decision-making process by clearly presenting its structure, underlying logic, and data processing mechanisms in a way that is intelligible to users.

### 4.4.1 Linearity

The principle of linearity pertains to the explainability method's capacity to represent a combination of features through a linear composition. This principle asserts that if the model's output is a linear function of multiple components or similarities—such as spatial similarity and content similarity—then the explanation should also be derivable through a linear separation of these components. Adhering to the principle of linearity in explanations enhances the explainability of the model, enabling more accurate predictions of the individual and combined effects of input components. This characteristic is particularly vital for the in-depth analysis of complex models, such as object detection systems, as it allows for a more precise understanding of the model's decision-making structure [35].

## 4.5 Comprehensibility

Comprehensibility refers to the ability of an explainability method to provide explanations that are easily understandable by users with diverse backgrounds, enabling them to grasp the underlying meaning and establish an effective connection with the model.

### 4.5.1 User Trust

The user trust metric assesses the degree to which end users accept and trust the explanations generated by XAI (Explainable Artificial Intelligence) methods. This metric reflects how well the explanations align with human judgment and intuition, and whether they are deemed practical and acceptable under real-world conditions. Studies on user trust typically involve empirical evaluations with non-expert participants to assess the clarity and persuasiveness of model explanations. These evaluations play a vital role in determining the practical utility and success of explainability methods.

Consequently, this metric is not only linked to comprehensibility but is also closely associated with user satisfaction and partially overlaps with criteria such as persuasiveness and cognitive relief [39].

### 4.6 Completeness

Does the explanation provide a comprehensive and holistic account of the model's behavior?

#### 4.6.1 Efficiency

The principle of efficiency in explainability emphasizes that the sum of all attribution scores in a saliency map should match the model’s output (i.e., the predicted score). This principle effectively ensures that the explanation provided by the method represents a complete and precise breakdown of each feature’s contribution to the model’s prediction. When this principle is upheld, it can be confidently asserted that the model's explanation quantitatively aligns with the model itself, without missing or superfluous information.

Assessment of this criterion is carried out by computing the difference between the total value of the saliency map and the model’s output. The smaller this discrepancy, the better the explainability method performs in satisfying the principle of efficiency. These criteria have been developed specifically to tailor the evaluation of explainability methods for object detection models. They directly assess the quality of saliency maps in predicting individual objects within a multi-object image. Employing these criteria in research related to explainable artificial intelligence (XAI) in object detection can enhance both the precision and generalizability of analytical outcomes [35].

## 5 Results and Experiments

Given the diversity of explainability methods proposed in the domain of object detection, a rigorous and standardized evaluation of their performance using established metrics and well-known datasets plays a pivotal role in analyzing the strengths and weaknesses of each method.

Among the reviewed studies, the MS-COCO and PASCAL VOC datasets have been most frequently utilized as benchmark datasets for assessing explainability methods. Similarly, three core evaluation metrics—Insertion, Deletion and EPG (Explanation-based Pointing Game)—have emerged as the most commonly employed criteria in studies focused on evaluating explainability techniques in object detection.

The following tables present the evaluation results of various explainability methods based on these metrics using the widely adopted MS-COCO and PASCAL VOC datasets. These results provide a foundation for comparing performance and analyzing the advantages and limitations of each method.

In this table, the performance of various explainability methods in the context of object detection on the COCO dataset is compared. Three main evaluation metrics—Energy-based Pointing Game (EPG)**,** Insertion, and Deletion—are considered as indicators for assessing the quality of explanation heatmaps. Higher values for the EPG and Insertion metrics, and lower values for Deletion, indicate better performance of the method in generating more accurate and effective explanatory maps. The FSOD method demonstrates superior performance across most metrics compared to other methods, while the results for the BSED methods are also noteworthy, with optimal performance in the Deletion metric.

*Table 8: Comparison of the Performance of Explainable Methods on the MS-COCO Dataset Based on EPG, Insertion, and Deletion Metrics.Higher values for the EPG and Insertion metrics, and lower values for the Deletion metric, indicate better performance. This evaluation was conducted using the YOLOv5s model on a random 10% subset of the validation dataset [9], [35].*

| Method | EPG ↑ | Insertion ↑ | Deletion ↓ |
|---|---|---|---|
| **Grad-CAM(2016)** | 0.154 | 0.515 | 0.147 |
| **SS-GradCAM(2022)** | 0.729 | 0.467 | 0.176 |
| **CRP(2019)** | 0.321 | 0.158 | 0.170 |
| **CRP (YOLO)(2022)** | 0.166 | 0.461 | 0.264 |
| **E2X(2018)** | 0.332 | 0.358 | 0.097 |
| **D-RISE(2021)** | 0.163 | 0.085 | 0.047 |
| **KernelSHAP(2017)** | 0.260 | 0.198 | 0.148 |
| **FSOD (2022)** | **0.791** | 0.614 | 0.065 |
| **BSED (K=1)(2024)** | 0.211 | 0.642 | **0.034** |
| **BSED (K=2)(2024)** | 0.227 | 0.660 | 0.060 |
| **BSED (K=4)(2024)** | 0.244 | **0.667** | **0.034** |

The table below presents the performance of explainability methods on the VOC dataset, evaluated and compared across three metrics: Energy-based Pointing Game (EPG), Insertion, and Deletion. These metrics correspond to the following: EPG measures the concentration of the heatmap on the target area, Insertion reflects the incremental impact of important pixels on the model's output, and Deletion indicates the model's sensitivity to the removal of those pixels. Higher values for EPG and Insertion, and lower values for Deletion, signify better performance of the explainability method. The FSOD method demonstrates a significantly superior performance in the EPG metric, while the D-RISE method achieves the best performance in the Deletion metric. BSED methods have also shown notable results.

*Table 9: Comparison of the Performance of Explainable Methods on the VOC Dataset Based on EPG, Insertion, and Deletion Metrics. Higher values in the EPG and Insertion metrics, and lower values in the Deletion metric, indicate better performance. This evaluation was conducted using the YOLOv5s model on a random 10% subset of the validation dataset [9], [35].*

| Method | EPG ↑ | Insertion ↑ | Deletion ↓ |
|---|---|---|---|
| **Grad-CAM(2016)** | 0.165 | **0.933** | 0.133 |
| **SS-GradCAM(2022)** | 0.732 | 0.657 | 0.179 |
| **CRP(2019)** | 0.321 | 0.158 | 0.170 |
| **CRP (YOLO)(2022)** | 0.260 | 0.359 | 0.200 |
| **E2X(2018)** | 0.283 | 0.254 | 0.088 |
| **D-RISE(2021)** | 0.160 | 0.565 | **0.039** |
| **KernelSHAP(2017)** | 0.267 | 0.207 | 0.141 |
| **FSOD (2022)** | **0.777** | 0.591 | 0.072 |
| **BSED (K=1)(2024)** | 0.314 | 0.552 | 0.042 |
| **BSED (K=2)( 2024)** | 0.328 | 0.571 | 0.057 |
| **BSED (K=4)( 2024)** | 0.338 | 0.581 | 0.041 |

Based on the results gathered in the tables above, it can be observed that the explainability methods examined have demonstrated varying performances depending on the type of evaluation metric and the dataset used. Although a precise and fair comparison of all methods was not possible due to differences in experimental setups, baseline models, and evaluation metrics, the analysis of the available results can provide a better understanding of the effective trends in the development of explainability methods.

In particular, in some studies, the FSOD method has shown relatively better performance in the EPG and Insertion metrics, indicating its high capability in identifying important regions of the image. On the other hand, the D-RISE method has delivered favorable results in the Deletion metric, which could suggest its accuracy in determining the model's sensitivity to the removal of key information. Furthermore, the stable and acceptable performance of the BSED method across different versions of K was also evident in several evaluations.

Overall, although differences in experimental design prevent providing a definitive conclusion, identifying the relative strengths of each method can assist researchers in selecting explainability tools that align with specific needs in the field of object detection. The use of multi-metric and cross-evaluations across different datasets could pave the way for the development of more accurate and real-world compatible explainability methods in the future.

### 5.1 Visual Comparison in Datasets

In this section, examples of visual comparisons between different explainability methods in well-known datasets such as MS-COCO and PASCAL VOC are presented. These comparisons are intended to evaluate the methods' ability to generate explanatory heatmaps that align with human attention or the actual regions of objects.

In several studies utilizing the MS-COCO dataset, the alignment of the model-generated explanatory maps with human attention was assessed by directly comparing the heatmaps produced by various models, such as Grad-CAM, FullGradCam, FullGrad-Cam++, and Grad-CAM++, with attention maps collected from human users. The figure below illustrates an example of this comparison between the heatmaps of these models at the image level.

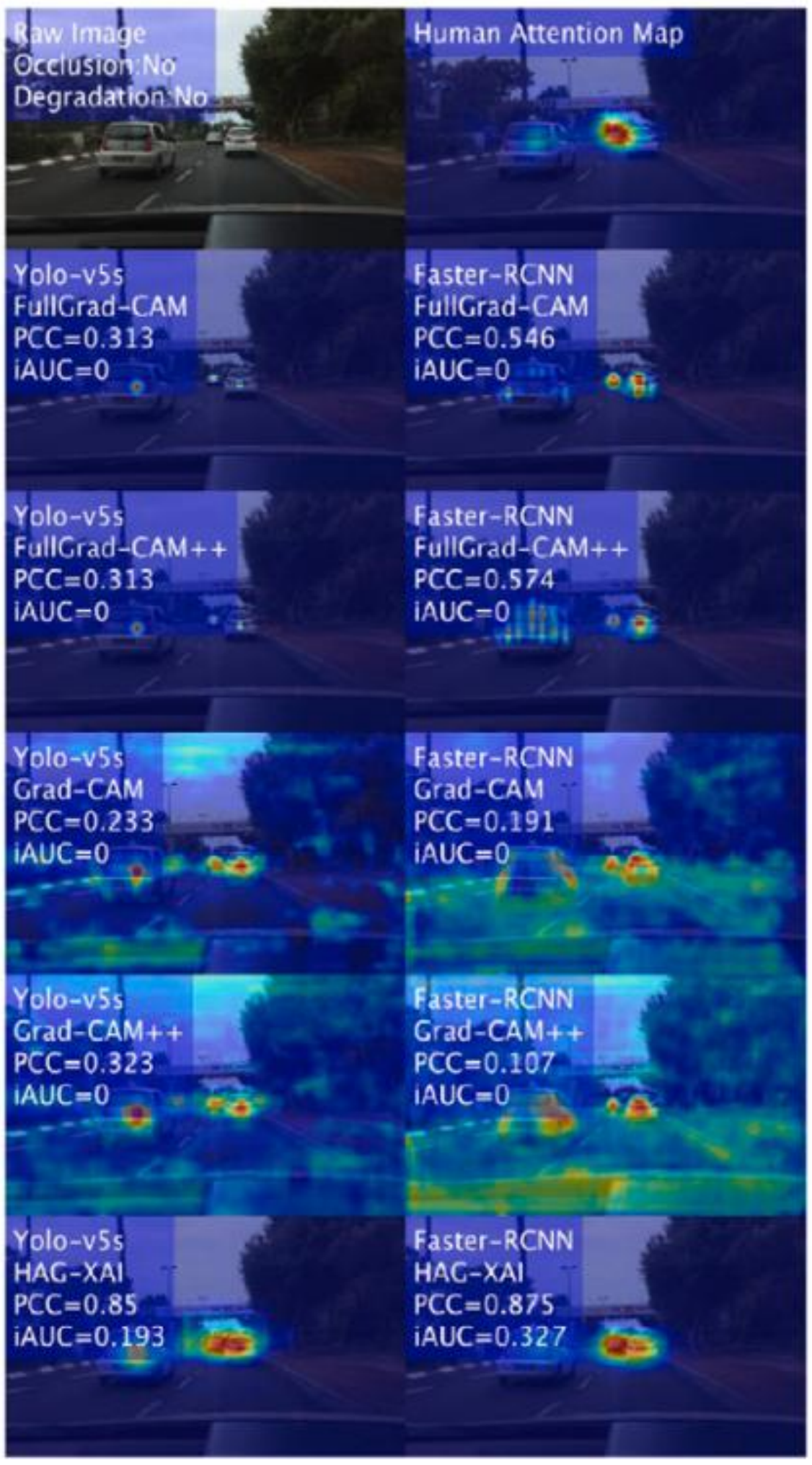

*Figure 43: Comparison of Fidelity and Plausibility at the Image Level. Human attention maps and the outputs of various explainability methods (such as Grad-CAM and Grad-CAM++) for two models, VGG-SS and Faster R-CNN, are displayed. In each case, evaluation metrics such as Pearson Correlation Coefficient (PCC), Root Mean Square Deviation (RMSD), and Area Under the Curve (AUC) have been reported to assess the degree of alignment with human attention [39].*

In addition to numerical metrics, visual comparisons between heatmaps generated by various methods also play a significant role in the qualitative understanding of their performance. In studies conducted on the MS-COCO dataset, the following figure presents an example of these visual comparisons for the ODAM method versus methods such as D-RISE, Grad-CAM, and Grad-CAM++.

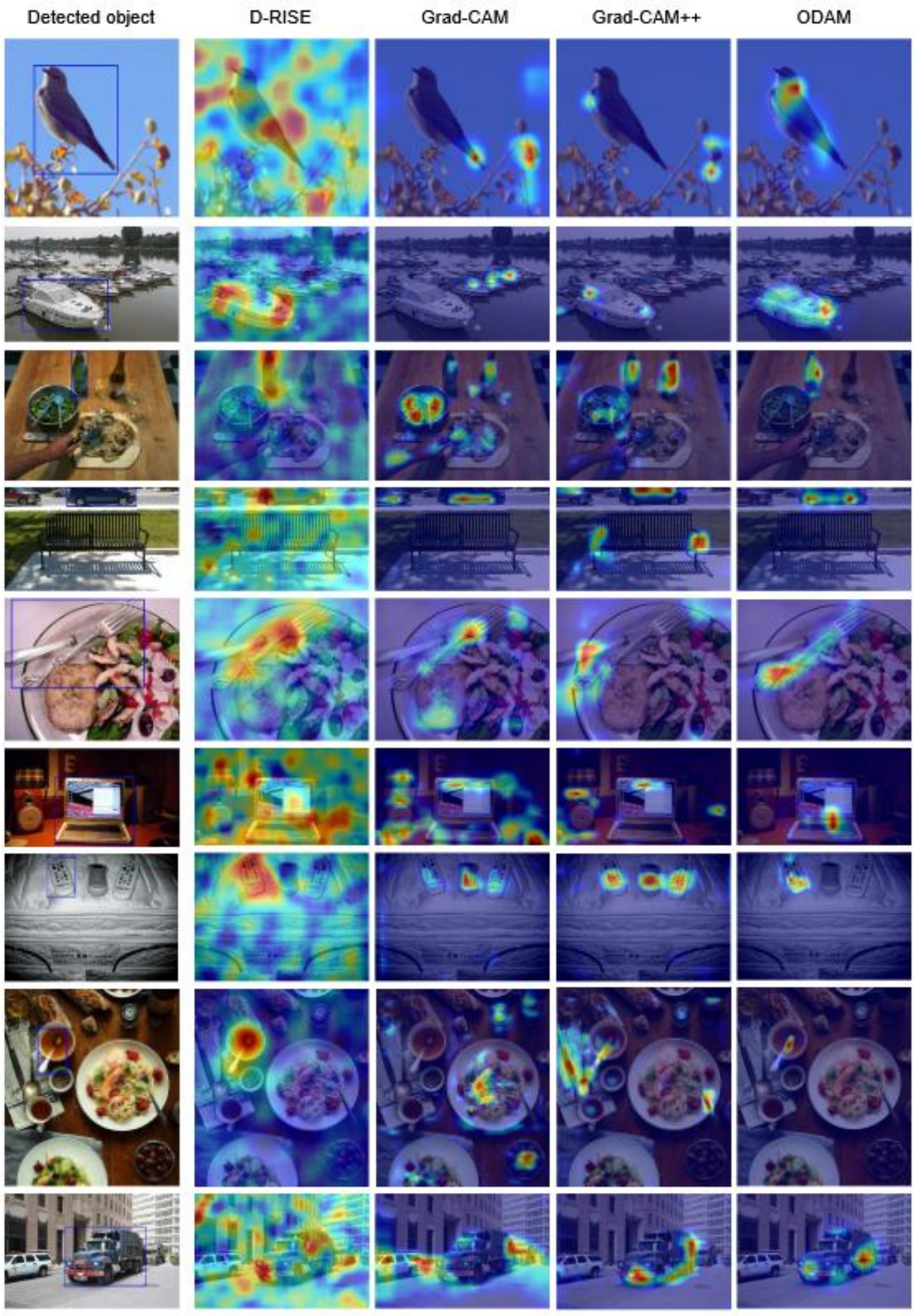

*Figure 44: Visual Comparison of the Proposed ODAM Method with Other Explainability Methods in Object Detection. This visual comparison illustrates the performance of the proposed ODAM method in relation to other explainability methods for object detection, including D-RISE, Grad-CAM, and Grad-CAM++. Each row presents a sample input image with a blue box indicating the detected object and the corresponding heatmaps generated by each method. The comparison highlights the differences in accuracy and focus of each method in emphasizing the regions related to the detected object [43].*

In several studies conducted on the MS-COCO dataset, the relative performance of various explainability methods has been evaluated in terms of accuracy in identifying meaningful regions of an image. The image below compares the outputs of three methods—Grad-CAM, SS-Grad-CAM, and SS-Grad-CAM++—on a sample image, in order to analyze how each method highlights the regions associated with the target object.

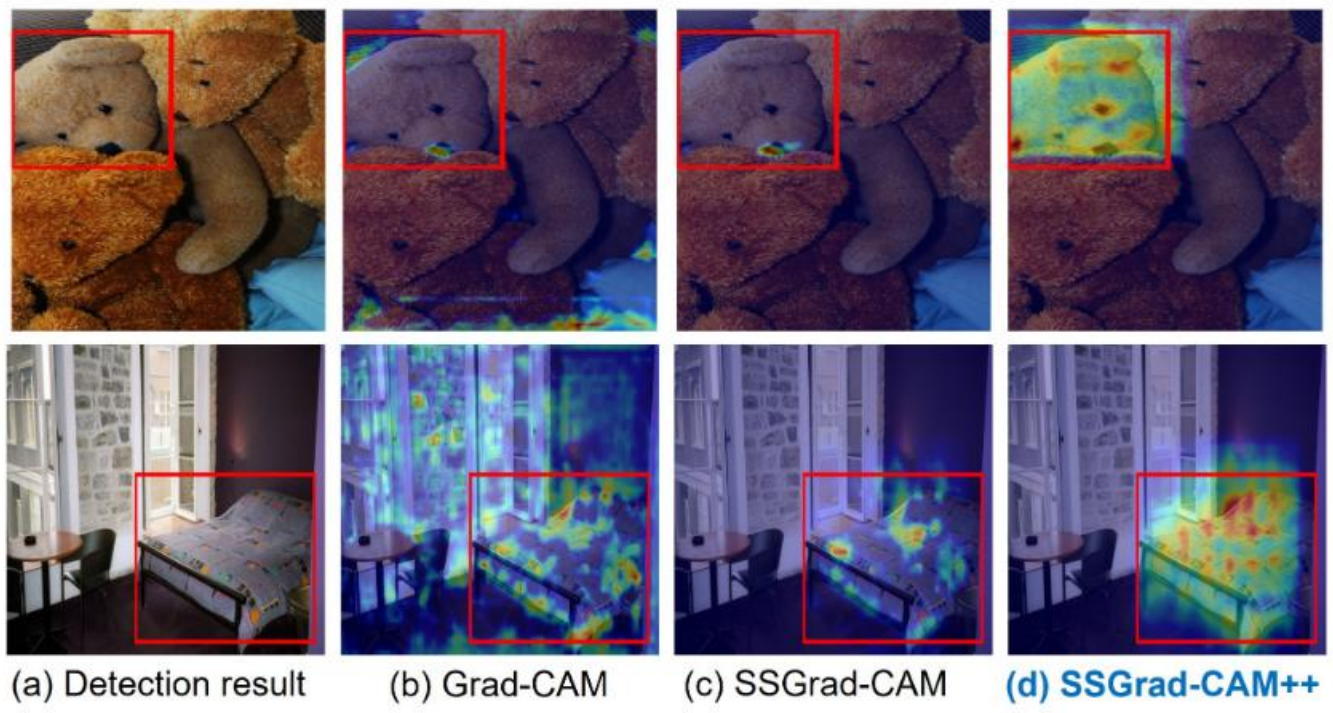

*Figure 45: Comparison of Heatmaps Generated by Different Methods for an Object Detection Sample. While Grad-CAM (b) only highlights regions related to the object class, SS-Grad-CAM (c) operates in an instance-specific manner but only emphasizes portions of the important regions. The SS-Grad-CAM++ method (d) more accurately identifies the important areas related to the detected object, encompassing not only prominent features like the nose but also other characteristics such as the eyes and ears [42].*

The figure below illustrates the performance of several different explainability methods in scenarios where multiple objects are present in an image and overlap with each other, using the Pascal VOC dataset. This comparison demonstrates how each method is able to identify and highlight the regions associated with the target object.

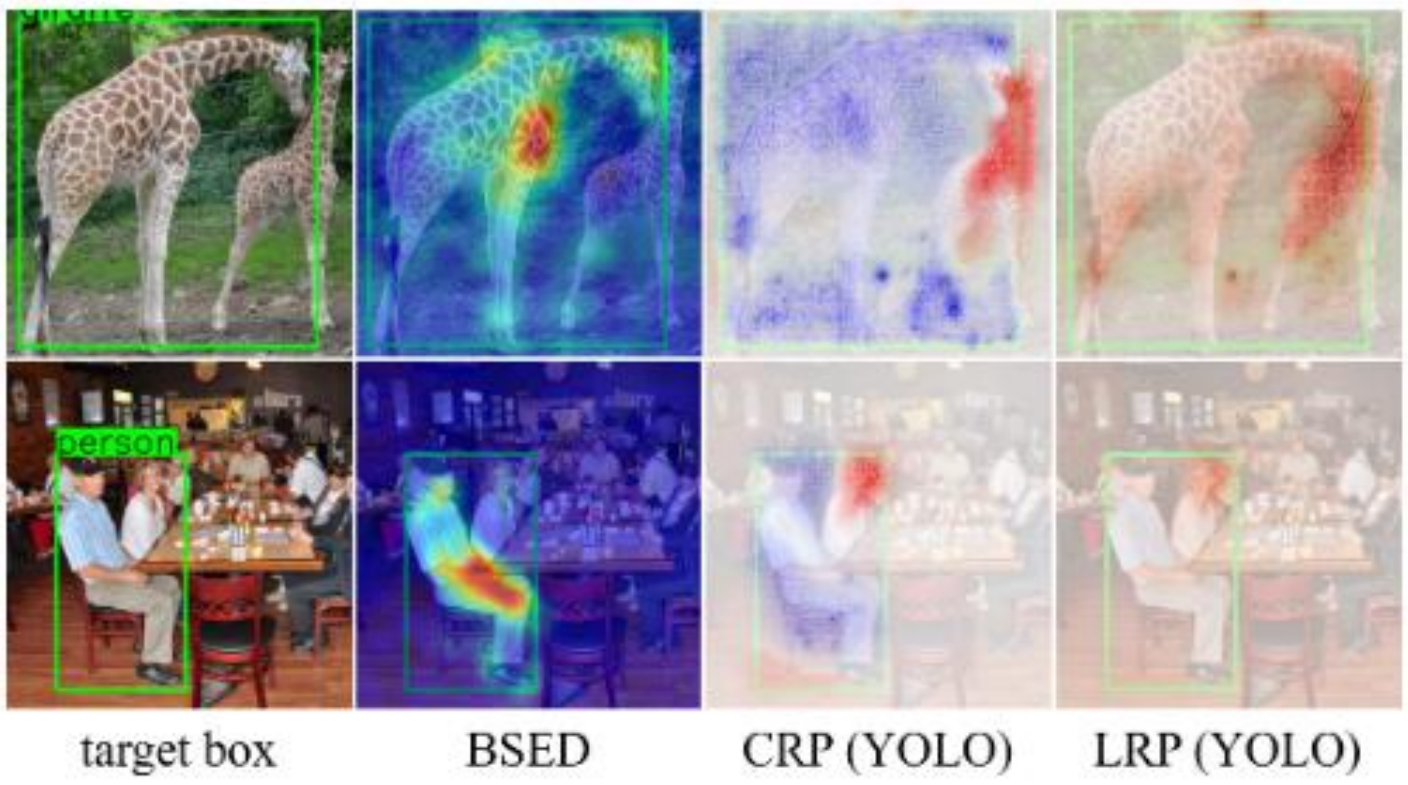

*Figure 46: This image demonstrates the performance of different explainability methods in scenes containing multiple objects and overlapping objects. The BSED method generates focused and precise maps that align well with the target object regions (such as giraffes or a seated person), while CRP and LRP produce more scattered and sometimes misleading maps. These results highlight BSED's ability to generate accurate explanatory maps even in overlapping and cluttered scene conditions [35].*

The figure below presents a visual comparison of the saliency maps generated by various XAI methods, along with the processing time for each method on the MS-COCO dataset. This image effectively illustrates the differences in spatial accuracy and execution speed of each method in real-time applications.

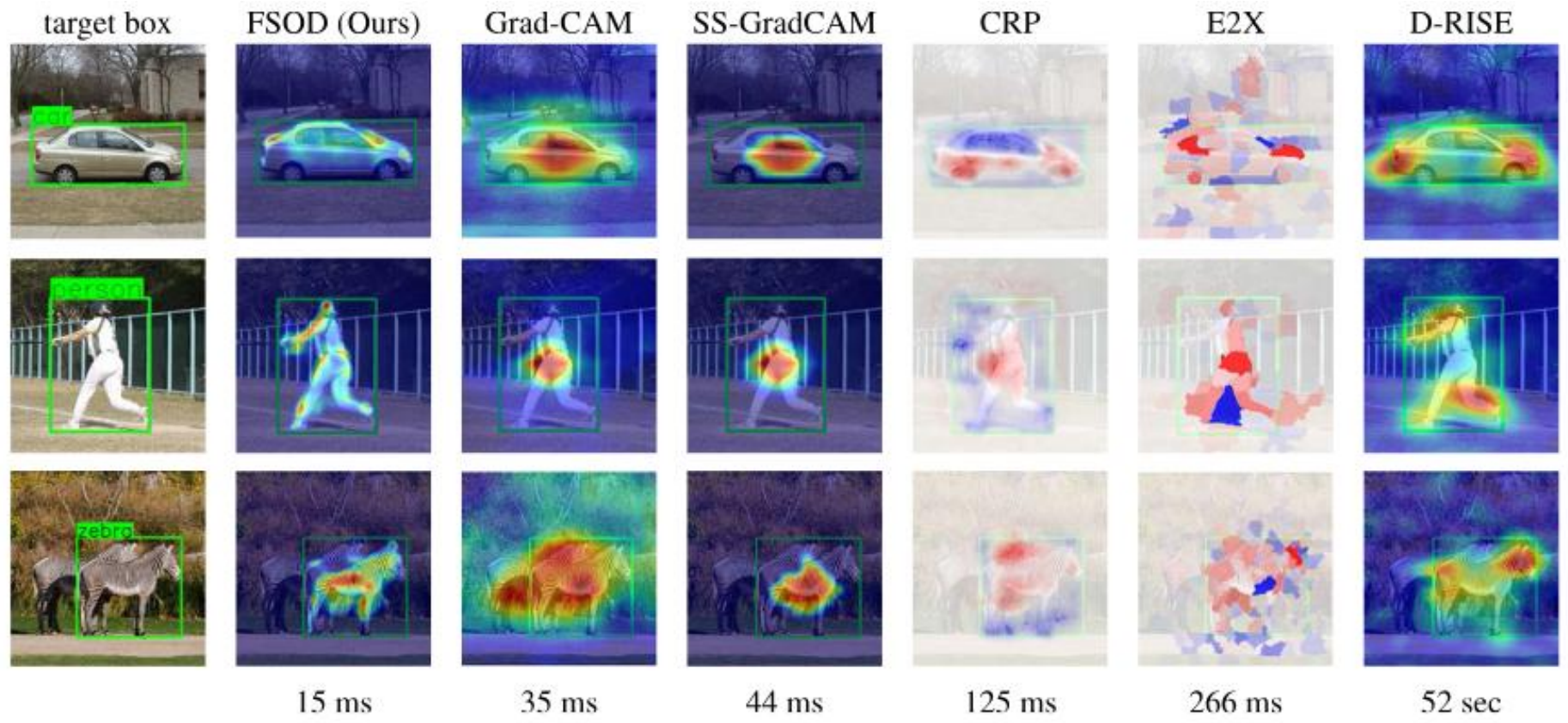

*Figure 47: Visual Comparison of Saliency Maps and Processing Time for Different XAI Methods on the MS-COCO Dataset. The FSOD method generates focused saliency maps aligned with the object region, with a processing time of just 15 milliseconds per sample. In terms of spatial accuracy and speed, it outperforms conventional methods such as Grad-CAM, D-RISE, and KernelSHAP. Although methods like CRP and E2X exhibit relative spatial sensitivity, they are prone to noise and accuracy degradation. These results demonstrate the superiority of FSOD in real-time applications that require precise explainability [9].*

Overall, the presented results demonstrate significant advancements in the field of explainability methods for object detection. Each method has its own strengths and limitations based on various criteria such as spatial accuracy, time efficiency, model behavior fidelity, and human acceptability. Additionally, both visual and quantitative comparisons reveal that the selection of an appropriate method should be based on the specific needs of the problem, the type of data, and the desired evaluation criteria. Ultimately, these reviews provide valuable guidance for researchers and developers to better understand the strengths and challenges involved, aiding in the development of more accurate and practical explainability methods for object detection within the field of XAI.

# 6 Conclusion

In this comprehensive review, we explored a wide range of explainable artificial intelligence (XAI) methods developed for deep learning-based object detection systems. As deep neural networks continue to deliver state-of-the-art performance in detecting and localizing objects, their inherent opacity raises critical concerns about reliability, safety, and trustworthiness—especially in high-stakes domains like healthcare, autonomous vehicles, and security systems.

We systematically categorized XAI methods into four main groups: perturbation-based, gradient-based, backpropagation-based, and graph-based approaches. Each category was thoroughly analyzed in terms of its methodology, advantages, limitations, and applicability to specific object detection architectures such as YOLO, SSD, Faster R-CNN, and EfficientDet. Particular attention was paid to recently proposed techniques like E2X, FSOD, D-CLOSE, and BODEM, which offer promising capabilities for generating fine-grained, object-level explanations in a model-agnostic fashion.

The increasing trend in the number of publications from 2022 to 2025 reflects the growing awareness and importance of model interpretability in the AI research community. Nevertheless, significant challenges remain. These include the computational cost of certain methods, difficulty in evaluating explanation quality, lack of standardized benchmarks, and limited generalization to real-world conditions.

Future research should aim to develop more efficient, scalable, and human-aligned XAI techniques that are robust across various detection tasks and deployment environments. Additionally, integrating human attention data, causal reasoning, and multi-modal explanations could further enhance the interpretability and usability of object detection systems.

Ultimately, achieving trustworthy AI requires not only high-performance models but also transparent and interpretable decision-making mechanisms. XAI will continue to play a pivotal role in shaping the future of responsible and explainable object detection.

**Statements and Declarations**

**Competing Interests:** The authors declare that they have no competing interests.